\documentclass[sigconf,table]{acmart}

\AtBeginDocument{%
  \providecommand\BibTeX{{%
    \normalfont B\kern-0.5em{\scshape i\kern-0.25em b}\kern-0.8em\TeX}}}

\copyrightyear{2022} 
\acmYear{2022} 
\setcopyright{acmlicensed}\acmConference[FAccT '22]{2022 ACM Conference on Fairness, Accountability, and Transparency}{June 21--24, 2022}{Seoul, Republic of Korea}
\acmBooktitle{2022 ACM Conference on Fairness, Accountability, and Transparency (FAccT '22), June 21--24, 2022, Seoul, Republic of Korea}
\acmPrice{15.00}
\acmDOI{10.1145/3531146.3533127}
\acmISBN{978-1-4503-9352-2/22/06}

\usepackage{soul}
\usepackage{amsmath}
\usepackage{stfloats}
\usepackage{graphicx}
\usepackage{xcolor,colortbl}

\usepackage{cleveref} 
\usepackage{booktabs}
\usepackage[T1]{fontenc} 
\usepackage[utf8]{inputenc}
\usepackage{textcomp}
\usepackage{gensymb} 
\usepackage{multirow}
\usepackage{subcaption}

\usepackage[ruled,vlined]{algorithm2e}
\usepackage{pifont}

\definecolor{red_1}{rgb}{1.0,0,0}
\definecolor{red_2}{rgb}{1.0,0.5,0.5}
\definecolor{red_3}{rgb}{1.0,0.75,0.75}
\definecolor{red_4}{rgb}{1.0,0.9,0.9}

\definecolor{lightgray}{gray}{0.955}
\definecolor{gray2}{gray}{0.78}
\let\oldtabular\tabular
\let\endoldtabular\endtabular
\renewenvironment{tabular}{\oldtabular}
{\endoldtabular}

\newcommand{\ra}[1]{\renewcommand{\arraystretch}{#1}}

\renewcommand{\vec}[1]{\mathbf{#1}}

\begin{document}

\title{Human Interpretation of Saliency-based Explanation Over Text}

\author{Hendrik Schuff}
\authornote{Both authors contributed equally to this research.}
\affiliation{%
  \institution{Bosch Center for\\Artificial Intelligence and \\University of Stuttgart}
  \country{Germany}}
\email{hendrik.schuff@de.bosch.com}

\author{Alon Jacovi}\authornotemark[1]
\affiliation{%
  \institution{Bar Ilan University}
  \country{Israel}}
\email{alonjacovi@gmail.com}

\author{Heike Adel}
\affiliation{%
  \institution{Bosch Center for\\Artificial Intelligence}
  \country{Germany}}
\email{heike.adel@de.bosch.com}

\author{Yoav Goldberg}
\affiliation{%
  \institution{Bar Ilan University and \\Allen Institute for\\Artificial Intelligence}
  \country{Israel}
}
\email{yoav.goldberg@gmail.com}

\author{Ngoc Thang Vu}
\affiliation{%
  \institution{University of Stuttgart}
  \country{Germany}}
\email{thang.vu@ims.uni-stuttgart.de}

\renewcommand{\shortauthors}{Schuff and Jacovi, et al.}

\begin{abstract}
While a lot of research in explainable AI focuses on producing effective explanations, less work is devoted to the question of how people understand and interpret the explanation. In this work, we focus on this question through a study of saliency-based explanations over textual data. 
Feature-attribution explanations of text models aim to communicate which parts of the input text were more influential than others towards the model decision. Many current explanation methods, such as gradient-based or Shapley value-based methods, provide measures of importance which are well-understood mathematically.
But how does a person receiving the explanation (the explainee) comprehend it? And does their understanding match what the explanation attempted to communicate? 
We empirically investigate the effect of various factors of the input, the feature-attribution explanation, and visualization procedure, on laypeople's interpretation of the explanation.
We query crowdworkers for their interpretation on tasks in English and German, and fit a GAMM model to their responses considering the factors of interest.
We find that people often mis-interpret the explanations: superficial and unrelated factors, such as word length, influence the explainees' importance assignment despite the explanation communicating importance directly.
We then show that some of this distortion can be attenuated:
we propose a method to adjust saliencies based on model estimates of over- and under-perception, and explore bar charts as an alternative to heatmap saliency visualization.
We find that both approaches can attenuate the distorting effect of specific factors, leading to better-calibrated understanding of the explanation.
\end{abstract}

\ccsdesc[500]{Human-centered computing~Empirical studies in visualization}
\ccsdesc[500]{Computing methodologies~Natural language processing}
\ccsdesc[500]{Computing methodologies~Machine learning}

\keywords{feature attribution, text, saliency, explainability, interpretability, human, perception, cognitive bias, generalized additive mixed model}

\maketitle

\section{Introduction}
Machine learning models' application in various domains (e.g., criminal justice and healthcare) has motivated the development of explanation methods to understand their behavior.
One popular class of explanation methods
explains model decisions by specifying the parts of the input which are most salient in the model's decision process \cite{DBLP:journals/jair/BurkartH21,PMID:33079674,DBLP:journals/corr/abs-2112-04417}.
In natural language processing (NLP), this refers to
which words, phrases or sentences in the input contributed
most to the model prediction \cite{DBLP:journals/corr/abs-2108-04840,DBLP:conf/ijcnlp/DanilevskyQAKKS20}.
While much research exists on developing and verifying such explanations \cite{arras__2017,adebayo_sanity_2018,kindermans_reliability_2019,tuckey_saliency_2019,wang_gradient-based_2020,madsen_evaluating_2021}, less is known about the information that human explainees actually understand from them \cite{miller19-social,dinu_challenging_2020,DBLP:journals/corr/abs-2111-04138,DBLP:journals/corr/abs-2112-09669}.
\begin{figure*}
  \centering
  \includegraphics[width=0.89\textwidth]{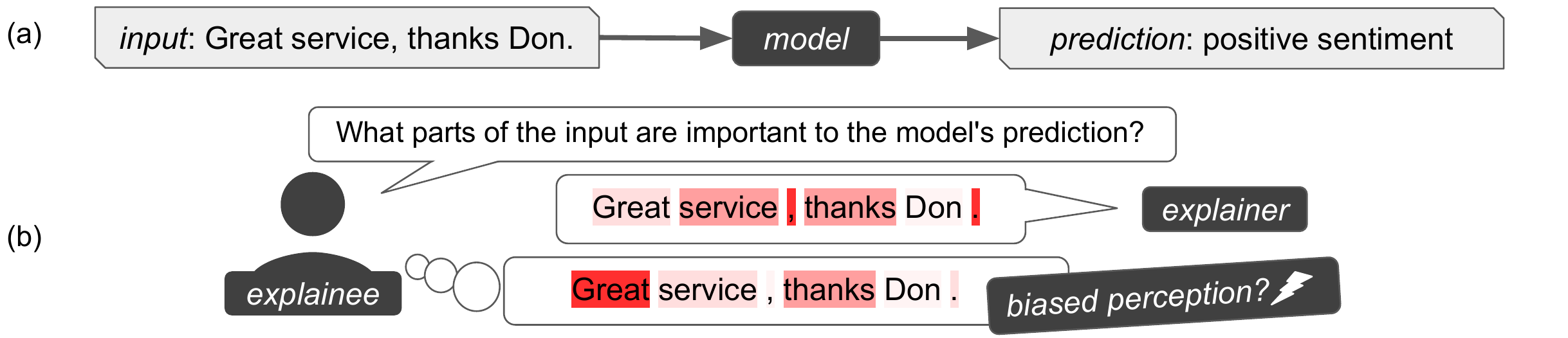}
   \caption{
   A saliency explantion is generated to answer the human's need to understand the model.
   We investigate whether the saliency explanation can be systematically mis-perceived by humans and which factors influence its perception.
   }
  \label{fig:teaser}
\end{figure*}

In the explainable NLP literature, it is generally (implicitly) assumed that the explainee interprets the information ``correctly'', as it is communicated \cite{arras__2017, DBLP:conf/iui/FengB19, DBLP:journals/corr/abs-2111-04138}:
e.g., when one word is explained to be influential in the model's decision process, or more influential than another word, it is assumed that the explainee understands this relationship \cite{DBLP:journals/tacl/JacoviG21}. 
We question this assumption: research in the social sciences describes modes in which the human explainee may be biased---via some cognitive habit---in their interpretation of processes \cite{malle2003,miller19-social,epley2007-anthropomorphism-three-factor,watson2020-anthropomorphism-rhetoric}. Additional research shows this effect manifests in practice in AI settings \cite{DBLP:conf/acl/GonzalezRS21,darling2015-anthropomorphism-robots,ehsan2021-ai-experts-explanation-perception,Hartzog2015UNFAIRAD,10.1145/3319502.3374839}.
This means, for example, that the explainee may underestimate the influence of a punctuation token, even if the explanation reports that this token is highly significant (Figure~\ref{fig:teaser}), because the explainee is attempting to understand how the model reasons \emph{by analogy to the explainee's own mind} which is an instance of
\emph{anthropomorphic bias} \cite{johnson2018-anthropomorphic-bias,Dacey2017-anthropomorphic-bias,Zlotowski2015AnthropomorphismOA} and \textit{belief bias} \cite{evans_conflict_1983,DBLP:conf/acl/GonzalezRS21}.

We identify three different such biases which may influence the explainee's interpretation: (i) \textit{anthropomorphic bias} and \textit{belief bias}: influence by the explainee's self projection onto the model; (ii) \textit{visual perception bias}: influence by the explainee's visual affordances for comprehending information; (iii) \textit{learning effects}: observable temporal changes in the explainee's interpretation as a result of interacting with the explanation over multiple instances.

We thus address the following question in this paper: \emph{When a human explainee observes feature-attribution explanations,
does their comprehended information differ
from what the explanation ``objectively'' attempts to communicate? If so, how?}

We propose a methodology to investigate whether explainees exhibit biases when interpreting feature-attribution explanations in NLP, which effectively distort the objective attribution into a subjective interpretation of it (Section~\ref{sec:specification}).
We conduct user studies in which we show an input sentence and a feature-attribution explanation (i.e., saliency map) to explainees, ask them to report their subjective interpretation, and analyze their responses for statistical significance across multiple factors, such as word length, total input length, or dependency relation, using GAMMs (Section~\ref{sec:study_all}).\footnote{We release the collected data and analysis code: \url{https://github.com/boschresearch/human-interpretation-saliency}.}

We find that word length, sentence length, the position of the sentence in the temporal course of the experiment, the saliency rank, capitalization, dependency relation, word position, word frequency as well as sentiment can significantly affect user perception.
In addition to \textit{whether} a factor has a significant influence, we also investigate \textit{how} this factor affects perception.
We find that, for example, short words overall decrease importance ratings while short sentences or intense sentiment polarities increase them.

Finally, we propose two visualization interventions to mitigate learning effect and visual perception biases:
model-based color correction and bar charts.
We conclude that (a) model-based color correction can predict and mitigate distorting temporal effects and (b) bar charts can successfully remove the influence of word length.

Overall, our results show that
\emph{supposedly irrelevant factors such as word length do affect how explainees perceive the influence of words in feature-attribution explanations, despite the explanations explicitly communicating this influence.}
This is a surprising result, which raises
important questions for explainability in NLP, and
in general, about the ability of feature-attribution tools available today to 
convey
the information that they intend to communicate: even in the case of a relatively straightforward explanation, such as directly informing importance regions in the input, cognitive biases of explainees run deep, and may erroneously affect the understanding of the given information.

We show that bar charts and color correction result in better-aligned human assessments in our setting on multiple bias factors.
We urge researchers to not blindly trust that users perceive explanations as communicated, and to investigate if our findings transfer to their respective audience and context.

\section{Feature-Attribution Explanations}
Feature-attribution explanations aim to convey which parts of the input to a model decision are ``important'', ``responsible'' or ``influential'' to the decision \cite{arras__2017,ribeiro16-lime,diogo2019-survey,DBLP:journals/corr/abs-2108-04840,DBLP:journals/tetci/ZhangTLT21}. This class of methods is a prevalent mode of describing NLP processes \cite{DBLP:journals/corr/abs-2108-04840,DBLP:conf/ijcnlp/DanilevskyQAKKS20,10.1145/3313831.3376219,tenney2020language}, due to two main strengths: (1) it is flexible and convenient, with many different measures developed to communicate some aspect of feature importance; (2) it is intuitive, with---seemingly, as we discover---straightforward interfaces of relaying this information.  Here we cover background on feature-attribution  on two fronts: the underlying technologies (Section~\ref{subsec:mechanism}) and the information which they communicate to humans (Section~\ref{subsec:function}). 

\subsection{Attribution Methods} \label{subsec:mechanism}
We consider feature-attribution explanations generally as scoring (or ranking) functions that map portions of the input to scores that communicate some aspect of importance about the aligned portion:
$E_f(f(\mathbf{x})):\Sigma^{n}\rightarrow \mathbb{R}^{n},$
where $E_f$ is the explanation method with respect to $f$, $f$ is the model and $\mathbf{x} \in \Sigma^n$ the input text to the model,
i.e., the input consists of $n$ tokens which are are elements of an alphabet $\Sigma$.
For simplicity, we assume that a high score implies high importance.

The loose definition proposed above for feature-attribution explanations as communicating ``important'' portions of the input (words, sub-words, or characters)
is often interpreted with causal lens: that by intervening on the tokens assigned a high score, the model behavior will change more than by intervening on the tokens assigned a low score \cite{DBLP:journals/tacl/JacoviG21,grimsley-etal-2020-attention,arras__2017}.
This perspective is relaxed in various ways to produce various softer measures of importance: for example, \textit{gradient-based methods} measure the change required in the embedding space to cause change in model output, while \textit{Shapley-value methods} measure the change with respect to the ``average case'' in the data.

The granularity provided in the scoring function may vary greatly, from a binary measure---important or not important---to a complete saliency map, depending on the tokenization granularity, the method and visualization. Most commonly, the explanation is given as a colorized saliency map over word tokens \cite[e.g.,][]{arras__2017,wang_gradient-based_2020,tenney2020language,DBLP:journals/corr/abs-2112-09669}.
Note that this work is \textit{not} concerned with a particular feature-attribution method, but rather how feature-attribution explanations generally communicate information to human explainees, and what the explainees comprehend from them.

\subsection{Social Attribution: The Case of Text Marking}  \label{subsec:function}
Is it really possible for the explainee to comprehend feature-attribution explanations differently from what they objectively communicate? What is the nature of any discrepancy in this perception?\footnote{This question is distinct from the question of whether the explanation faithfully communicates information about the model \cite{jacovi2020-faithfully,wiegreffe-pinter-2019-attention}: even if the feature-attribution information is entirely faithful, discrepancies may still arise in how humans comprehend this information.} As \citet{miller19-social} writes, literature in the social sciences about how humans comprehend explanations and behavior can help illuminate this problem.

In particular, we assume that the human explainee comprehends the explanation with respect to their own reasoning. By assigning human-like reasoning to the model behavior being explained
\cite{miller19-social},
the explainee may fill any incompleteness in the explanation with assumptions from their own priors about what is plausible to them \cite{Dacey2017-anthropomorphic-bias,DBLP:conf/acl/GonzalezRS21}. 

To demonstrate, consider the case of binary feature-attribution---marking parts of the input as ``important'' and ``not important'', also known as \textit{highlighting} or \textit{extractive rationalization} \cite{DBLP:conf/emnlp/LeiBJ16}.
Even this simple format of communicating information can be assigned human-like reasoning by the explainee, on account of ``\textit{who marked this text}'' and ``\textit{for what purpose}'': \citet{Marzouk2018TextMA}
identifies various objectives that humans follow when marking or observing marked text, e.g., marking forgettable secions (for memorization); marking as a summary (for subsequent reading); marking exemplifying text; marking contradicting or surprising text, etc.
In the context of NLP models, \citet{DBLP:journals/tacl/JacoviG21} note two possible central objectives: reducing the input to a summary which comprehensively informs the decision, or identifying influential evidence in the input which non-comprehensively supports the decision.

These many different objectives can influence the choice of marking, and the information that it communicates.
This means that both the marked text, and the choice of what text to mark, are information which the explainee comprehends when observing the explanation. Therefore, how the explanation is perceived is influenced by both factors.

Text marking is a special case of feature-attribution. The above demonstrates how the explainee's interpretation is potentially shaped by aspects of the explanation which are implicit or unintended---leading to an ``erroneous'' interpretation of the explanation. We identify three biases that may cause this effect, as motivation for our investigation: (i) anthropomorphic bias and belief bias, via the explainee's a-priori opinion on human-like or plausible reasoning; (ii) visual perception bias, via characteristics of the explainee's visual affordances for comprehending information; (iii) learning effects, as observable influence in the explainee's interpretation by previous explanation attempts in-context.

\section{Study Overview}
\begin{figure*}
    \centering
    \fbox{\includegraphics[width=0.7\textwidth]{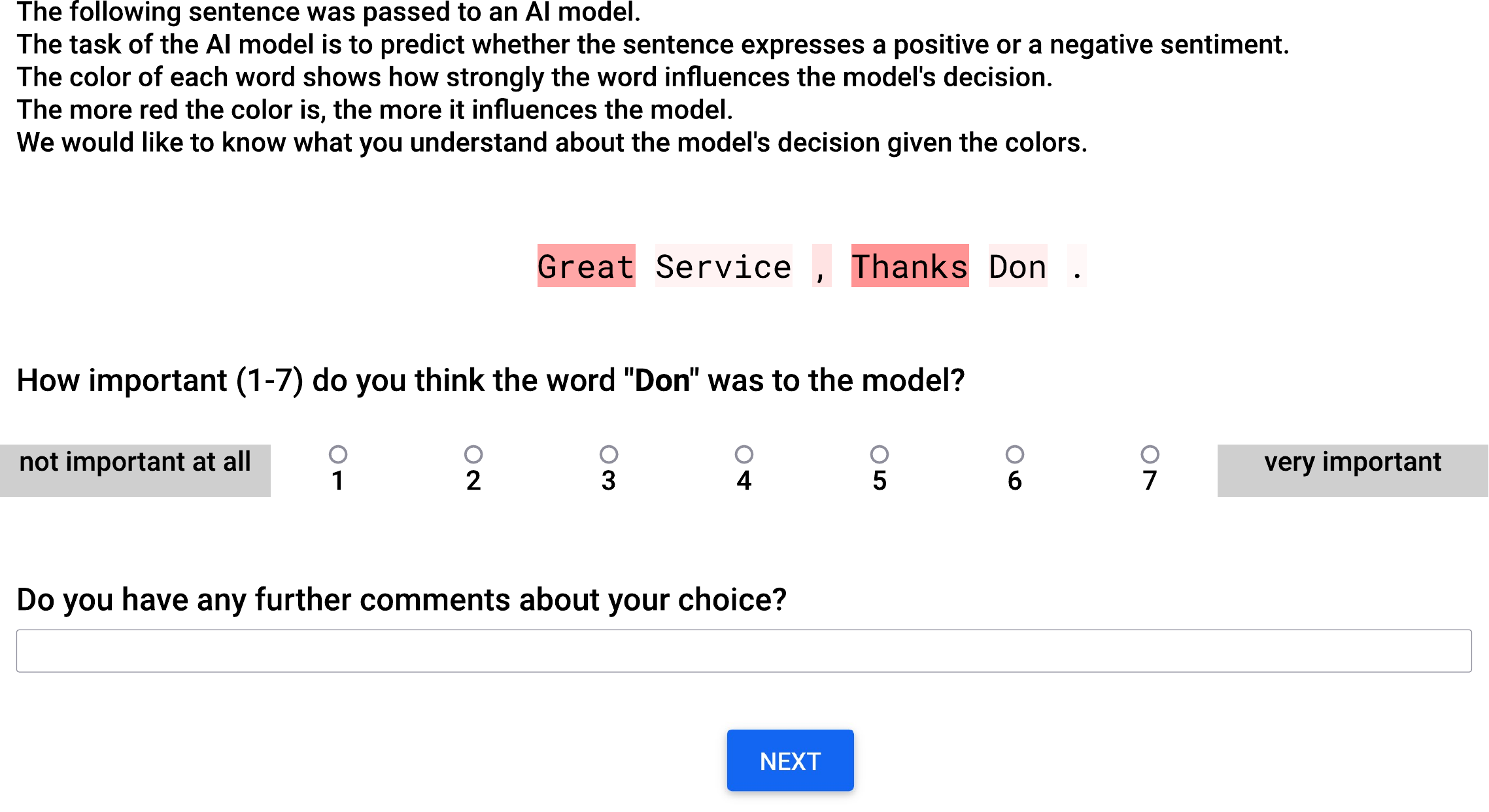}}
    \caption{Screenshot of the importance rating interface for English sentiment sentences using saliency visualization.}
    \label{fig:screenshot_en_saliency}
\end{figure*}
\textit{Research Question.}
The core research question in this work is to probe
into which, if any, factors in the explanation process---aside from the saliency itself---may influence the explainee's interpretation of the saliency information.
Formally, we view the saliency explanation as a process whose
result is the explainee's interpretation of the saliency scores. 
The ``input'' to this process is the original text as well as the saliency information and the visualization method.
Then, we ask which factors in the original text have statistically significant effects on the explainee's interpretation and how properties of the saliency score and the visualization method affect it. 

Notably, a key challenge in analyzing the explainees' saliency understanding is that we want to identify influencing factors on the explainee's ratings without the existence of an inherently correct ground truth perception.\vskip5pt

\noindent\textit{Proposed Methodology.}
We propose a combination of study design and statistical analysis to quantify the influence of arbitrary factors such as word length, sentiment polarity or dependency relations.
We collect explainees' subjective interpretations of the saliency scores in a crowdsourcing setup.
We
relate this interpretation to the original explanation
considering various potentially
influencing factors
using an ordinal generalized additive mixed model (GAMM).
The result from this comparison is an answer on \textit{which} of the a-priori candidate explanatory factors indeed have significant effect on the explainee's interpretation and \textit{how} these factors functionally affect interpretation.

\section{Study Methodology Specification} \label{sec:specification}

The study consists of two phases: collecting subjective importance interpretations (Section~\ref{subsec:collection-methodology}), and
analyzing responses with an adequate statistical model
(Section~\ref{subsec:analysis-methodology}).
We release the the collected data and analysis code.

\subsection{Collecting Self-Reported Importance Ratings} \label{subsec:collection-methodology}
In our main study, we investigate the interpretation of color-coding saliency visualization of the feature-attribution by crowdsource laypeople (variations on this study will be described later).
We measure the perceived importance of a word within a saliency score explanation by directly probing human self-reported word importance.
In this instance, we ask ``How important (1-7) do you think the word "X" was to the model?'' (Figure~\ref{fig:screenshot_en_saliency}).
We collect answers on a single-item unipolar 7-point Likert scale ranging from \textit{not important at all} to \textit{very important}.
\\[0.4em] 
\noindent \textit{Texts.} \
We use sentences from the Universal Dependencies English Web Treebank \cite{silveira14gold}.\footnote{\url{https://github.com/UniversalDependencies/UD_English-EWT}}
This treebank contains comprehensive annotation, including
dependency relations of sentences, stemming from various domains such as newsgroups or online reviews.
We use sentences from the reviews group for a plausible framing of a sentiment analysis task.\footnote{We choose sentences without sub-token dependency relations (e.g., excluding ``it's'' because displaying it as two tokens breaks the orthography) and with unique word occurrences (i.e., excluding sentences that contain a word several times). From this subset we remove length outliers: sentences with number of tokens longer than one standard deviation above the mean (concretely, 11 tokens).}
We randomly select 150 sentences to be used.
\\[0.4em] 
\noindent \textit{Saliency Scores.} \
We assign random saliency scores to each token to uniformly sample the space of saliency intensities.
We are, at this stage, not interested in using a ``real'' model or saliency score (e.g., attention or integrated gradients), as we investigate general perception of arbitrary scores.
It is therefore useful to create saliency scores that ``do not make sense'' because a saliency score should reflect the model's reasoning which might very well not make sense at all.
We study an instance of ``real'' saliency scores (integrated gradients) later in Section~\ref{sec:study-ig}.
\\[0.4em] 
\noindent \textit{Study Interface.} \
See Figure~\ref{fig:screenshot_en_saliency} for the rating collection interface. We display all sentences using monospaced font and fixed whitespaces to obtain a direct mapping between the number of characters and the color area for each word.\footnote{Ligatures and other typographic attributes of non-monospaced fonts would break this mapping.}
\\[0.4em] 
\noindent \textit{Procedure.} \
We ask participants to rate the importance of a randomly-selected word in the sentence.\footnote{Alternatively, one can imagine a setting in which participants rate all words within the sentence.
We choose to ask for single-word ratings to (i) avoid carry-over effects from ratings of the first to the last words and (ii) collect ratings of more sentences within the same experiment time compared to splitting the set of sentences over participants which would introduce further difficulty in the statistical analysis.} We show all 150 sentences from the described review dataset to each participant,
displayed in a randomized order per participant. Saliency scores for all tokens are randomized for each participant (such that we collect responses to many different saliency maps, rather than numerous responses for the same set). 
We do so because our aim is not to obtain accurate (mean) estimates of single ratings as one would do in a corpus annotation, but to collect rich data to build an accurate model describing the underlying general phenomenon.
For each sentence, we collect the participant's importance rating, the completion time and a voluntary free-text comment. 
We choose to not include a dedicated training phase, e.g., showing the participants ten explanation instances before starting the data collection: we are explicitly interested in potential learning effects.
These can be crucial in real-world applications: for example, should we find a decaying learning effect, an effective model audit should make sure to include a sufficient number of model predictions.\footnote{In order to filter-out (a) participants that just ``click through'' the interface to obtain the study reward and (b)  noisy responses due to decreased participant attention towards the end of the experiment, we insert three trap sentences at random positions in the last two thirds of the real sentences. See example and more integration details in Figure~\ref{fig:screenshot_trap_question} in Appendix~\ref{sec:appendix_interface}.}
\\[0.4em] 
\noindent \textit{Participants.} \
We recruit 50 crowdworkers on Mechanical Turk.
One crowdworker failed all of the trap sentences, so we exlude this worker's responses and recruit one additional worker.
All other participants successfully passed all trap sentences.
In total, this yields 7500 importance ratings.

\subsection{Factors of Saliency Perception}
For our set of possible candidate factors, we model factors which are motivated by the three types of biases: anthropomorhic and belief biases, visual biases, and learning effects. Each factor will be
tested for statistical significance on the explainees' interpretations. Table~\ref{tab:factors} lists the factors we investigate in this work.

Selected factors in Table~\ref{tab:factors} include:
(i) \textit{word length} as longer words correspond to a larger colored draw area, which we hypothesize influences visual perception bias; 
(ii) \textit{word polarity} as we present participants a sentiment classification task and expect that the participants' own assessment of word importance influences their perception of how important it is to the model, which we hypothesize is an instance of belief bias; 
(iii) \textit{display index} as we hypothesize that participant ratings are affected by temporal effects such as learning;
(iv) \textit{word position} as we hypothesize that, e.g., words at the center of a sentence might be perceived more strongly due to the center bias (visual perception bias) which was observed in various eye-tracking studies, i.a., for natural scenes \cite{tseng_quantifying_2009}.\footnote{We derive word frequencies from the WikiMatrix corpus \cite{schwenk_wikimatrix_2021} and sentiment polarities from SentiWords \cite{gatti_sentiwords_2016}.}  
\begin{table*}
\centering
\ra{1.3}
\caption{List of factors that presupposedly affect saliency explanation perception along with the findings of our three user studies. EN refers to the English sentiment classification study, DE to the German fact checking study and EN-IG to the English sentiment classification study using integrated gradients as feature attribution method (without correction visualizations).}\label{tab:factors}
\resizebox{0.99\linewidth}{!}{%
\begin{tabular}{p{0.2\linewidth}p{0.7\linewidth}ccc}
\toprule
\textbf{Factor} & \textbf{Description} & \multicolumn{3}{c}{\textbf{Significant Effects}}\\ 
 & & \textit{EN} & \textit{DE} & \textit{EN-IG}\\
\midrule
Saliency & The color intensity specified as the saturation value ($S \in [0,1]$) in a $(H,S,V)$ color triple \cite{smith_color_1978}, e.g., (0\degree,0.5,1.0) (\textcolor{red_2}{$\blacksquare$}) and (0\degree,0.25,1.0) (\textcolor{red_3}{$\blacksquare$}). & \ding{51} & \ding{51} & \ding{51}\\
\rowcolor{lightgray}Word length & The number of characters in a word, e.g., 7 for ``example''. & \ding{51} & \ding{51} & \ding{51}\\
Word frequency & The word's normalized frequency, estimated on a large corpus.
&  &  & \ding{51}\\
\rowcolor{lightgray}Sentence length & Number of words in the sentence. & \ding{51} & \ding{51} &\\
Display index & The sentence's position within a sequence of sentences (e.g., the third sentence in the sequence of 150 sentences). This relates to temporal effects such as learning. & \ding{51} & \ding{51} &\\
\rowcolor{lightgray}Sentiment polarity & The sentiment polarity of a word (defined via its lemma) $\in [-1,1]$. & \ding{51} & -- &\\
Saliency rank & Normalized rank of a word's saliency score (i.e. color intensity) in comparison to the other words in its sentence $\in [0,1]$. & \ding{51} &  & \ding{51}\\
\rowcolor{lightgray}Word position & The index of the token's position within its sentence. & & \ding{51} &\\
Capitalization & The word's capitalization, e.g., ``example'', ``Example'' or ``EXAMPLE''. & & \ding{51}& \\
Dependency relation & Dependency relation to its parent within the dependency graph (36 types for \textit{EN}). & & \ding{51}& \\
\bottomrule
\end{tabular}
}
\end{table*}

\subsection{Statistical Analysis Using GAMMs} \label{subsec:analysis-methodology}
Given a set of inputs for which there are the feature-attribution scores, and the interpreted importance scores, we describe the analysis methodology aiming to derive the possible input factors that cause discrepancy between the two.

\subsubsection{Ordinal Generalized Additive Mixed Model}
We analyze the collected ratings of perceived importance using a ordinal generalized additive mixed model (GAMM).
Its key properties are that it (i) models the ordinal response variable (i.e., the importance ratings in our setting) on a continuous latent scale (\textit{ordinal generalized}), which is (ii) modeled as a sum of smooth functions of covariates (\textit{additive}) and (iii) accounts for random effects (\textit{mixed}).
The continuous latent scale is linked to ordinal categories by estimating threshold values that separate neighboring categories.
The smooth functions can comprise single covariates (\textit{univariate} smooths) such as $f_1(x_{1})$ or combinations of multiple covariates such as $f_2(x_{2}, x_{3})$.
Random effects allow to account for, e.g., systematic differences in individual participants' rating behaviour.
For example, a specific participant might have a tendency to give overall higher ratings than other participants.
Including a \textit{random effect} allows to disentangle this influence on the response variable from the influence of the covariates in question (such as word length) and thereby offers a clearer view on these \textit{fixed effects}.
The GAMM analysis enables us (i) to make statements about which factors significantly influence saliency perception, without prescribing any notion of ``correct perception'' and (ii) to study the relation between these factors and participants' importance ratings in detail, via an interpretation of the model's parametric terms (categorical factors) as well as smooth terms (numeric factors).
We provide a description of each of the ordinal GAMM's components in Appendix~\ref{sec:introduction_gamm}.\footnote{For further information on ordinal GAMMs, we refer to \citet{divjak_ordinal_2017}, who provide a comprehensive introduction.
For detailed information on GAM(M)s as well as explanations of implementations and analyses, we recommend the textbook by \citet{wood_generalized_2017}.}

\subsubsection{Model Details}
We include all factors listed in Table~\ref{tab:factors} into our model formula.
We use smooth terms for numeric factors and parametric terms for categorical factors.
Additionally, we include tensor product interactions for all pairs of smooth terms.\footnote{Such a functional ANOVA decomposition is supported by mgcv and allows to study, e.g., the interaction between word length and sentiment polarity in addition to the isolated main effects of word length and sentiment polarity.}
In order to statistically account for potentially confounding effects of individual participants or sentences, we include random intercepts as well as random slopes for each participant and each sentence.
Before fitting the model, we remove a small amount of outlier ratings.\footnote{We remove outliers from the intially 7500 importance ratings by excluding words with 20 or more characters (8 ratings) and ratings with a completion time of 60 seconds or more (50 ratings), leaving 7442 ratings left for analysis.
We apply the identical filters to the study described in Section~\ref{sec:bias_mitigation}. For the German study described in Section~\ref{sec:study_2}, we only apply the completion time filter.}
We use fast REML for smoothness selection and apply variable selection via double-penalty shrinkage (i.e., additionally penalizing the splines' null space).
We fit the model using discretized covariates as described in \citet{wood_generalized_2017-1} and \citet{li_faster_2020}.%
\footnote{We use R and mgcv \cite{mgcv1,mgcv2,mgcv3,wood_generalized_2017,mgcv5} (version 1.8-38) to fit all our models.}

\section{Study Results, Interpretation and Generalizations}\label{sec:study_all}
In the following, we conduct three user studies.
The first study (Section~\ref{sec:study_en}) investigates saliency perception for English and a sentiment classification task.
The second study (Section~\ref{sec:study_2}) extends the investigation to German language and a fact checking task to evaluate generalization of the findings.\footnote{\textit{Task} refers to the AI's task which operation is communicated to the explainee via the saliency explanation.}
Since these two studies use random saliency scores so as to not prescribe a specific feature-attribution method, we report a third study (Section~\ref{sec:study-ig}) which uses the wide-spread integrated gradient scores as a generalization to practically-used attribution methods.

\subsection{Sentiment Analysis in English}\label{sec:study_en}
We discuss quantitative results based on the fitted GAMM (Section~\ref{sec:study_en_quantitative}) as well as qualitative findings based on the participants written feedback (Section~\ref{sec:study_en_qualitative}).

\subsubsection{Quantitative}\label{sec:study_en_quantitative}
\begin{figure*}
    \centering
    \begin{subfigure}[t]{.23\textwidth}
        \centering
    \includegraphics[width=\textwidth]{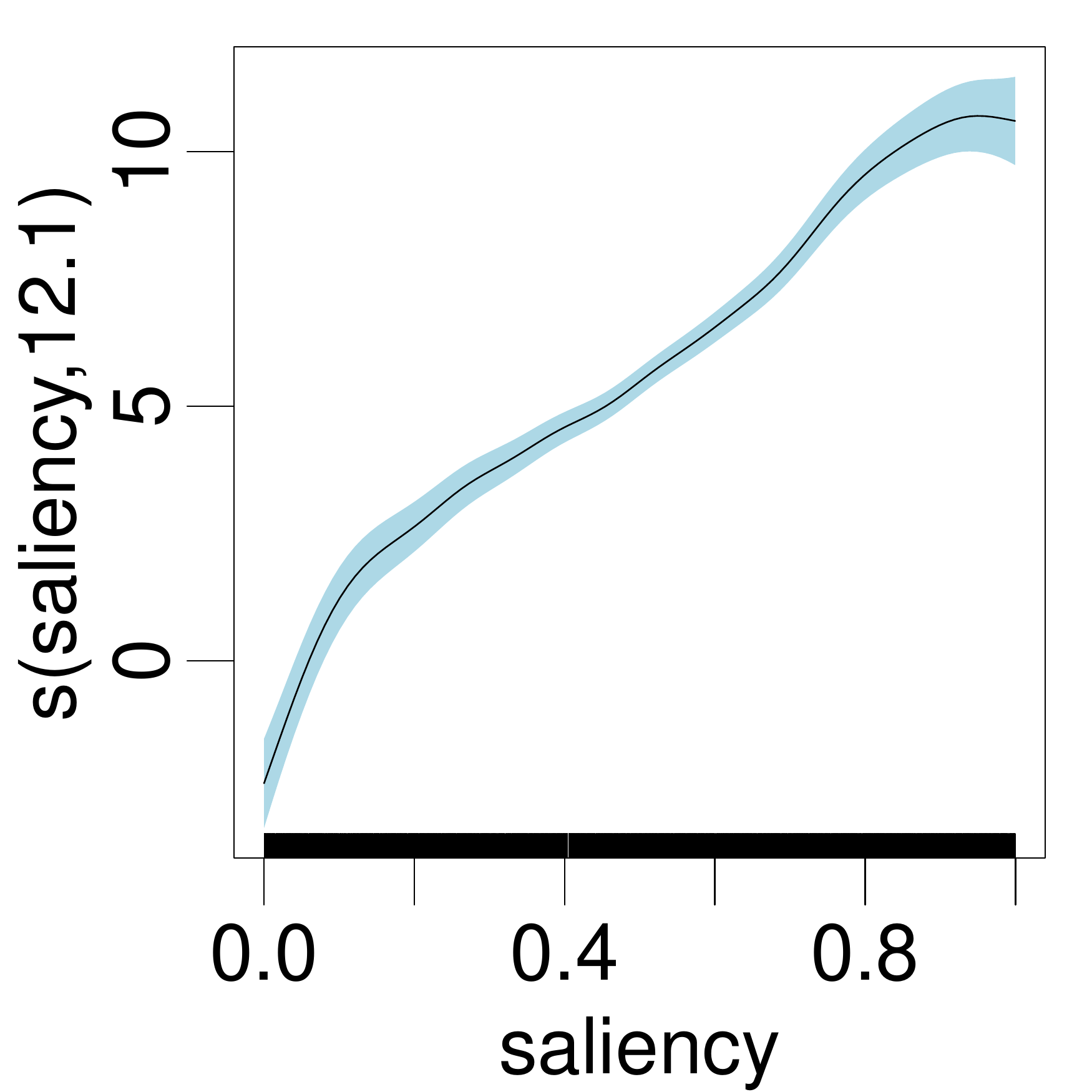}
    \caption{Saliency (saturation)}\label{fig:partial_effects_saliency_score}
    \end{subfigure}%
    \hspace{1cm}
    \begin{subfigure}[t]{.22\textwidth}
        \centering
    \includegraphics[width=\textwidth]{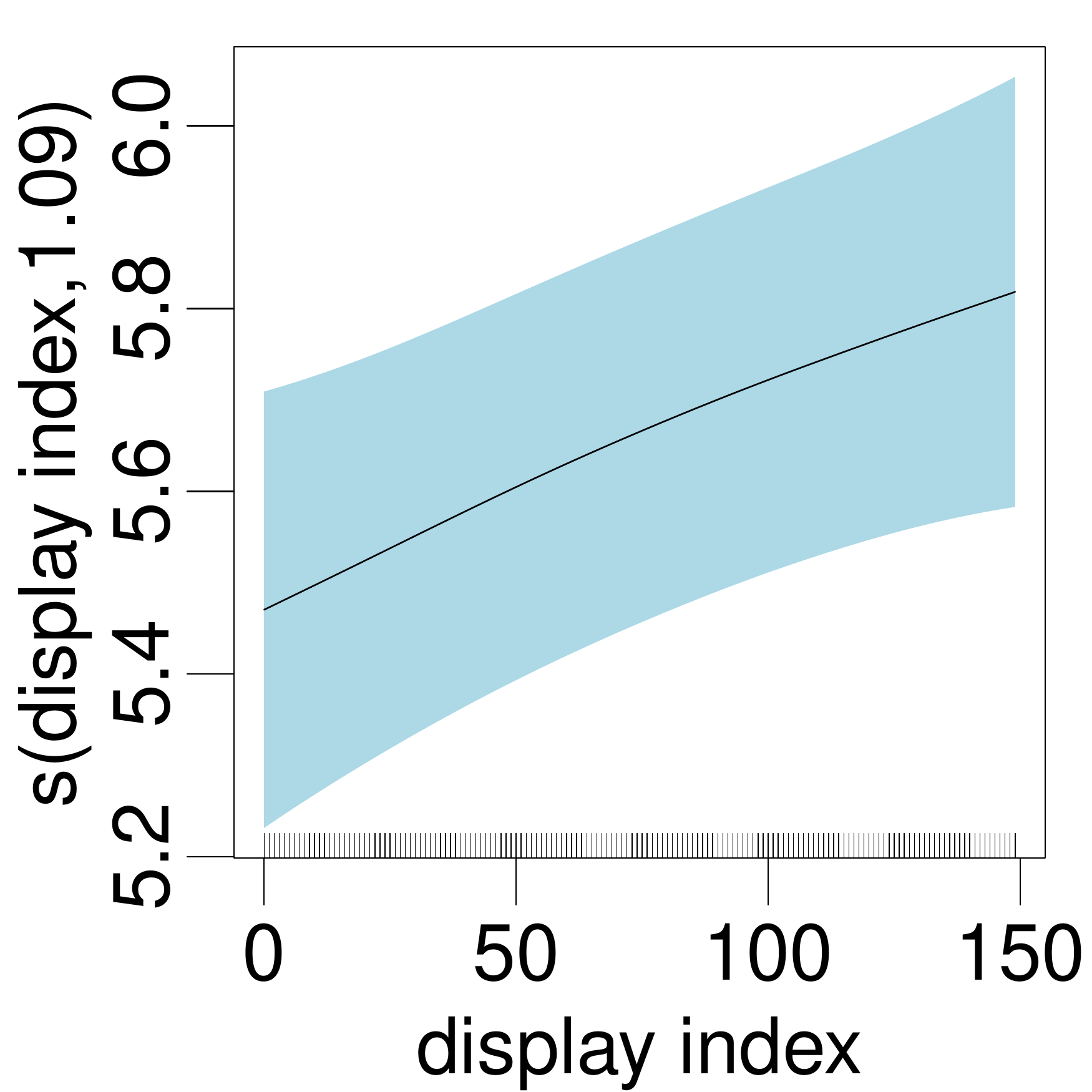}
    \caption{Temporal display index}\label{fig:partial_effects_display_index}
    \end{subfigure}%
    \hspace{1cm}
    \begin{subfigure}[t]{.22\textwidth}
        \centering
    \includegraphics[width=\textwidth]{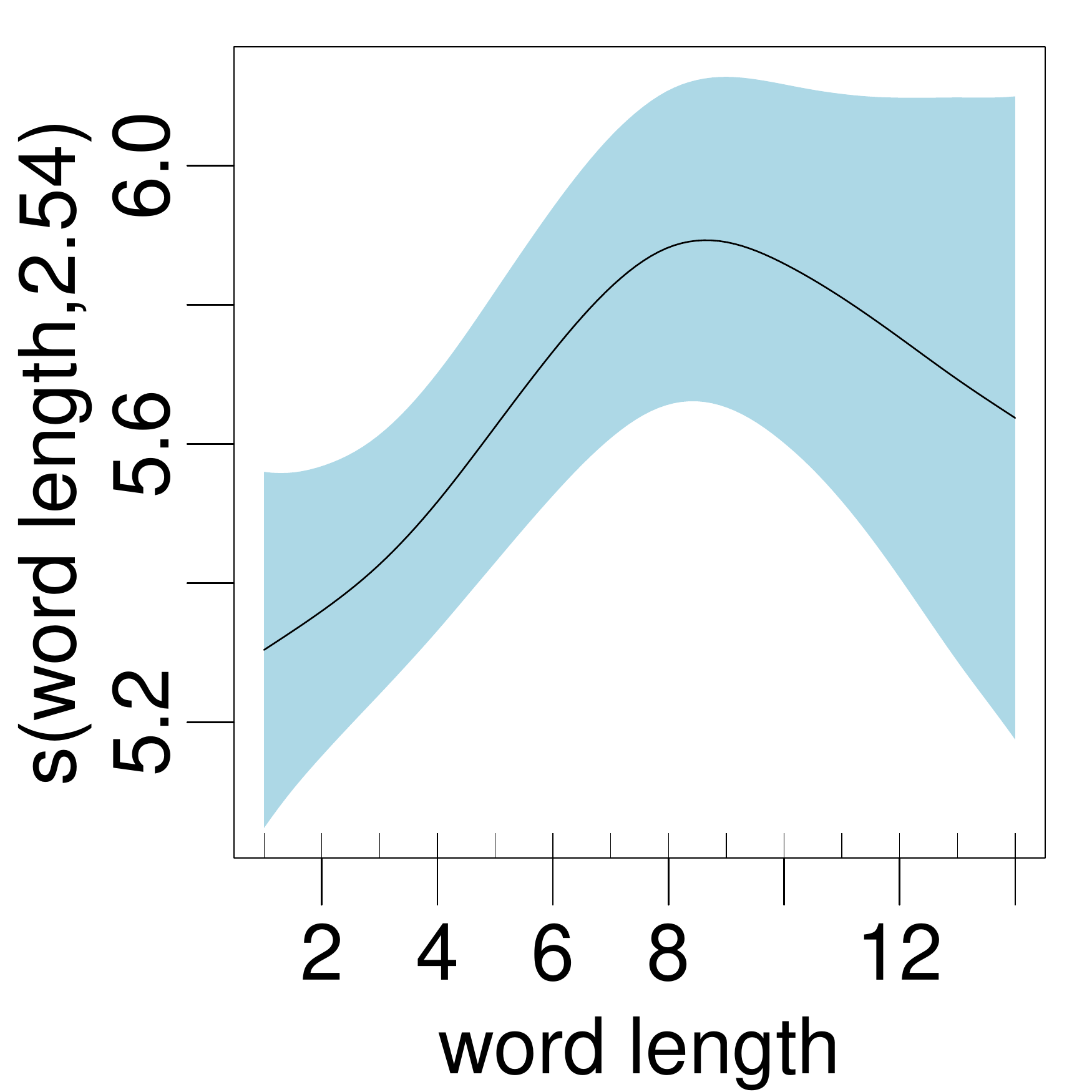}
    \caption{Word length}\label{fig:partial_effects_word_length}
    \end{subfigure}
    \hspace{1cm}
    \begin{subfigure}[t]{.22\textwidth}
        \centering
    \includegraphics[width=\textwidth]{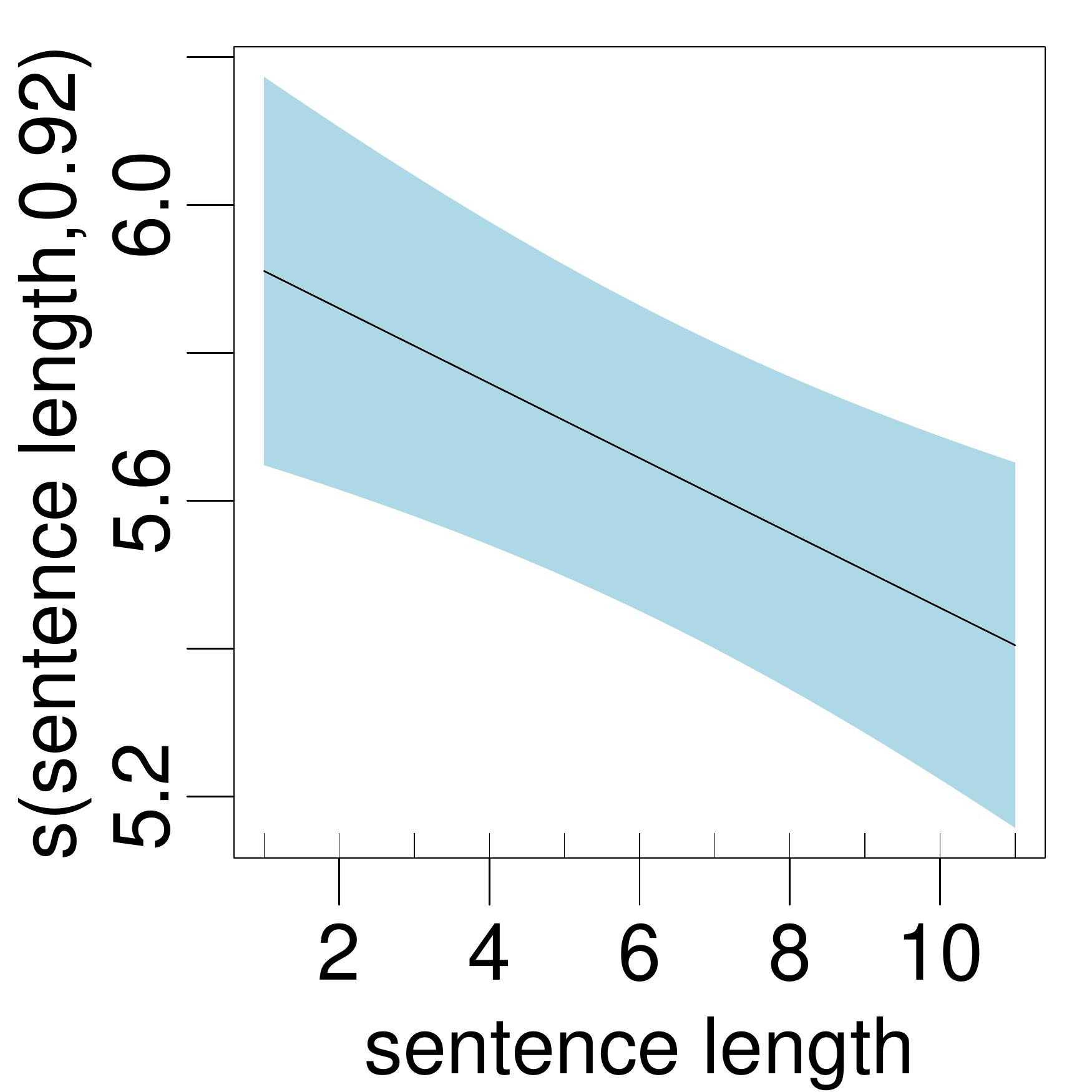}
    \caption{Sentence length}\label{fig:partial_effects_sentence_length}
    \end{subfigure}%
    \hspace{1cm}
    \begin{subfigure}[t]{.22\textwidth}
        \centering
    \includegraphics[width=\textwidth]{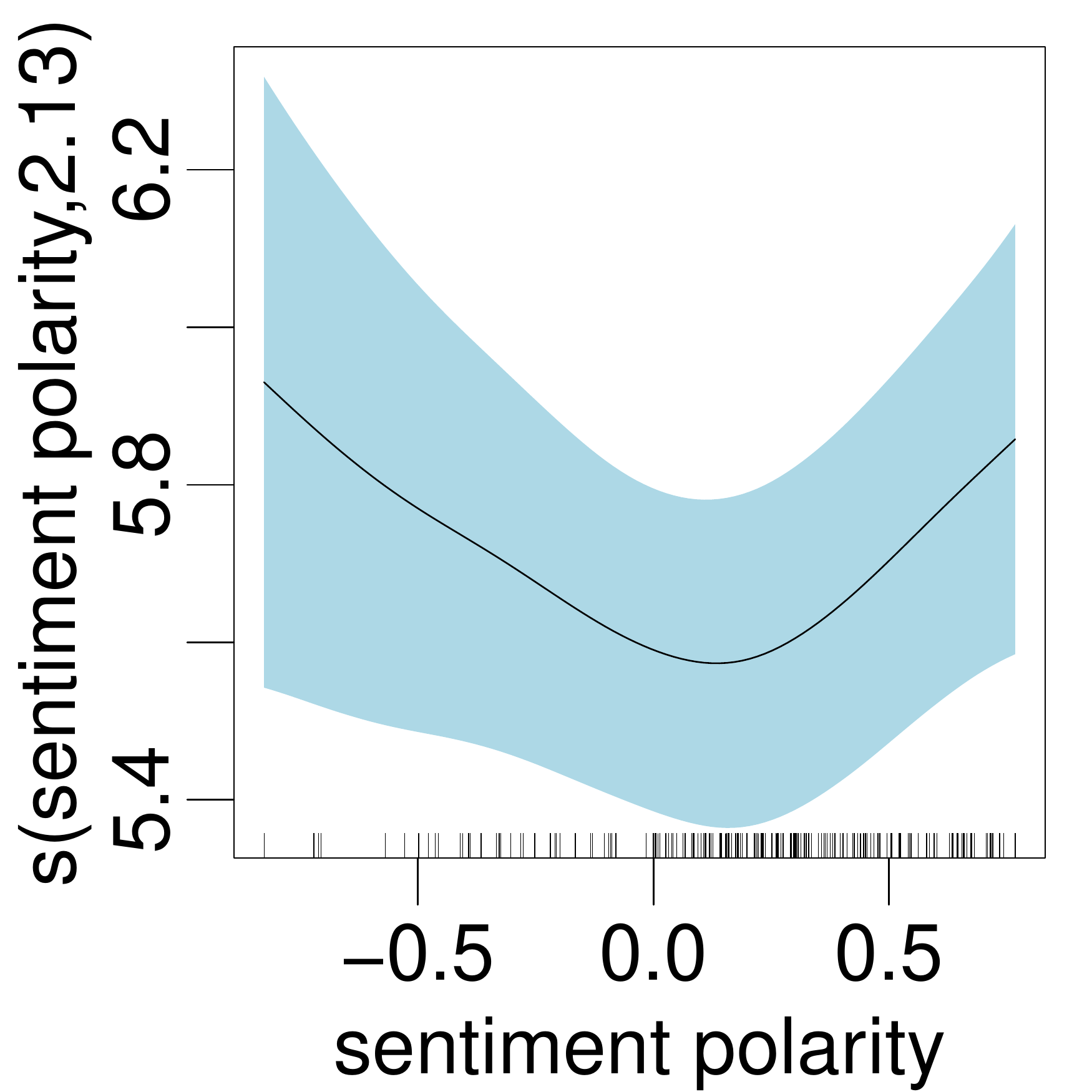}
    \caption{Word sentiment}\label{fig:partial_effects_polarity}
    \end{subfigure}%
    \hspace{1cm}
    \begin{subfigure}[t]{.22\textwidth}
        \centering
    \includegraphics[width=\textwidth]{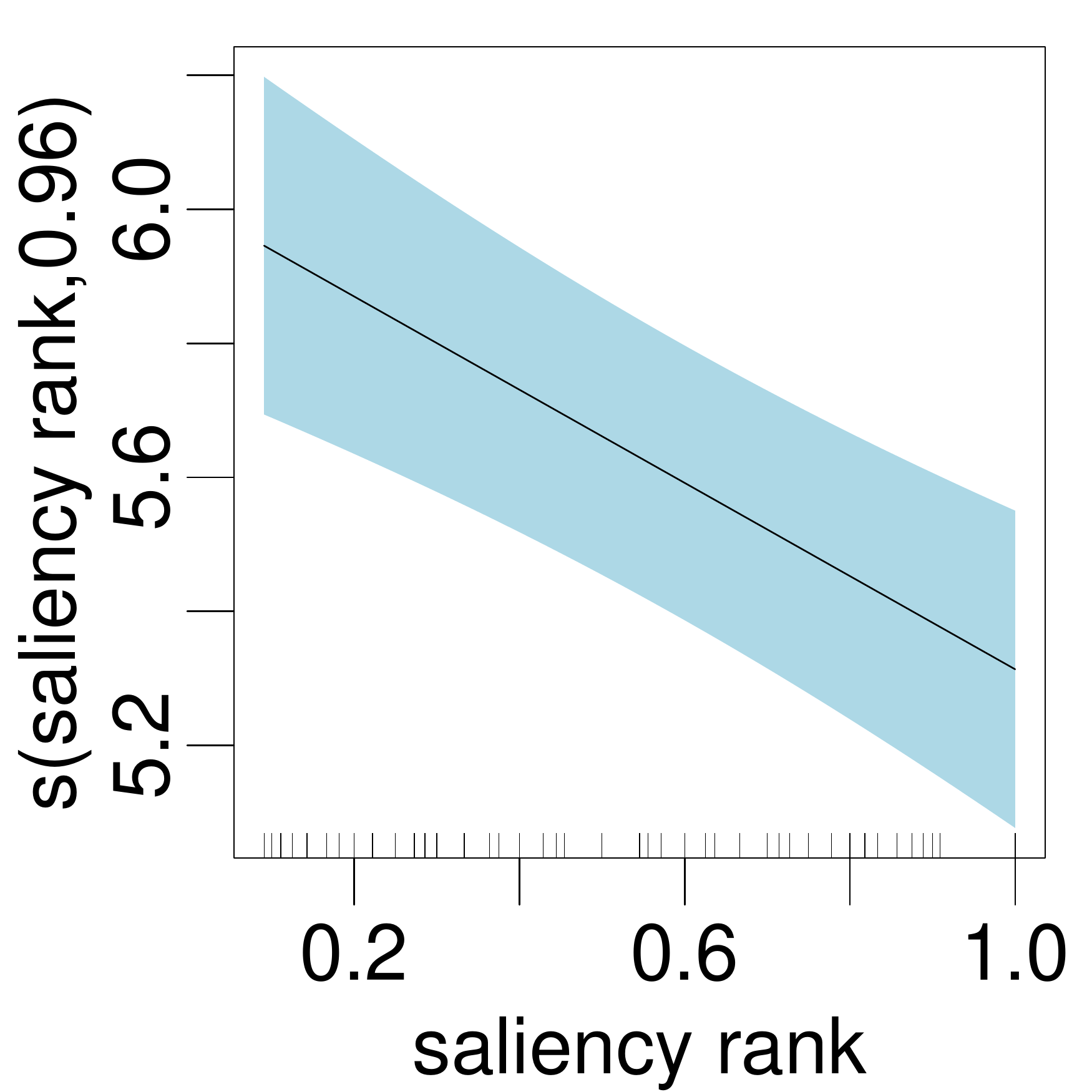}
    \caption{Saliency rank}\label{fig:partial_effects_saliency_rank}
    \end{subfigure}%
    \caption{Partial effect plots for all significant smooth terms (note that y-axes are scaled per effect). Numbers in y-axis labels are estimated degrees of freedom (edf) of the respective smooth. The shaded area displays confidence intervals (plus and minus one standard error) including uncertainty about the overall mean.}\label{fig:partial_effects}
\end{figure*}

Table~\ref{tab:tests_smooth_univariate} shows statistics for the univariate smooth terms in the fitted GAMM.
Figure~\ref{fig:partial_effects} shows partial effect plots of the respective significant smooth terms.
\begin{table}[t]
\centering
\ra{1.3}
\caption{Effective degrees of freedom (edf), reference df and Wald test statistics for the uniariate smooth terms of the first user study.}\label{tab:tests_smooth_univariate}%
\resizebox{1\linewidth}{!}{
\begin{tabular}{lrrrr}
  \hline
 & \textbf{edf} & \textbf{ref. df} & \textbf{F} & \textbf{p} \\ 
  \hline
\rowcolor{lightgray}s(saliency) & 12.0967 & 19 & 728.8738 & $<$ 0.0001 \\ 
    s(display index) & 1.0921 & 9 & 2.0872 & 0.0001 \\ 
\rowcolor{lightgray}s(word length) & 2.5416 & 9 & 4.1826 & $<$ 0.0001 \\ 
    s(sentence length) & 0.9200 & 9 & 1.7531 & 0.0001 \\ 
\rowcolor{lightgray} s(word frequency) & 0.0011 & 9 & 0.0001 & 0.1082 \\ 
    s(sentiment polarity) & 2.1281 & 9 & 1.6156 & 0.0065 \\ 
\rowcolor{lightgray}s(saliency rank) & 0.9580 & 9 & 4.4417 & $<$ 0.0001 \\ 
    s(word position) & 0.0005 & 9 & 0.0000 & 0.7882 \\ 
    \hline
\end{tabular}}
\end{table}

Regarding the parametric terms, neither a word's capitalization (df=2, F=1.84, p=0.16) nor its dependency relation (df=35, F=1.17, p=0.24) show a significant effect on perceived importance.
Regarding the smooth terms, we observe that saliency score, display index, word length, sentence length, word sentiment polarity and saliency rank show significant effects on perceived importance.
In the following, we discuss each effect in detail.
\\[0.4em] 
\noindent \textit{Saliency (Figure~\ref{fig:partial_effects_saliency_score}):} \ 
 The saliency (i.e., the color saturation) has the strongest impact on perceived importance as the graph spans the by-far widest y-axis range of all plots in Figure~\ref{fig:partial_effects}.
Except for the saliency scores around 1, the entire graph shows a monotonous relation between saliency score and perceived importance.
\\[0.4em]  
\noindent \textit{Display Index (Figure~\ref{fig:partial_effects_display_index}):} \
 Participants' ratings increased over the course of the experiment.
We hypothesize that the participants report more conservative ratings in the beginning of the experiment to ``leave enough room'' for more extreme sentences and adapt their ratings to a more ``calibrated'' level over the course of the experiment.
Interestingly, this trend does not seem to stop after our maximum number of 150 sentences. We leave the study of sufficient amount of training required for the effect to reach a peak to future work.
\\[0.4em]  
\noindent \textit{Word Length (Figure~\ref{fig:partial_effects_word_length}):} \
 With increasing word length, importance ratings rise up until a length of approximately eight characters and decrease again afterwards.
We hypothesize that the initial increase corresponds to an increase of the colored area that a longer word directly causes, as the saliency score is visualized within a box which is proportional to the number of characters.
To interpret the subsequent decrease of perceived importance, we consider the interactions between word length and other factors.
We find significant pairwise interactions of word length with (i) saliency, (ii) display index and (iii) word frequency (Appendix~\ref{sec:appendix_en}).
For the interaction with display index, we observe that the decreasing effect of high word lengths grows with increasing display index up until around the 55th sentence.
After this point, the effect decreases.
While the latter decrease can be explained with the partial effect of increasing ratings with higher display indices (as shown in Figure~\ref{fig:partial_effects_display_index}), the former decrease demands detailed investigation in future work.
\\[0.4em] 
\noindent \textit{Sentence Length (Figure~\ref{fig:partial_effects_sentence_length}):} \ 
Importance ratings decrease for words in longer sentences.
A longer sentence leads to a higher number of color samples and therefore also to a larger expected color range.
We argue that such an increased color range inhibts users to make very high importance ratings due to a missing ``maximum color'' anchor.
\\[0.4em] 
\noindent \textit{Sentiment Polarity (Figure~\ref{fig:partial_effects_polarity}):} \ 
 The effect of a word's lemma's sentiment polarity on importance ratings.
We observe a parabola-shaped curve with a minimum at slightly-positive sentiment.
To the left, importance ratings increase with increasingly negative polarity and to the right importance ratings increase with increasingly positive polarity.
This indicates that users ratings of ``what was important to the model when classifying the sentence'' are biased by their answer to ``what is important to me when classifying the sentence myself''.
Such a substitution of a presumably complex-to-compute target attribute with a simpler heuristic attribute is a known cognitive bias and often referred to as \textit{attribute substitution} or \textit{substitution bias} \cite{kahneman_representativeness_2002}
\\[0.4em] 
\noindent \textit{Saliency Rank (Figure~\ref{fig:partial_effects_saliency_rank}):} \
 The partial effect of a word's normalized saliency rank on participants' importance ratings.
We normalize the rank by dividing by sentence length, as low ranks (i.e., larger numbers) would otherwise be strongly correlated to sentence length, and potentially cause stability issues within the model estimation.
We observe that an increased rank (a value of one corresponds to the last rank, i.e., the lowest saliency score) corresponds to a decrease in rated importance.
In contrast to the effect of saliency score shown in Figure~\ref{fig:partial_effects_saliency_score}, the saliency rank is not only a property of a word but of a word in context of its sentence.
A word's saliency score can remain unchanged while at the same time its rank can be arbitrarily modified by changing the saliency scores of the other words in its sentence.
We argue that the significant effect of saliency rank indicates that users interpret saliencies \textit{in relation} to each other, i.e., their judgements are relative and lack a fixed anchoring point. This is supported by qualitative analysis in Section~\ref{sec:study_en_qualitative}.

In addition to the significant partial effects, we also find numerous significant interactions.
We provide the statistics of Wald tests for
all pairwise tensor product interactions (following a functional ANOVA decomposition) as well as summed effect plots of all significant pairwise interactions in Table~\ref{tab:tests_smooth_interactions} and Figure~\ref{fig:summed_effects} in Appendix~\ref{sec:appendix_en}.

\subsubsection{Qualitative}\label{sec:study_en_qualitative}
In addition to the statistical evaluation, we also evaluate the participants' voluntary free-text comments.
Table~\ref{tab:participant_comments} shows a selection of comments grouped into four categories:
\\[0.4em] 
\noindent \textit{Relative Judgement:} \
Participants explicitly state that they make relative importance judgements.
This supports our argumentation of relative judgments discussed for the effects of sentence length and saliency rank.
\\[0.4em] 
\noindent \textit{Own Opinion:} \
Similarly, participant comments support our hypothesis that users' ratings are subject to the cognitive bias of attribute substitution as discussed for the effect of word sentiment polarity.
\\[0.4em] 
\noindent \textit{Light Color:} \
Participants seem to make a categorial distinction between \textit{very light color} and \textit{seemingly no color} although this distinction does not exist in terms of the attribution score.
This can be important when communicating very low influences and should be addressed in more detail in future work.
\\[0.4em] 
\noindent \textit{Other:} \
Miscellaneous comments on, e.g., issues of word-level attribution and the resulting ambiguity in interpretation.

\definecolor{c_255_197_197}{RGB}{255,197,197}
\definecolor{c_255_74_74}{RGB}{255,74,74}
\definecolor{c_255_62_62}{RGB}{255,62,62}
\definecolor{c_255_127_127}{RGB}{255,127,127}
\definecolor{c_255_1_1}{RGB}{255,1,1}
\definecolor{c_255_77_77}{RGB}{255,77,77}
\definecolor{c_255_231_231}{RGB}{255,231,231}
\definecolor{c_255_107_107}{RGB}{255,107,107}
\definecolor{c_255_113_113}{RGB}{255,113,113}
\definecolor{c_255_167_167}{RGB}{255,167,167}
\definecolor{c_255_220_220}{RGB}{255,220,220}
\definecolor{c_255_51_51}{RGB}{255,51,51}
\definecolor{c_255_136_136}{RGB}{255,136,136}
\definecolor{c_255_232_232}{RGB}{255,232,232}
\definecolor{c_255_46_46}{RGB}{255,46,46}
\definecolor{c_255_227_227}{RGB}{255,227,227}
\definecolor{c_255_222_222}{RGB}{255,222,222}
\definecolor{c_255_213_213}{RGB}{255,213,213}
\definecolor{c_255_81_81}{RGB}{255,81,81}
\definecolor{c_255_184_184}{RGB}{255,184,184}
\definecolor{c_255_19_19}{RGB}{255,19,19}
\definecolor{c_255_138_138}{RGB}{255,138,138}
\definecolor{c_255_240_240}{RGB}{255,240,240}
\definecolor{c_255_249_249}{RGB}{255,249,249}
\definecolor{c_255_91_91}{RGB}{255,91,91}
\definecolor{c_255_103_103}{RGB}{255,103,103}
\definecolor{c_255_218_218}{RGB}{255,218,218}
\definecolor{c_255_168_168}{RGB}{255,168,168}
\definecolor{c_255_133_133}{RGB}{255,133,133}
\definecolor{c_255_65_65}{RGB}{255,65,65}
\definecolor{c_255_185_185}{RGB}{255,185,185}
\definecolor{c_255_129_129}{RGB}{255,129,129}
\definecolor{c_255_99_99}{RGB}{255,99,99}
\definecolor{c_255_127_127}{RGB}{255,127,127}
\definecolor{c_255_160_160}{RGB}{255,160,160}
\definecolor{c_255_21_21}{RGB}{255,21,21}
\definecolor{c_255_22_22}{RGB}{255,22,22}
\definecolor{c_255_63_63}{RGB}{255,63,63}
\definecolor{c_255_202_202}{RGB}{255,202,202}
\definecolor{c_255_175_175}{RGB}{255,175,175}
\definecolor{c_255_195_195}{RGB}{255,195,195}
\definecolor{c_255_139_139}{RGB}{255,139,139}
\definecolor{c_255_244_244}{RGB}{255,244,244}
\definecolor{c_255_127_127}{RGB}{255,127,127}
\definecolor{c_255_177_177}{RGB}{255,177,177}
\definecolor{c_255_18_18}{RGB}{255,18,18}

\definecolor{c_255_216_216}{RGB}{255,216,216}
\definecolor{c_255_152_152}{RGB}{255,152,152}
\definecolor{c_255_189_189}{RGB}{255,189,189}
\definecolor{c_255_239_239}{RGB}{255,239,239}
\definecolor{c_255_70_70}{RGB}{255,70,70}

\definecolor{c_255_227_227}{RGB}{255,227,227}
\definecolor{c_255_76_76}{RGB}{255,76,76}
\definecolor{c_255_234_234}{RGB}{255,234,234}
\definecolor{c_255_85_85}{RGB}{255,85,85}
\definecolor{c_255_30_30}{RGB}{255,30,30}
\definecolor{c_255_189_189}{RGB}{255,189,189}
\definecolor{c_255_249_249}{RGB}{255,249,249}
\definecolor{c_255_60_60}{RGB}{255,60,60}
\definecolor{c_255_241_241}{RGB}{255,241,241}
\definecolor{c_255_220_220}{RGB}{255,220,220}
\definecolor{c_255_96_96}{RGB}{255,96,96}
\definecolor{c_255_177_177}{RGB}{255,177,177}
\definecolor{c_255_83_83}{RGB}{255,83,83}
\definecolor{c_255_35_35}{RGB}{255,35,35}
\definecolor{c_255_118_118}{RGB}{255,118,118}
\definecolor{c_255_130_130}{RGB}{255,130,130}
\definecolor{c_255_253_253}{RGB}{255,253,253}
\definecolor{c_255_249_249}{RGB}{255,249,249}
\definecolor{c_255_185_185}{RGB}{255,185,185}
\definecolor{c_255_253_253}{RGB}{255,253,253}
\definecolor{c_255_177_177}{RGB}{255,177,177}
\definecolor{c_255_221_221}{RGB}{255,221,221}
\definecolor{c_255_63_63}{RGB}{255,63,63}
\definecolor{c_255_102_102}{RGB}{255,102,102}
\definecolor{c_255_89_89}{RGB}{255,89,89}

\definecolor{c_255_133_133}{RGB}{255,133,133}
\definecolor{c_255_65_65}{RGB}{255,65,65}
\definecolor{c_255_185_185}{RGB}{255,185,185}
\definecolor{c_255_129_129}{RGB}{255,129,129}
\definecolor{c_255_99_99}{RGB}{255,99,99}
\definecolor{c_255_127_127}{RGB}{255,127,127}
\definecolor{c_255_160_160}{RGB}{255,160,160}
\definecolor{c_255_21_21}{RGB}{255,21,21}
\definecolor{c_255_22_22}{RGB}{255,22,22}
\definecolor{c_255_63_63}{RGB}{255,63,63}
\definecolor{c_255_202_202}{RGB}{255,202,202}
\definecolor{c_255_210_210}{RGB}{255,210,210}
\definecolor{c_255_14_14}{RGB}{255,14,14}
\definecolor{c_255_20_20}{RGB}{255,20,20}
\definecolor{c_255_25_25}{RGB}{255,25,25}
\definecolor{c_255_227_227}{RGB}{255,227,227}
\definecolor{c_255_184_184}{RGB}{255,184,184}
\definecolor{c_255_3_3}{RGB}{255,3,3}
\definecolor{c_255_3_3}{RGB}{255,3,3}
\definecolor{c_255_51_51}{RGB}{255,51,51}
\definecolor{c_255_72_72}{RGB}{255,72,72}
\definecolor{c_255_25_25}{RGB}{255,25,25}
\definecolor{c_255_28_28}{RGB}{255,28,28}
\definecolor{c_255_229_229}{RGB}{255,229,229}

\begin{table*}
\centering

\ra{1.3} 
\caption{Comments of the participants of the English sentiment study. Participants were asked to rate the \underline{underlined} word or symbol.}\label{tab:participant_comments}

\includegraphics{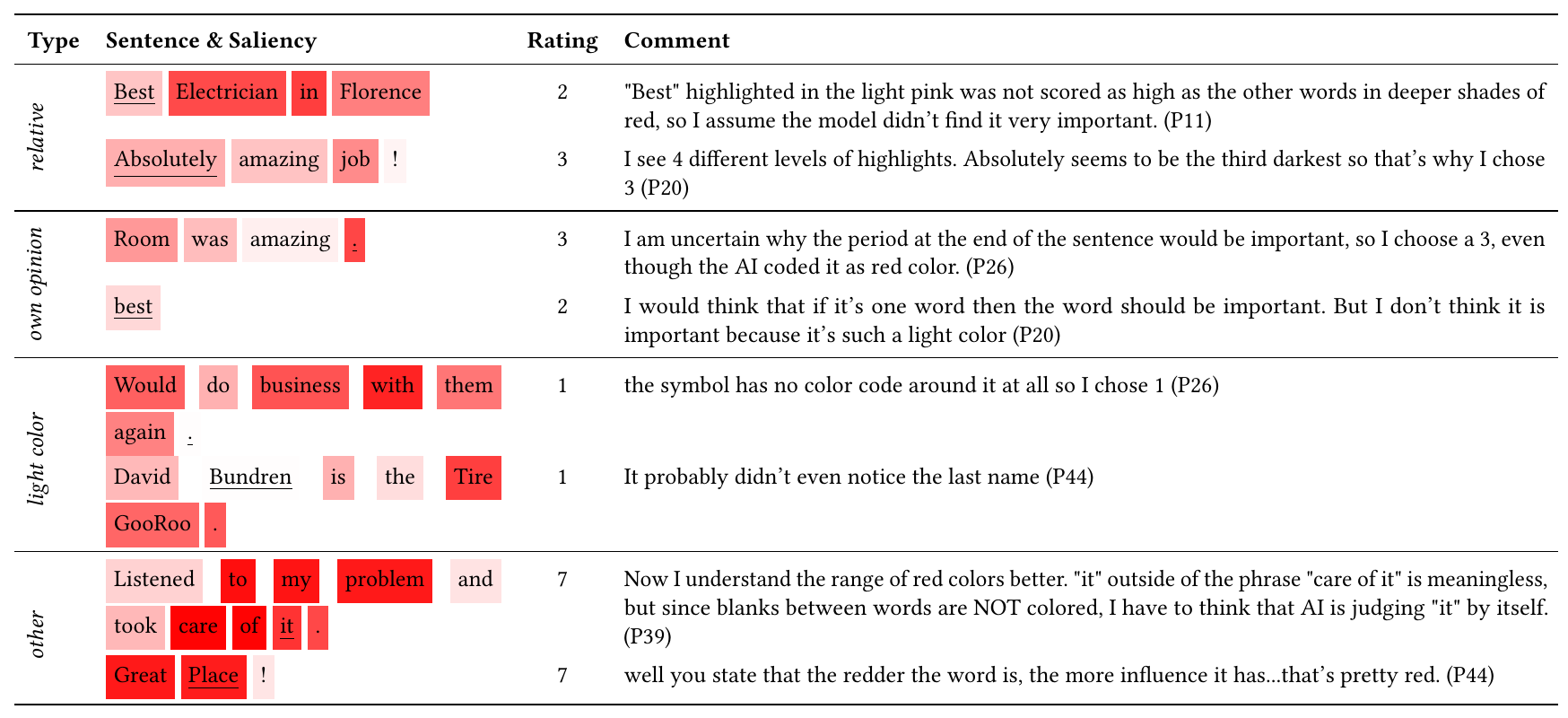}
\end{table*}

\subsection{Generalization Across Tasks and Languages: Fact Checking in German}\label{sec:study_2}
So far, we found indication that numerous factors (word length, saliency rank, etc.) significantly influence users' subjective importance ratings.
Two important limitations are that (i) the findings are limited to English, and (ii) they are limited to one AI task (sentiment classification).
To assess whether the findings do generalize to another language and another task, we repeat the study identically with German sentences from the
PUD Corpus\footnote{\url{https://universaldependencies.org/treebanks/de_pud/index.html}.} with a fact checking AI task.
We collect responses from 25 German-speaking participants from a participant pool including Germany, Austria and Switzerland.
In total, this corresponds to 3750 ratings.

\subsubsection{Confirmed Effects}
Our analysis confirms the significants effects of saliency, display index, word length and sentence length.
Figure~\ref{fig:partial_effects_de} displays the respective partial effect plots.
While the smooths for saliency (Figure~\ref{fig:partial_effects_saliency_score_de}) and sentence length (Figure~\ref{fig:partial_effects_sentence_length_de}) show high similarity to the respective smooths of the English study (see Figures~\ref{fig:partial_effects_saliency_score} and \ref{fig:partial_effects_sentence_length}), we observe slight differences for display index (Figures~\ref{fig:partial_effects_display_index_de} and \ref{fig:partial_effects_display_index}) and word length (Figures~\ref{fig:partial_effects_word_length_de} and \ref{fig:partial_effects_word_length}).
While the English display index smooth grows more or less linearly (edf=1.09), the respective German smooth reaches a plateau after around half the sentences (edf=1.60).
We hypothesize that such a saturation effect will also be visible for English, but requires a larger number of sentences.
We argue that this is caused by the fact that the sentences in the German study are longer than in the English study, which makes participants of the German study see more colored words and thereby ``calibrates'' their ratings faster in terms of number of sentences. 
Similarly, the German word length smooth saturates after around 15 characters, while the English smooth decreases after around 8 characters.
We hypothesize that this difference can be attributed to the overall longer words in German as well as the difference in compounding.

The effect of saliency rank cannot be confirmed in the German experiment.
Like in the English study, we find no indication that word frequency has a significant effect on importance ratings.
We provide test statistics of paramatric and smooth terms (univariate smooths and pairwise interactions) as well as coefficient estimates in Appendix~\ref{sec:appendix_de}. 

As in the English study, we additionally qualitatively analyse the participants' free text comments and observe (as in the English study) numerous instances for which participants mix their own estimate of importance with the communicated importance.
We provide exemplary instances in Table~\ref{tab:participant_comments_de} in Appendix~\ref{sec:appendix_de}.

\subsubsection{Additional Effects}
\begin{figure*}
    \centering
    \begin{subfigure}[t]{.195\textwidth}
        \centering
    \includegraphics[width=\textwidth]{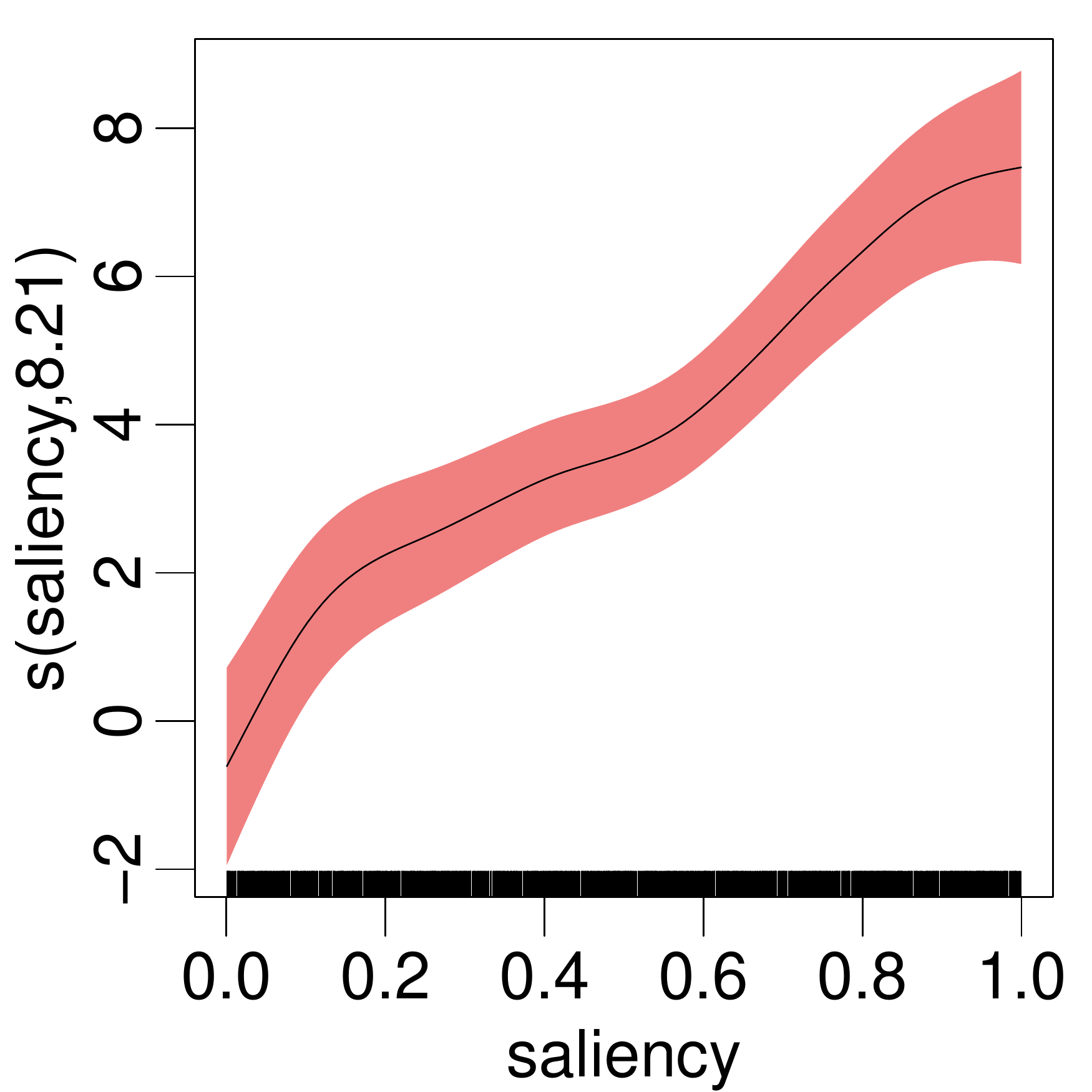}
    \caption{Saliency (saturation)}\label{fig:partial_effects_saliency_score_de}
    \end{subfigure}%
    \hfill
    \begin{subfigure}[t]{.195\textwidth}
        \centering
    \includegraphics[width=\textwidth]{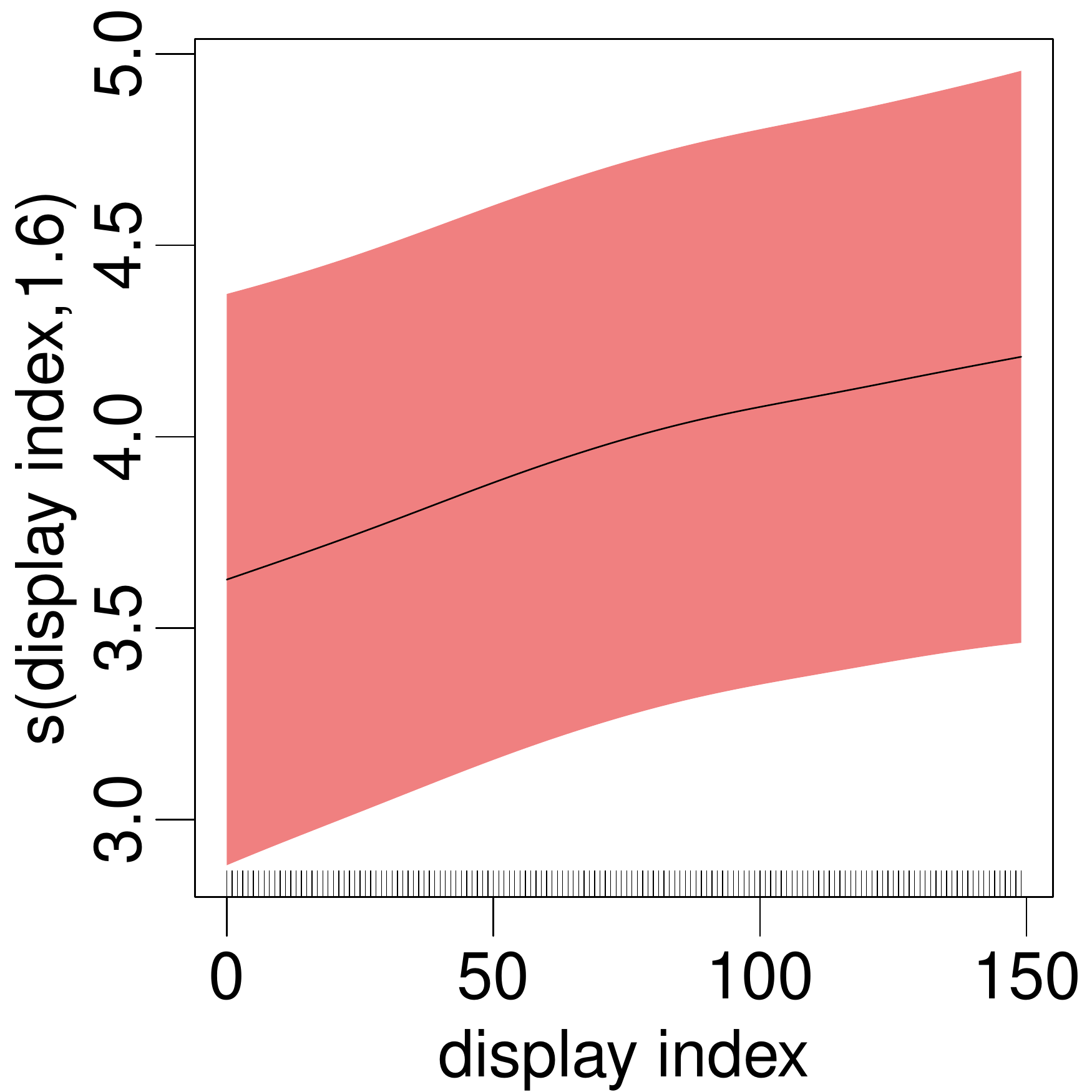}
    \caption{Temporal display index}\label{fig:partial_effects_display_index_de}
    \end{subfigure}%
    \hfill
    \begin{subfigure}[t]{.195\textwidth}
        \centering
    \includegraphics[width=\textwidth]{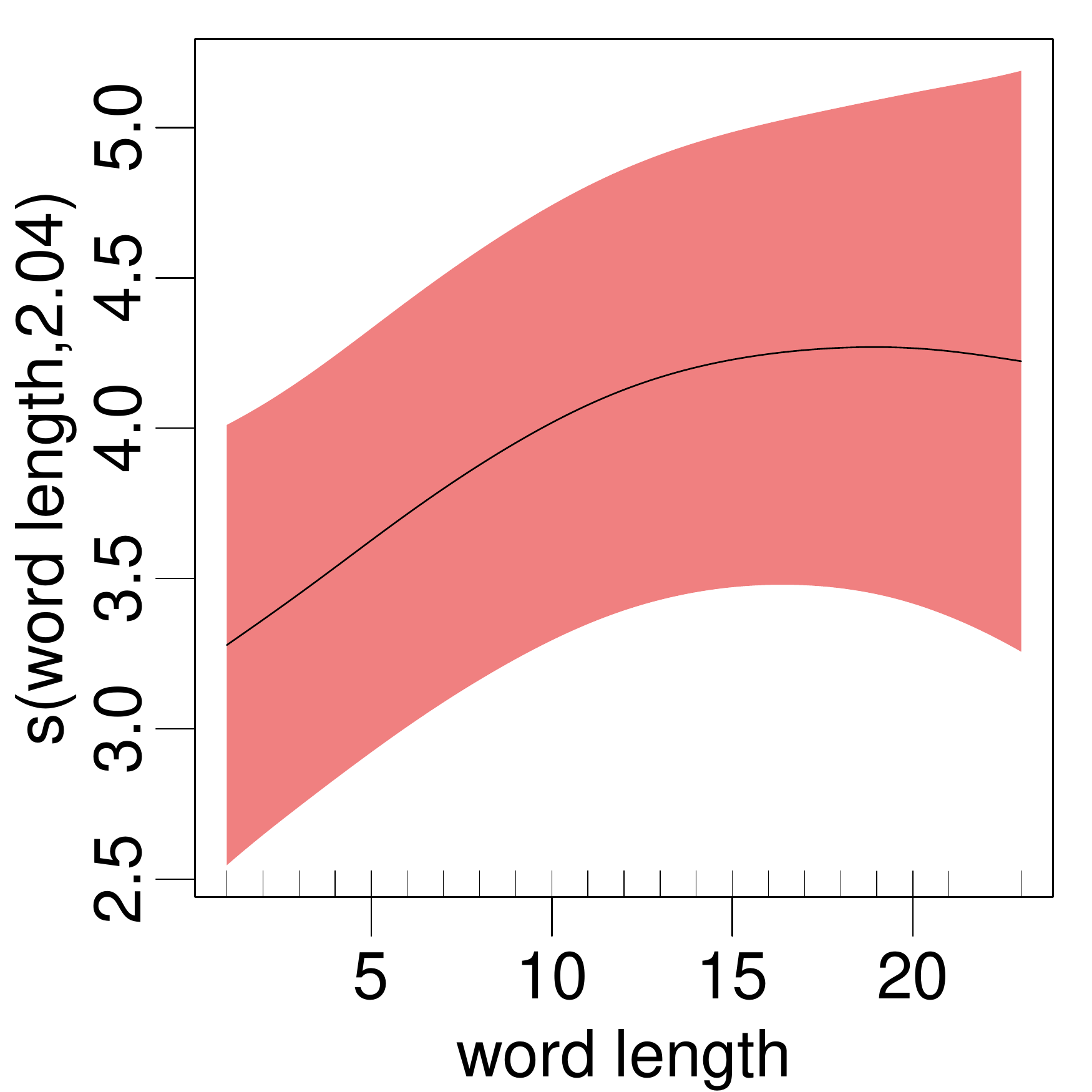}
    \caption{Word length}\label{fig:partial_effects_word_length_de}
    \end{subfigure}
    \begin{subfigure}[t]{.195\textwidth}
        \centering
    \includegraphics[width=\textwidth]{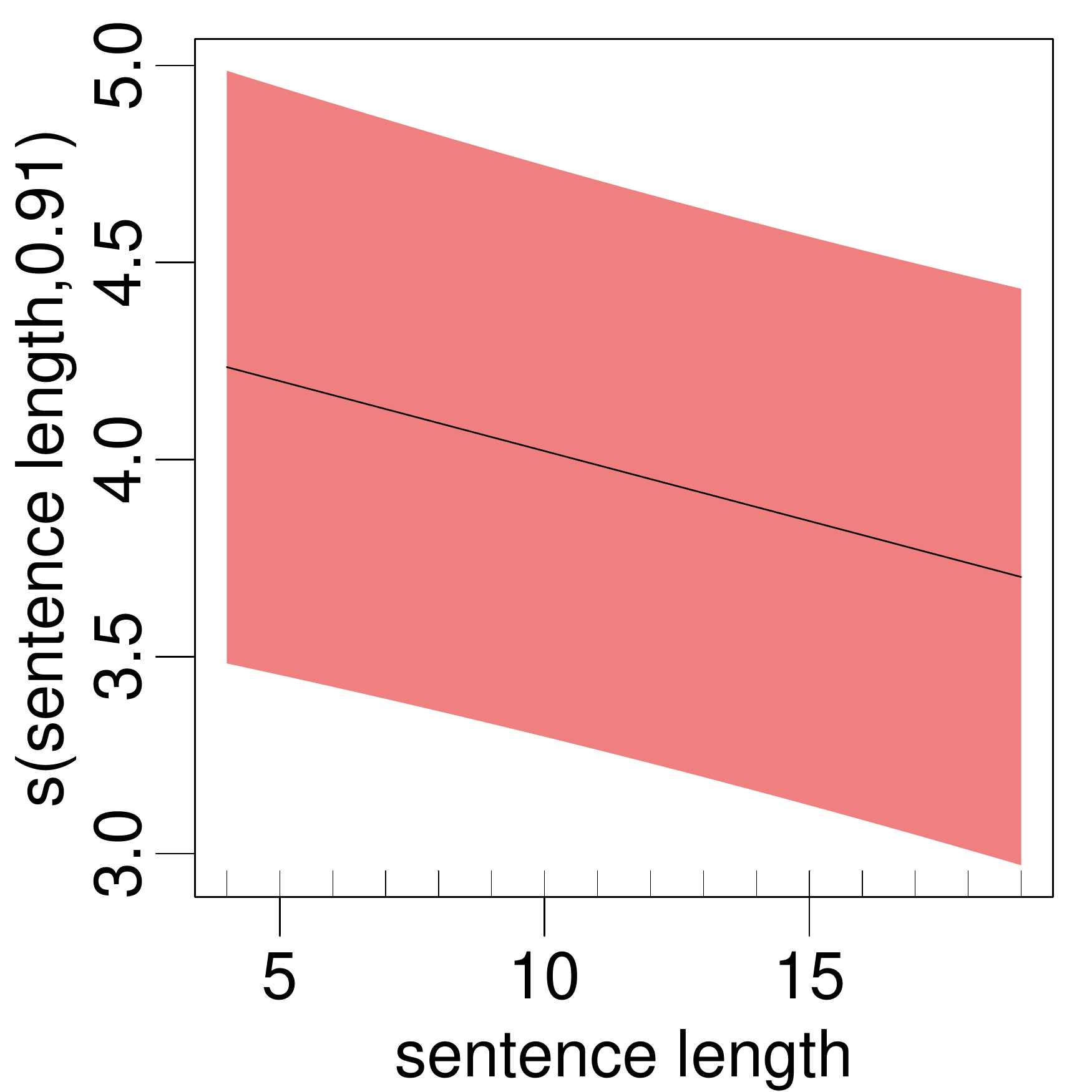}
    \caption{Sentence length}\label{fig:partial_effects_sentence_length_de}
    \end{subfigure}%
    \begin{subfigure}[t]{.195\textwidth}
        \centering
    \includegraphics[width=\textwidth]{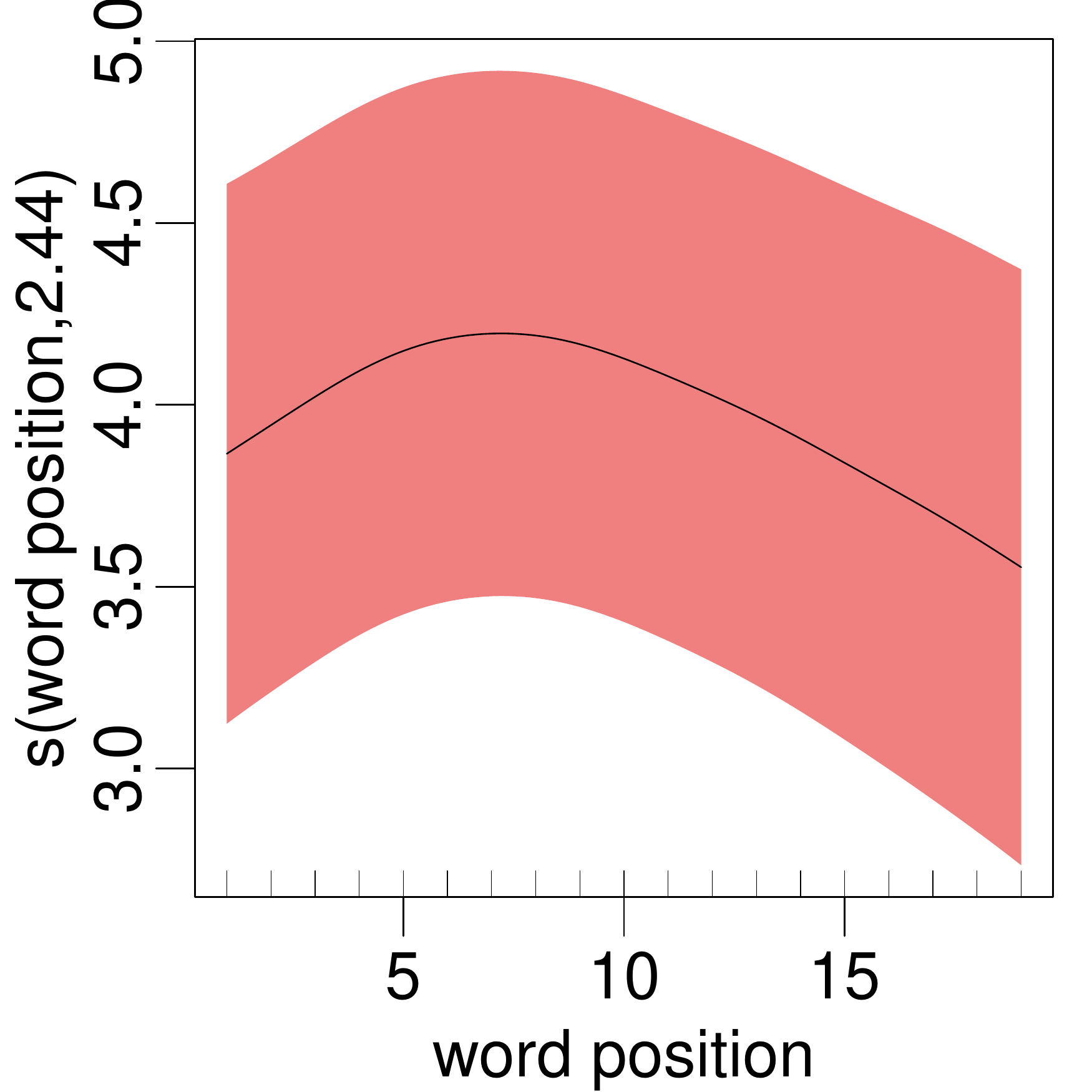}
    \caption{Word position}\label{fig:word_index_in_sentence_de}
    \end{subfigure}%
    \caption{Partial effect plots for all significant smooth terms (note that y-axes are scaled per effect) for the German experiment. Numbers in y-axis labels are estimated degrees of freedom (edf) of the respective smooth. The shaded area displays confidence intervals (plus and minus one standard error) including uncertainty about the overall mean. (a) refers to color saturation.}\label{fig:partial_effects_de}
\end{figure*}

In addition to the effects that we already observed in the English study, we also find that the word's position within the sentence (Figure~\ref{fig:word_index_in_sentence_de}) as well as
capitalization and
dependency relation have significant effects on importance ratings.
A full list of coefficient estimates along with further details is provided in Table~\ref{tab:coefficients_details_de} in Appendix~\ref{sec:appendix_de}.

The estimate for fully capitalized words is 1.91 (SE=0.9638), the respective estimate for words with the first letter capitalized is 0.41 (SE=0.12).\footnote{The estimate for lower-cased words is fixed to zero as the reference level. For dependency relations, we choose the (most frequent) punctuation relation.}
This confirms the intuition that fully capitalized words receive the highest importance ratings, followed by first-letter-capitalized words.
We argue that this effect---in particular for first-letter-capitalized words---is more visible in the German experiment as German uses more frequent capitalization (e.g., for all nouns).

Regarding dependency relations, the highest estimate can be observed for temporal modifiers (obl:tmod, $\beta=1.70$, SE=0.55) like ``today'' and numerical modifiers (nummod, $\beta=1.39$, SE=0.36) like ``one''.
The lowest estimate can be observed for clausal modifier of nouns (acl, $\beta=-1.22$, SE=0.64) like ``sees'' in ``the issues as he sees them'' and indirect objects (iobj, $\beta=-0.48$, SE=0.52) like ``me'' in ``she gave me the book''.  We hypothesize that the grammatical function effect is larger here than in the previous experiment because properties such as the use of temporality, numerals and embedded clauses are more important for determining factuality than for determining sentiment.

\subsection{Generalization to Model-based Saliencies (derived via Integrated Gradients)}\label{sec:study-ig}
We want to assess whether our findings on the random saliency scores used in the previous two studies also hold for practically-used feature attribution scores.
Therefore, we conduct an additional user study
using integrated gradients \cite{sundararajan_axiomatic_2017} instead of random saliencies.\footnote{We make use of the Language Interpretability Toolkit \cite{tenney2020language} to obtain normalized integrated gradient scores with respect to the SST2-base sentiment model and 30 interpolation steps.}

\subsubsection{Study Modification: Within-Subject Design}
We combine the evaluation of integrated gradient scores with a within-subject evaluation of three visualization methods which we detail in Section~\ref{sec:bias_mitigation}.
In this section, we focus on the unmodified visualization as it is used in the two previously described studies.
In the remainder of this paper, this visualization method is referred to as \textit{saliency}.
We sample another 150 sentiment sentences from the sentence pool described in Section~\ref{subsec:collection-methodology} and present them in the same sentiment classification context.
Instead of using one saliency visualization method for all 150 sentences, we now use the three visualizations and show each participant 50 sentences per visualization.\footnote{The order of visualization methods is balanced across participants.
Sentence order is fixed to ensure identical ordering effects for the three visualizations.
}
We collect 9000 importance ratings from 60 participants and exclude participants of the previous study to avoid carry-over effects from previous exposures.

\subsubsection{Model Modification: Factor-Smooth Interactions}
We again use an ordinal GAMM model using the same covariates as
described in Section~\ref{subsec:analysis-methodology}.
In addition, we add a parametric term for the visualization condition to account for overall differences in rating intensities between the visualization conditions.\footnote{We additionally include a random intercept to account for visualization order.}
We use factor-smooth interactions for each variable which leads to separate estimates for each variable per visualization (e.g., three smooths for word length, one per visualization).
First, this yields smooths for the ``orginal'' saliency visualization, i.e., the heatmap visualization without saliency corrections.
In contrast to our first study, these smooths now correspond to effects on integrated gradient attribution scores instead of random scores.
First, comparing the smooths allows us to compare how factors influence importance ratings across the three visualizations, e.g., to assess whether the bar visualization did mitigate the biasing effect of word length.
We discuss the respective results in Section~\ref{sec:empirical_evaluation_correction}.
Second, analyzing the smooths relating to the original saliency visualization allows us to evaluate which of the effects we observed in the first study do generalize to the integrated gradients attribution scores.
We discuss the respective results in the following paragraph.

\subsubsection{Results}
We find significant effects of saliency score, word length, relative word frequency and saliency rank.
We provide details and test statistics on all parametric coefficients as well as smooth terms in Table~\ref{tab:model_params_corr_all} in Appendix~\ref{sec:appendix_IG}.
All of these variables except relative word frequency were also found to be significant in our first study and all of them except relative word frequency and saliency rank were confirmed in our German study.
The significant influence of relative word frequency was observed for the first time.

\noindent\textbf{Overall}, three studies confirmed the presumably biasing influence of word length, (pairs of) two studies respectively confirmed the effect of sentence length, display index and saliency rank, and one study (each) found significant effects of word position, sentiment polarity, word frequency, capitalization and dependency relation. Together, these reflect the three sources of bias: anthropomorphism and belief bias, visual perception, and learning effects.

\section{Mitigating Visual Perception and Learning Effect Biases}\label{sec:bias_mitigation}

\begin{figure*}
    \centering
    \begin{subfigure}[t]{.25\textwidth}
        \centering
    \includegraphics[width=\textwidth]{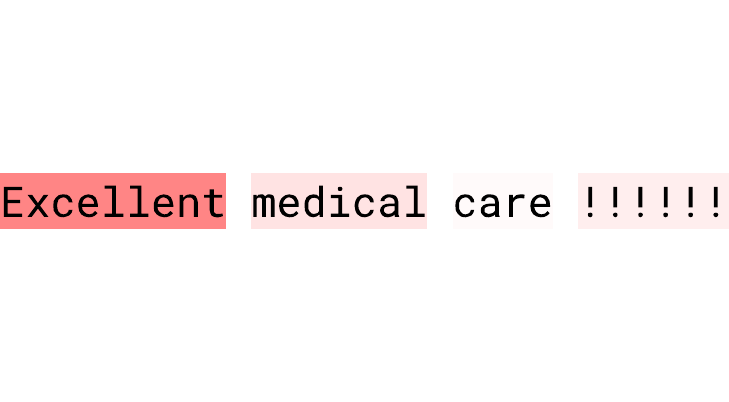}
    \caption{Original saliency}\label{fig:saliency_viz}
    \end{subfigure}%
    \hfill
    \begin{subfigure}[t]{.25\textwidth}
        \centering
    \includegraphics[width=\textwidth]{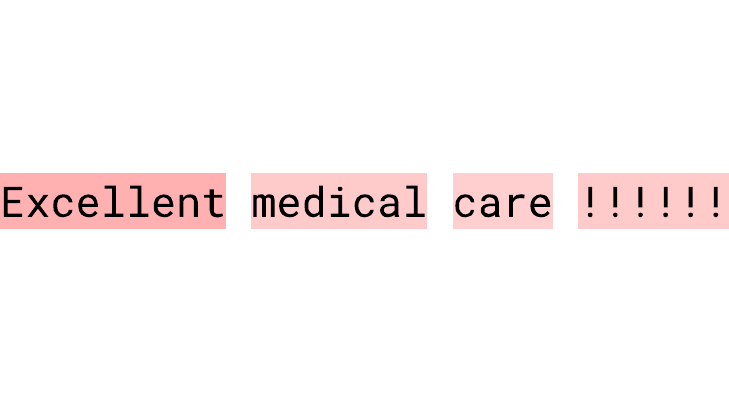}
    \caption{Corrected saliency}\label{fig:corrected_saliency_viz}
    \end{subfigure}
    \hfill
    \begin{subfigure}[t]{.25\textwidth}
        \centering
    \includegraphics[width=\textwidth]{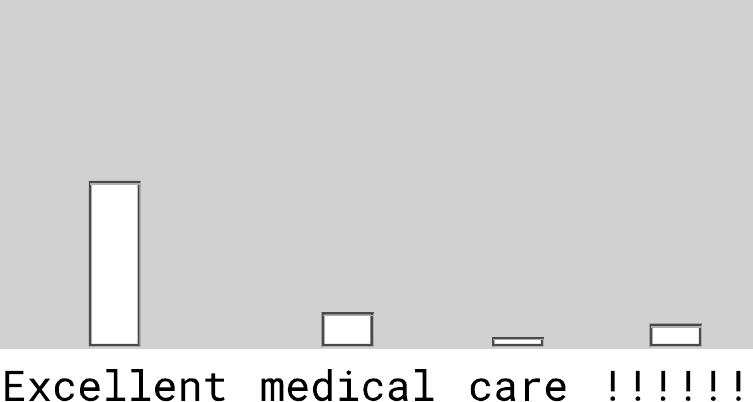}
    \caption{Bars}\label{fig:bar_viz}
    \end{subfigure}%
    \caption{The three different visualization methods we compare in our third study (Section~\ref{sec:bias_mitigation}). 
    }\label{fig:visualization-types}
\end{figure*}

So far, we observed that various seemingly irrelevant factors influence human perception in unintended ways from the explicit and objective saliency information across different languages, tasks and feature-attribution scores.
Next, we explore two methods to decrease the bias in human perception (Figure~\ref{fig:visualization-types}): (i) controlling for the bias by modifying the color-coding to account for over-estimation and under-estimation of importance (over-estimated tokens will receive decreased color saturation, and vice versa); (ii) replacing the color-coding visualization with bar chart visualization. 

\subsection{Model-Based Color Correction Technique}
We compute an alternative color-coding visualization which a-priori accounts for over-estimation and under-estimation of tokens based on the data collected in the previous experiments. Here we investigate whether it is possible to ``correct'' the explainees' saliency perception by superimposing the initial saliency values with a correction signal.

We require a procedure which increases the saliency scores for words which are predicted to be under-perceived (e.g., short words and words that appear in long sentences) and decrease the saliency scores for words that are predicted to be over-perceived (e.g., words with a high sentiment polarity or words that appear in short sentences). 
Briefly, the trained GAMM model from the English study (Section~\ref{sec:study_en}) allows us to map a combination of a saliency score together with word/sentence properties to a perceived importance score (on a continuous latent scale).
By grounding this prediction of perceived importance to a prediction conditioned on a particularly chosen reference level,
we can iteratively, globally correct the explained scores over the sentence such that the (predicted) perception bias is decreased in each iteration.
Table~\ref{tab:correction_examples_main} displays examples of the application of this correction.
In Appendix~\ref{sec:saliency-correction-algorithm}, we discuss the full algorithm including its components and motivating details and provide an extended list of example applications in Table~\ref{tab:correction_examples_appendix} as well as an example of the gradual correction of one sentence over the course of 100 correction steps in Table~\ref{tab:correction_detailed}.

\subsection{Bar Chart Visualization}\label{sec:bar_charts}

As an alternative to color-coding visualization, we consider bar charts (Figure~\ref{fig:bar_viz}): we investigate whether a sufficiently distinct visualization
will result in different perception.
We hypothesize that this is related to visual perception bias.

We note two visual qualities of bars which differentiate it from color-coding, and therefore make it a relevant alternative visualization candidate: (i) The bars are communicated with objective reference points of 0 and 1 (the top and bottom of the draw area), while the results in Section~\ref{sec:study_en} indicate that participants perceive colored saliency in relation to each other, instead of in reference to 0 and 1 (pure white and pure red, respectively); (ii) The draw area for the bars is separate from the draw area for the input text, in contrast to color-coding, where they occupy the same space.
This means that in color-coding, for example, a word with more characters will receive a larger area of color, in comparison to a shorter word with the same color.
As our studies in Section~\ref{sec:study_all} show, word length influenced explainee perception.
In the bar chart visualization scheme, all words are treated identically within the draw area which communicates importance, and this draw area is disconnected from the text display area.

\subsection{Results}\label{sec:empirical_evaluation_correction}

\definecolor{c_255_88_88}{RGB}{255,88,88}
\definecolor{c_255_249_249}{RGB}{255,249,249}
\definecolor{c_255_172_172}{RGB}{255,172,172}
\definecolor{c_255_135_135}{RGB}{255,135,135}
\definecolor{c_255_237_237}{RGB}{255,237,237}
\definecolor{c_255_135_135}{RGB}{255,135,135}
\definecolor{c_255_0_0}{RGB}{255,0,0}
\definecolor{c_74_74_255}{RGB}{74,74,255}
\definecolor{c_107_107_255}{RGB}{107,107,255}
\definecolor{c_255_225_225}{RGB}{255,225,225}
\definecolor{c_255_254_254}{RGB}{255,254,254}
\definecolor{c_254_254_255}{RGB}{254,254,255}
\definecolor{c_255_29_29}{RGB}{255,29,29}
\definecolor{c_255_242_242}{RGB}{255,242,242}
\definecolor{c_255_237_237}{RGB}{255,237,237}
\definecolor{c_255_98_98}{RGB}{255,98,98}
\definecolor{c_255_236_236}{RGB}{255,236,236}
\definecolor{c_255_189_189}{RGB}{255,189,189}
\definecolor{c_255_0_0}{RGB}{255,0,0}
\definecolor{c_206_206_255}{RGB}{206,206,255}
\definecolor{c_104_104_255}{RGB}{104,104,255}
\definecolor{c_254_254_255}{RGB}{254,254,255}
\definecolor{c_255_254_254}{RGB}{255,254,254}
\definecolor{c_254_254_255}{RGB}{254,254,255}
\definecolor{c_255_103_103}{RGB}{255,103,103}
\definecolor{c_255_229_229}{RGB}{255,229,229}
\definecolor{c_255_176_176}{RGB}{255,176,176}
\definecolor{c_255_140_140}{RGB}{255,140,140}
\definecolor{c_255_196_196}{RGB}{255,196,196}
\definecolor{c_255_140_140}{RGB}{255,140,140}
\definecolor{c_255_0_0}{RGB}{255,0,0}
\definecolor{c_25_25_255}{RGB}{25,25,255}
\definecolor{c_87_87_255}{RGB}{87,87,255}
\definecolor{c_253_253_255}{RGB}{253,253,255}
\definecolor{c_255_254_254}{RGB}{255,254,254}
\definecolor{c_255_130_130}{RGB}{255,130,130}
\definecolor{c_255_243_243}{RGB}{255,243,243}
\definecolor{c_255_252_252}{RGB}{255,252,252}
\definecolor{c_255_25_25}{RGB}{255,25,25}
\definecolor{c_255_242_242}{RGB}{255,242,242}
\definecolor{c_255_198_198}{RGB}{255,198,198}
\definecolor{c_255_238_238}{RGB}{255,238,238}
\definecolor{c_255_96_96}{RGB}{255,96,96}
\definecolor{c_255_198_198}{RGB}{255,198,198}
\definecolor{c_46_46_255}{RGB}{46,46,255}
\definecolor{c_123_123_255}{RGB}{123,123,255}
\definecolor{c_255_0_0}{RGB}{255,0,0}
\definecolor{c_87_87_255}{RGB}{87,87,255}
\definecolor{c_255_203_203}{RGB}{255,203,203}
\definecolor{c_254_254_255}{RGB}{254,254,255}
\definecolor{c_254_254_255}{RGB}{254,254,255}
\definecolor{c_186_186_255}{RGB}{186,186,255}
\definecolor{c_255_241_241}{RGB}{255,241,241}
\definecolor{c_255_132_132}{RGB}{255,132,132}
\definecolor{c_255_247_247}{RGB}{255,247,247}
\definecolor{c_255_250_250}{RGB}{255,250,250}
\definecolor{c_255_230_230}{RGB}{255,230,230}
\definecolor{c_255_183_183}{RGB}{255,183,183}
\definecolor{c_255_241_241}{RGB}{255,241,241}
\definecolor{c_255_231_231}{RGB}{255,231,231}
\definecolor{c_134_134_255}{RGB}{134,134,255}
\definecolor{c_255_117_117}{RGB}{255,117,117}
\definecolor{c_182_182_255}{RGB}{182,182,255}
\definecolor{c_0_0_255}{RGB}{0,0,255}
\definecolor{c_254_254_255}{RGB}{254,254,255}
\definecolor{c_255_254_254}{RGB}{255,254,254}
\definecolor{c_255_254_254}{RGB}{255,254,254}
\definecolor{c_254_254_255}{RGB}{254,254,255}

\begin{table*}
\centering

\caption{Examples of the bias reduction procedure. 
The \textit{saliency} column shows the saliency explanations (how users would see them) before and after the bias correction procedure.
The \textit{bias} column shows the color-coded bias estimates. Predicted over-estimations are in red whereas predicted under-estimations are in blue. More examples in \Cref{tab:correction_examples_appendix} in \Cref{sec:saliency-correction-algorithm}.
}\label{tab:correction_examples_main}
\includegraphics{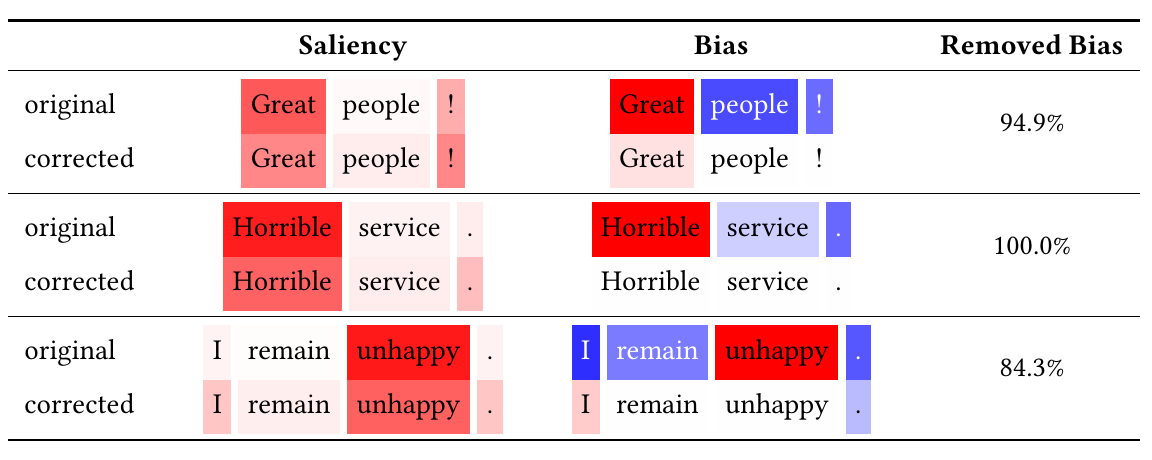}
\end{table*}

\begin{figure*}
    \centering
    \begin{subfigure}[t]{.26\textwidth}
        \centering
    \includegraphics[width=\textwidth]{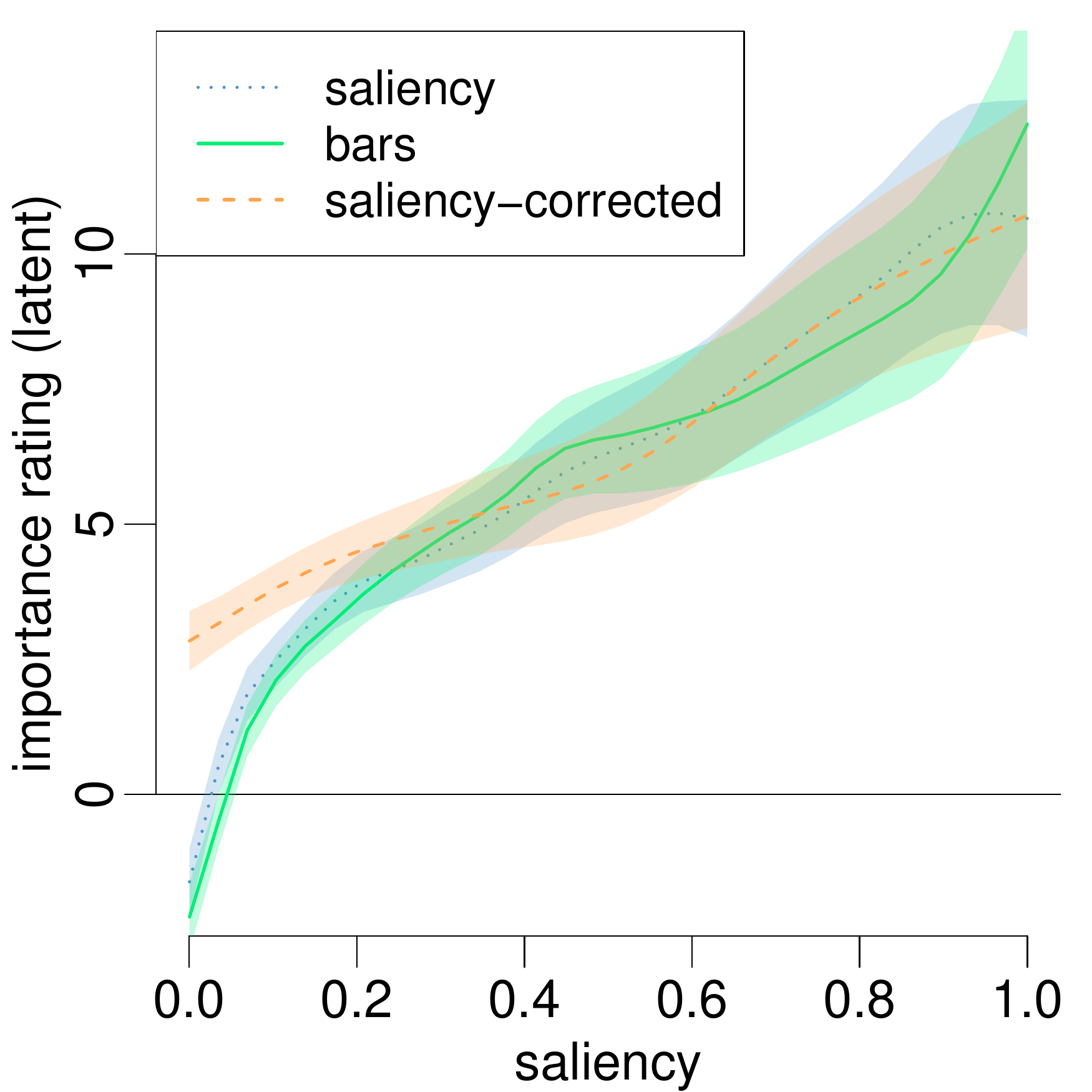}
    \caption{Saliency}\label{fig:correction_comparison_saliency_score_main}
    \end{subfigure}%
    \hfill
    \begin{subfigure}[t]{.26\textwidth}
        \centering
    \includegraphics[width=\textwidth]{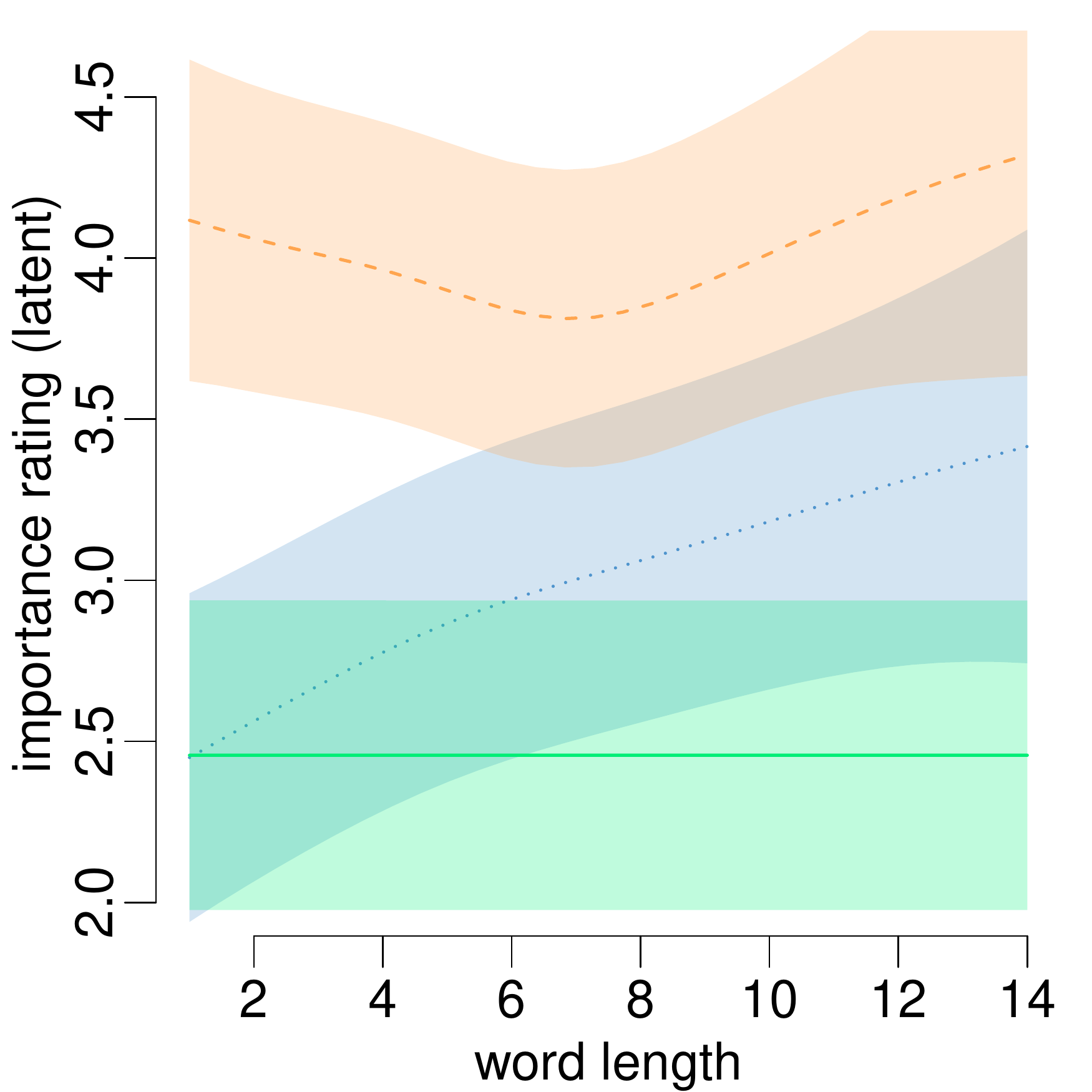}
    \caption{Word length}\label{fig:correction_comparison_word_length_main}
    \end{subfigure}%
    \hfill
    \begin{subfigure}[t]{.26\textwidth}
        \centering
    \includegraphics[width=\textwidth]{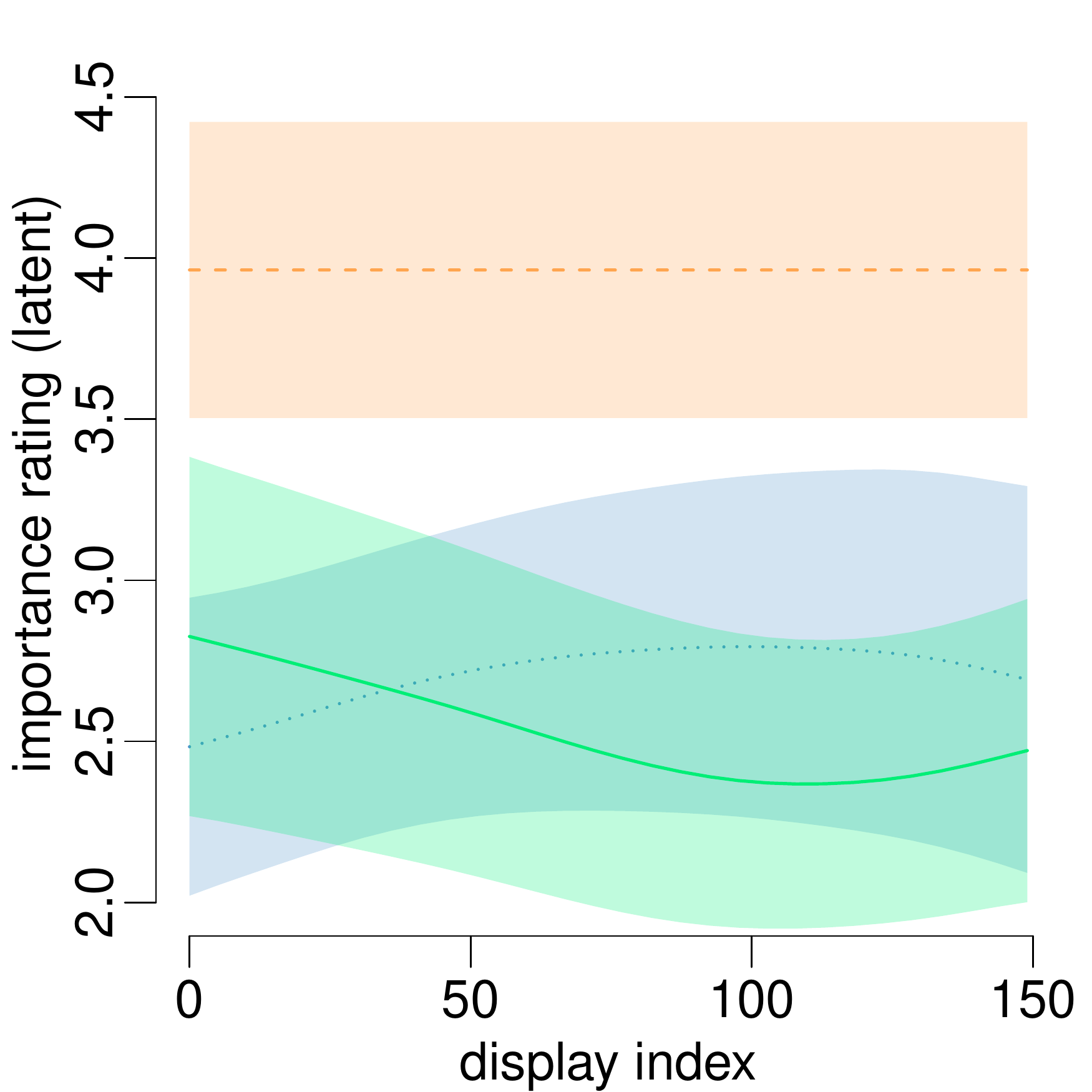}
    \caption{Temporal display index}\label{fig:correction_comparison_display_index_main}
    \end{subfigure}
    \caption{Selected summed-effects comparison plots of the visualization alternatives.}\label{fig:comparison_correction_main}
\end{figure*}

We investigate how well
the two proposed visualization alternatives
counteract bias in user perception within the study described in Section~\ref{sec:study-ig}.
We find that
visualization
has a significant effect on importance ratings (df=2, F=35.45, p<0.0001)
where the bar visualization leads to lower importance ratings ($\beta=-0.5991$, SE=0.1579) and the correction visualization leads to higher importance ratings ($\beta=1.1102$, SE=0.2515).
Regarding the visualizations' effect on smooth terms, we focus on the smooths for color saturation, word length and display index in Figure~\ref{fig:comparison_correction_main}.

Figure~\ref{fig:correction_comparison_saliency_score_main} shows that the saliency scores' effect on importance ratings is similar to the original saliency and the bar visualizations, while the corrected visualization leads to higher importance ratings in the lower end of the color saturation spectrum.
These differences are neither ``good'' nor ``bad''---we argue that the similarity between the original saliency visualization and the bar visualization is remarkable as the two visualizations are fundamentally different. 

Figure~\ref{fig:correction_comparison_word_length_main} shows that the biasing effect of word length in the original visualization is successfully eliminated using the bar visualization as shown by the nearly constant smooth of the bar visualization (edf=0.0009).
This confirms our hypothesis that a bar visualization evades word length bias.
The correction visualization leads to a different effect than the original visualization, however, this effect indicates a different but equally distorting bias of word length.

Figure~\ref{fig:correction_comparison_display_index_main} indicates a successful application of our color correction technique.
While the original visualization as well as the bar visualization show a biasing effect regarding the model smooths, the saliency correction visualization leads to a nearly constant smooth (edf=0.0009). 
Regarding the original and the bar visualizations, the smooths indicate that, in contrast to the original visualization, the bar visualization leads to an initial overestimation of importances which decreases over time, while the original visualization lead to a respective underestimation.
However, a difference plot between the two conditions (see Figure~\ref{fig:difference_bars_saliency_display_index} in Appendix~\ref{sec:appendix_IG}) shows no significant differences.

While these examples demonstrate indications for successful bias mitigation, we want to note that this mitigation cannot be observed for most of the other variables, in particular not for the effect of saliency rank, which we expected to be mitigated by the bar visualization.\footnote{
We provide summed-effect comparison plots for all effects under investigation in Figure~\ref{fig:comparison_correction_all}, difference plots between all conditions in Figures~\ref{fig:differences_corrected-saliency_saliency} and \ref{fig:differences_bars_saliency} as well as details and test statistics on all parametric coefficients as well as smooth terms in Table~\ref{tab:model_params_corr_all} in Appendix~\ref{sec:appendix_IG}.}

Tying back to our initial categorization of biases, we observe that our proposed visualization alternatives can successfully remove instances of visual bias (word length) and learning effect bias (display index).\footnote{We hypothesize that belief biases (such as sentiment polarity) exhibit more distinct expression across indiviuals, which requires subject-adaptive correction methods and should be addressed by online estimation of individual participant slopes and intercepts within our GAMM model in future work.}

\section{Conclusion}
We analyze feature-attribution for text from a novel, arguably under-explored perspective:
we investigate how humans interpret saliency explanations over text and which factors affect their perception.
We show that there can be discrepancy between the communicated information and how it is interpreted, even for a straight-forward and explicit explanation medium of feature-attribution for text.
This is achieved through a general methodology for investigating which factors in the input may cause this discrepancy.
We demonstrate the methodology for a lay-people audience of crowd-workers over multiple tasks, languages and visualizations, showing different setups yield similar but distinct distortions.
We find that word length, sentence length, learning effects, and within-sentence saliency relations affect human importance ratings across multiple user studies.
The methodology can and should be used on other audiences and tasks, before trusting a saliency visualization for this audience/task pair.
We present two bias correction methods and demonstrate their ability to compensate the distorting influence of word length and repeated exposure.
Our findings inform future design of saliency visualizations towards closing the gap between communicated and interpreted saliency explanations, and call for further research in the human factors in interpretation methods of AI, that study not only how the AI operates, but how humans perceive the communicated information.
    

\begin{acks}
We are grateful to Paul Roit, Diego Frassinelli, Yanai Elazar, Harald Baayen, Dagmar Divjak, Sophie Henning, Stefan Grünewald, the BCAI NLP\&KRR group, and the anonymous reviewers for valuable feedback.
A. Jacovi and Y. Goldberg received funding from the European Research Council (ERC) under the European Union’s Horizon 2020 research and innovation programme, grant agreement No. 802774 (iEXTRACT).
N.T. Vu is funded by Carl Zeiss Foundation.
\end{acks}

\bibliographystyle{ACM-Reference-Format}
\bibliography{references}


\clearpage
\appendix
\section{Brief Introduction to Ordinal GAMMs}\label{sec:introduction_gamm}
For an intuitive understanding, we sketch how one arrives at ordinal generalized additive mixed models (GAMMs) starting from linear models.
We follow notation by \citet{wood_generalized_2017}.\vskip5pt

\noindent\textit{Linear Model.}
In a linear model, the response variable $\vec{y}$ (e.g., a numeric rating of importance) is modeled as a function of explanatory variables $\vec{X}$ which are related to $\vec{y}$ \textit{linearly} via parameters $\vec{\beta}$ assuming additional noise $\vec{\epsilon}$:
\begin{equation}
    \vec{y} = \vec{X} \vec{\beta} + \vec{\epsilon}.
\end{equation}

\noindent\textit{Linear Mixed Model.}
In many scenarios, there are \textit{random effects} which one wants to account for in the model.
For example, we collect 150 word importance ratings per participant, i.e., we collect \textit{repeated measures} and are at danger of violating the independence assumption and introducing a confounding effect of the variable \textit{participant ID} because specific participants might have a tendency to give overall higher ratings than other participants.
Like the linear model, linear mixed models estimate \textit{fixed effects} but in addition they also model \textit{random effects} (e.g., of the participant ID) to disentangle their influence on the response variable and thereby offer a clearer view on the fixed effects.
The general formulation of a linear mixed model reads
\begin{equation}
    \vec{y} = \vec{X} \vec{\beta} + \vec{Z} \vec{b} + \vec{\epsilon},
\end{equation}
where $\vec{Z}$ corresponds to the random effects and $\vec{b}$ to the respective weights.\vskip5pt

\noindent\textit{Generalized Linear Model (GLM).}
While linear models require the response distribution to be normal, \textit{generalized} linear models \cite{nelder_generalized_1972} generalize to non-normal (exponential family) response distributions such as
categorical responses (e.g., dog or cat) or ordinal responses (e.g., Likert item ratings).
To achieve this generalization, GLMs link values on the response scale (e.g., categorial ratings) to a latent scale (e.g., logits) via a \textit{link function} $g(\cdot)$ (e.g., logit function):
For a row $i$, the general formulation reads:
\begin{equation}
    g(\mu_i) = \vec{X}_i \vec{\beta}.
\end{equation}

\noindent\textit{Generalized Additive Model (GAM).}
While a generalized linear model only allows to model linear relationships between the the explanatory variables and $g(\mu_i)$, a GAM \cite{hastie1990generalized} generalizes the linear relationship to a \textit{sum of smooth functions} of explanatory variables using:
\begin{equation}
    g(\mu_i) = \vec{X^*}_i \vec{\theta} + f_1(x_{1i}) + f_2(x_{2i}, x_{3i}) + ... \ ,
\end{equation}
where $f_1$ and $f_2$ are smooth functions that typically are chosen to be a sum of basis functions, such as splines and $\vec{X^*}$ corresponds to strictly parametric model components.
A regularized estimation of these functions allows GAMs to model complex functions, but also to fall back to simpler, e.g., constant or linear functions when an increase in model complexity is not sufficiently warranted by improved model fit.\vskip5pt

\noindent\textit{Ordinal Generalized Additive Mixed Model (ordinal GAMM).}
Having introduced the previous models, an ordinal GAMM can be described as a generalized additive model that additionally accounts for random effects and models ordinal ratings via a continuous latent variable that is separated into the ordinal categories via estimated threshold values.
For further details, \citet{divjak_ordinal_2017} provide a practical introduction to ordinal GAMs in a linguistic context and \citet{wood_generalized_2017} offers a detailed textbook on GAM(M)s including implementation and analysis details.

\section{Study Interfaces}\label{sec:appendix_interface}
\begin{figure*}[h!]
    \centering
    \fbox{\includegraphics[width=0.775\textwidth]{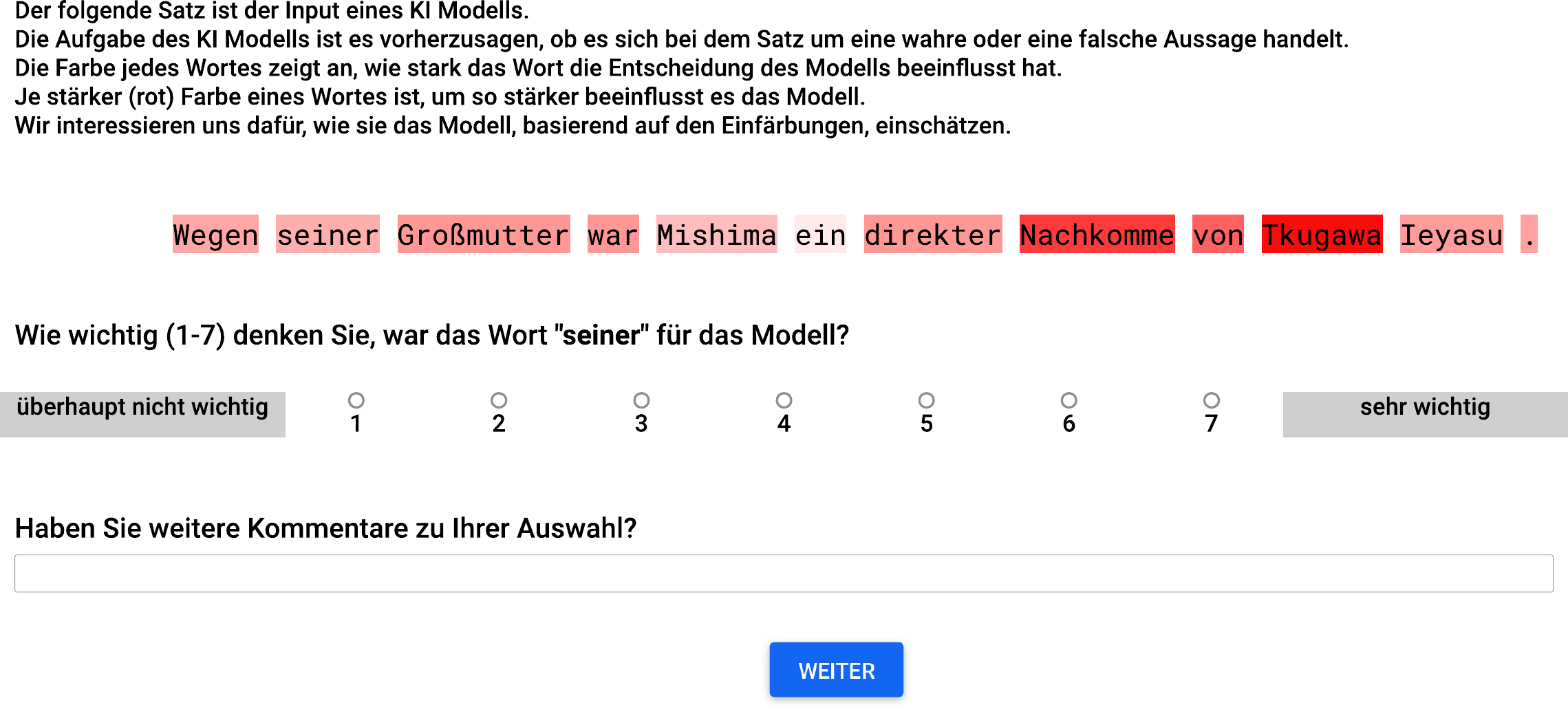}}
    \caption{Screenshot of the importance rating interface for German fact checking sentences using saliency visualization.}
    \label{fig:screenshot_de_saliency}
\end{figure*}

\begin{figure*}[h!]
    \centering
    \fbox{\includegraphics[width=0.775\textwidth]{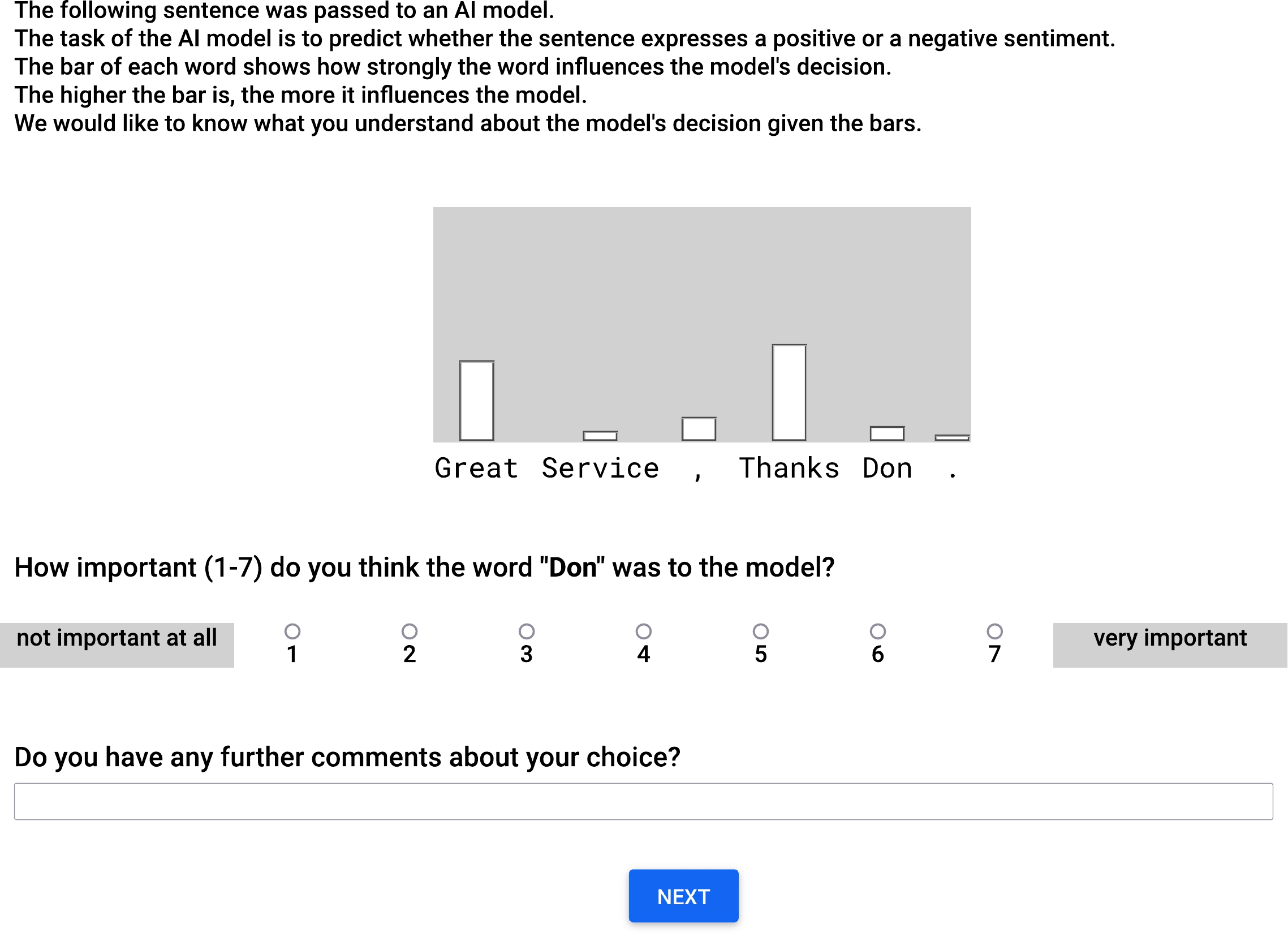}}
    \caption{Screenshot of the importance rating interface for English sentiment sentences using bar visualization.}
    \label{fig:screenshot_en_bars}
\end{figure*}

\begin{figure*}[h!]
    \centering
    \fbox{\includegraphics[width=0.65\textwidth]{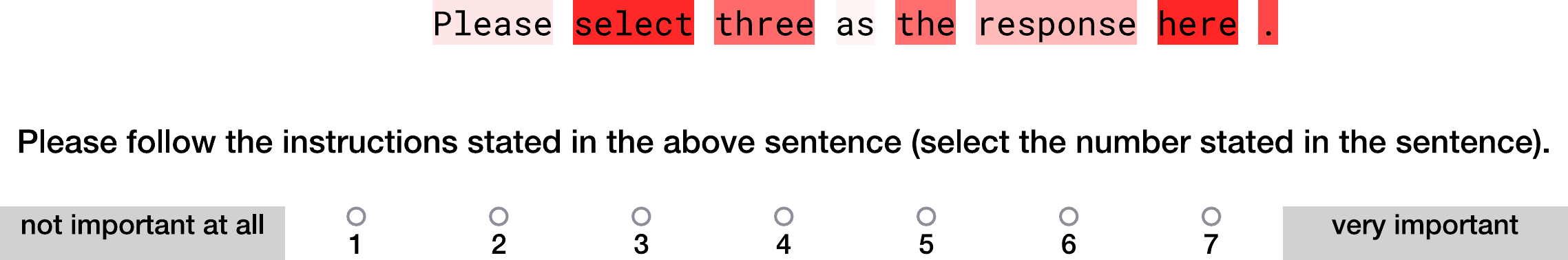}}
    \caption{Screenshot of one of three trap sentences used to validate that the participant pays attention to the task.}
    \label{fig:screenshot_trap_question}
\end{figure*}

In addition to the screenshot shown in Figure~\ref{fig:screenshot_en_saliency}, Figure~\ref{fig:screenshot_de_saliency} shows the interface of the German study and Figure~\ref{fig:screenshot_en_bars} shows an interface that uses the alternative bar chart visualization.
Figure~\ref{fig:screenshot_trap_question} displays one of the three trap questions we use to detect participants that do not pay attention to the task.

\section{English Study Details}\label{sec:appendix_en}
\begin{figure*}
    \centering
    \begin{subfigure}[t]{0.33\textwidth}
        \centering
    \includegraphics[width=0.75\textwidth]{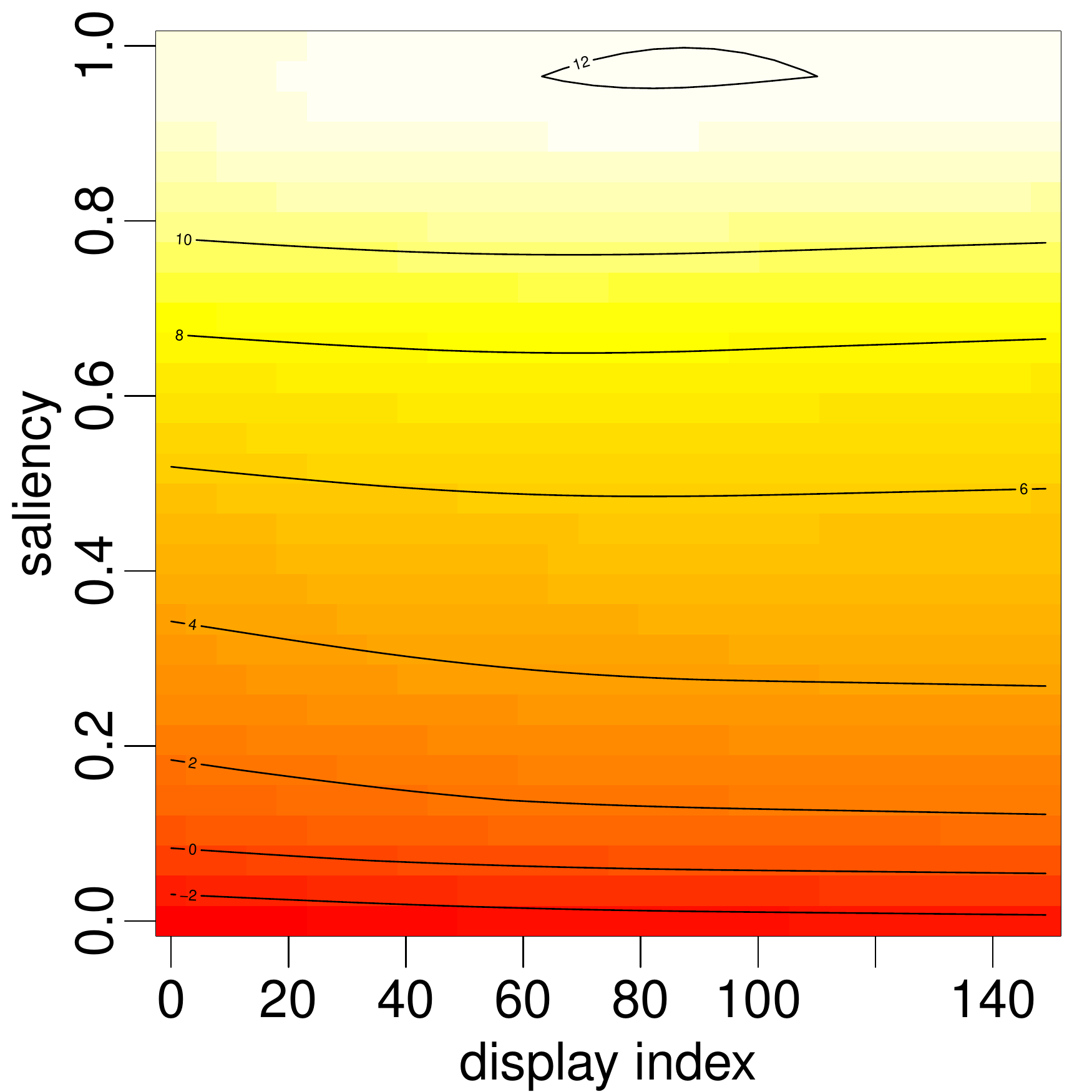}
    \caption{Saliency * display index}
    \end{subfigure}%
    \begin{subfigure}[t]{0.33\textwidth}
        \centering
    \includegraphics[width=0.75\textwidth]{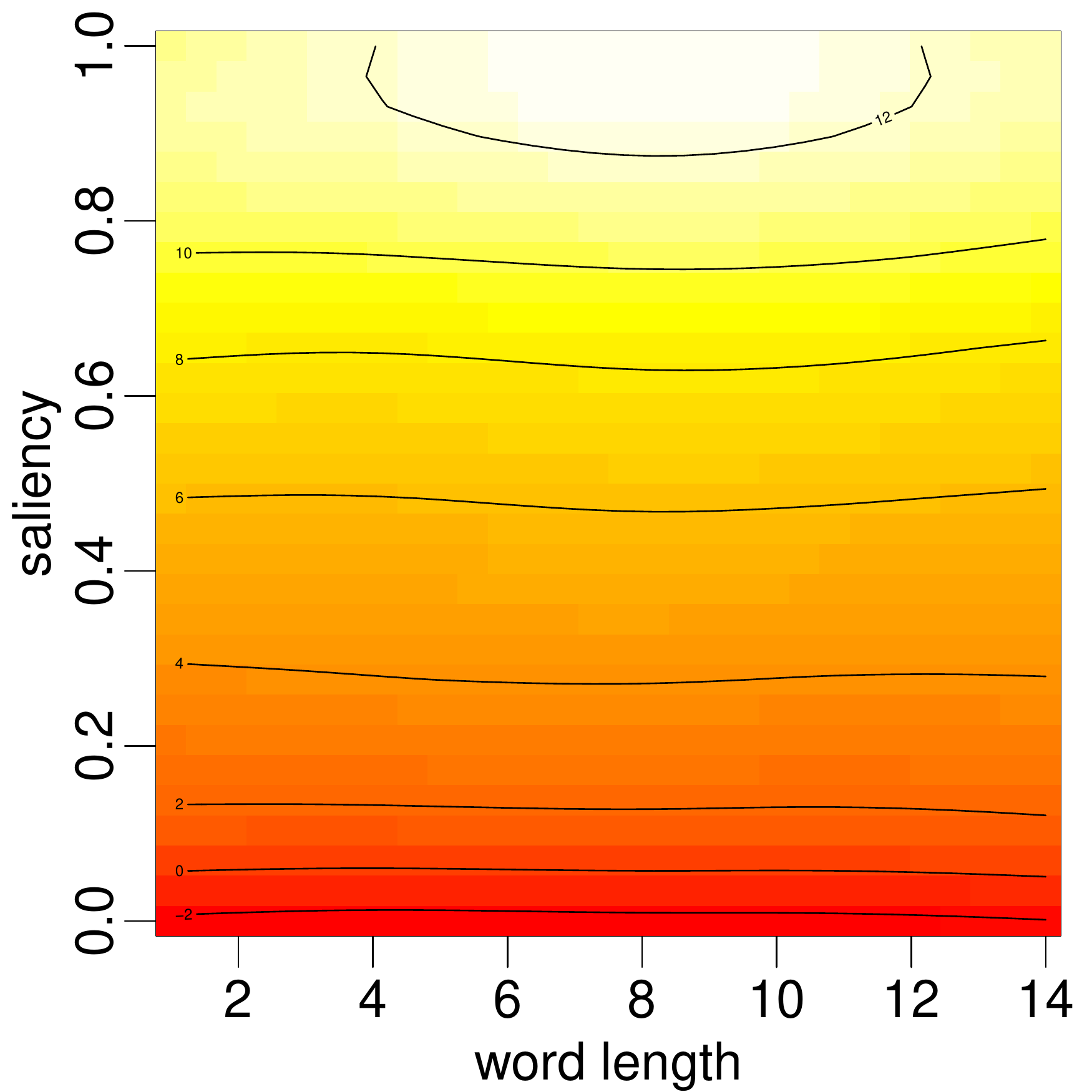}
    \caption{Saliency * word length}\label{fig:interaction_word_length_word_saliency_score}
    \end{subfigure}%
    \begin{subfigure}[t]{0.33\textwidth}
        \centering
    \includegraphics[width=0.75\textwidth]{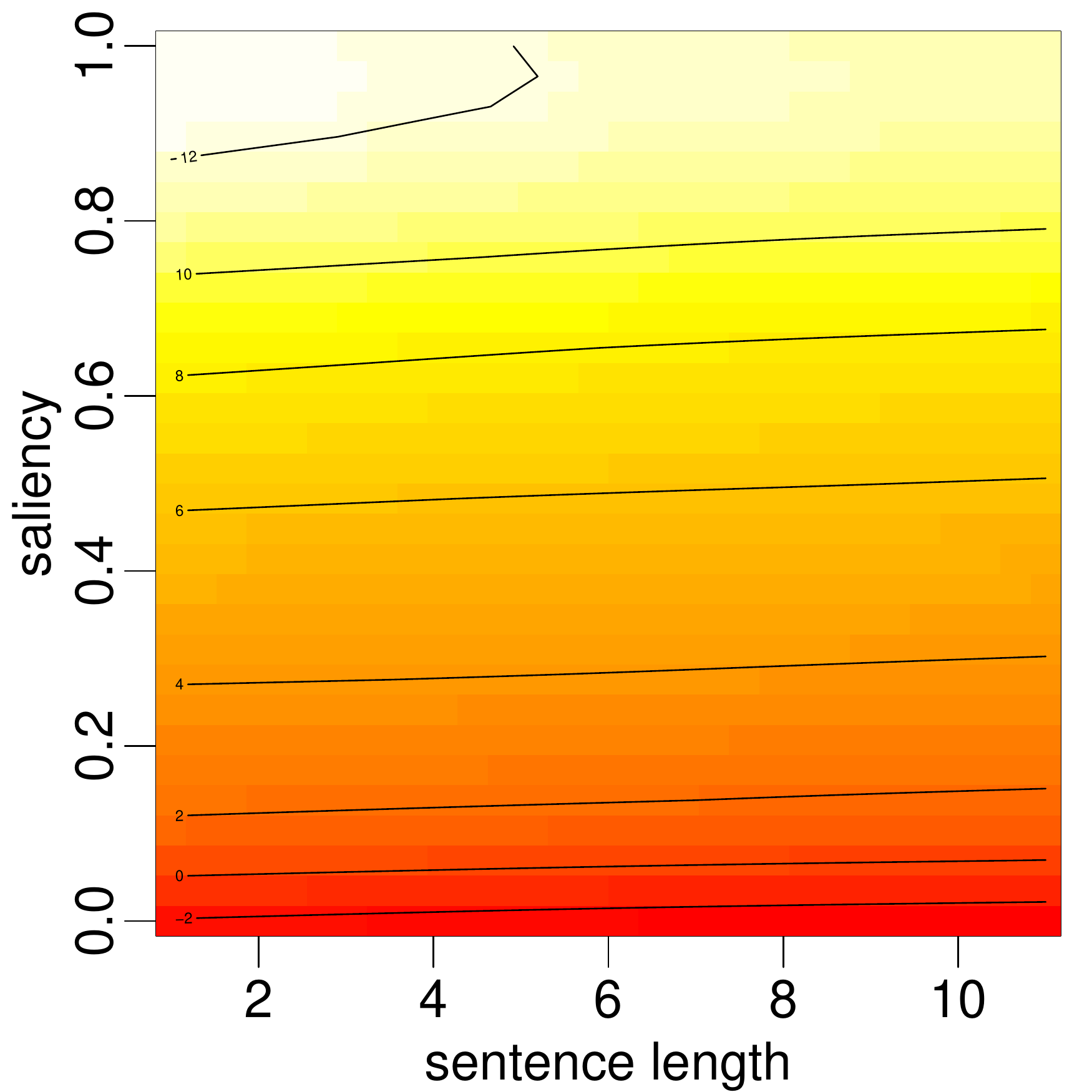}
    \caption{Saliency * sentence length}
    \end{subfigure}
    \begin{subfigure}[t]{0.33\textwidth}
        \centering
    \includegraphics[width=0.75\textwidth]{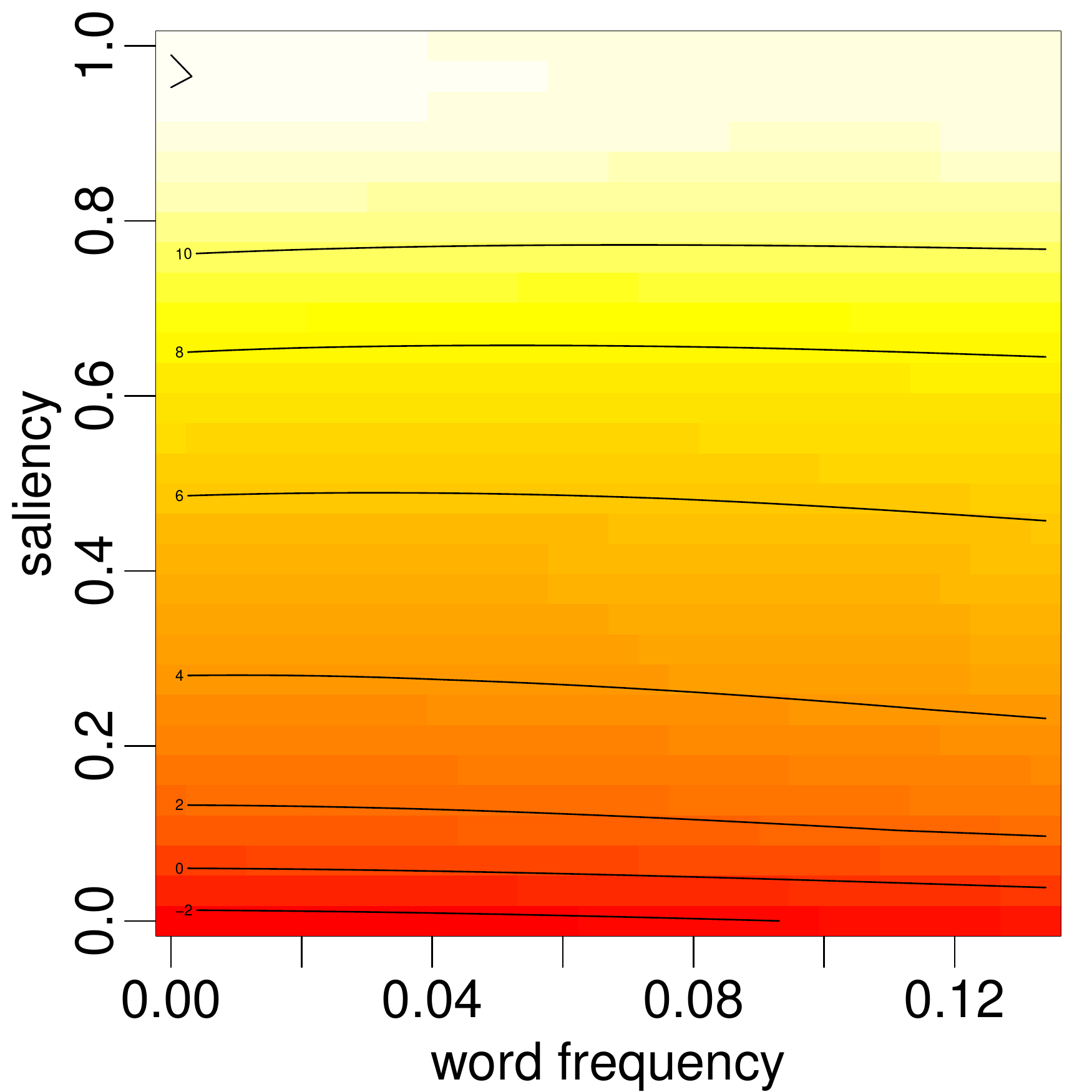}
    \caption{Saliency * word frequency}
    \end{subfigure}%
    \begin{subfigure}[t]{0.33\textwidth}
        \centering
    \includegraphics[width=0.75\textwidth]{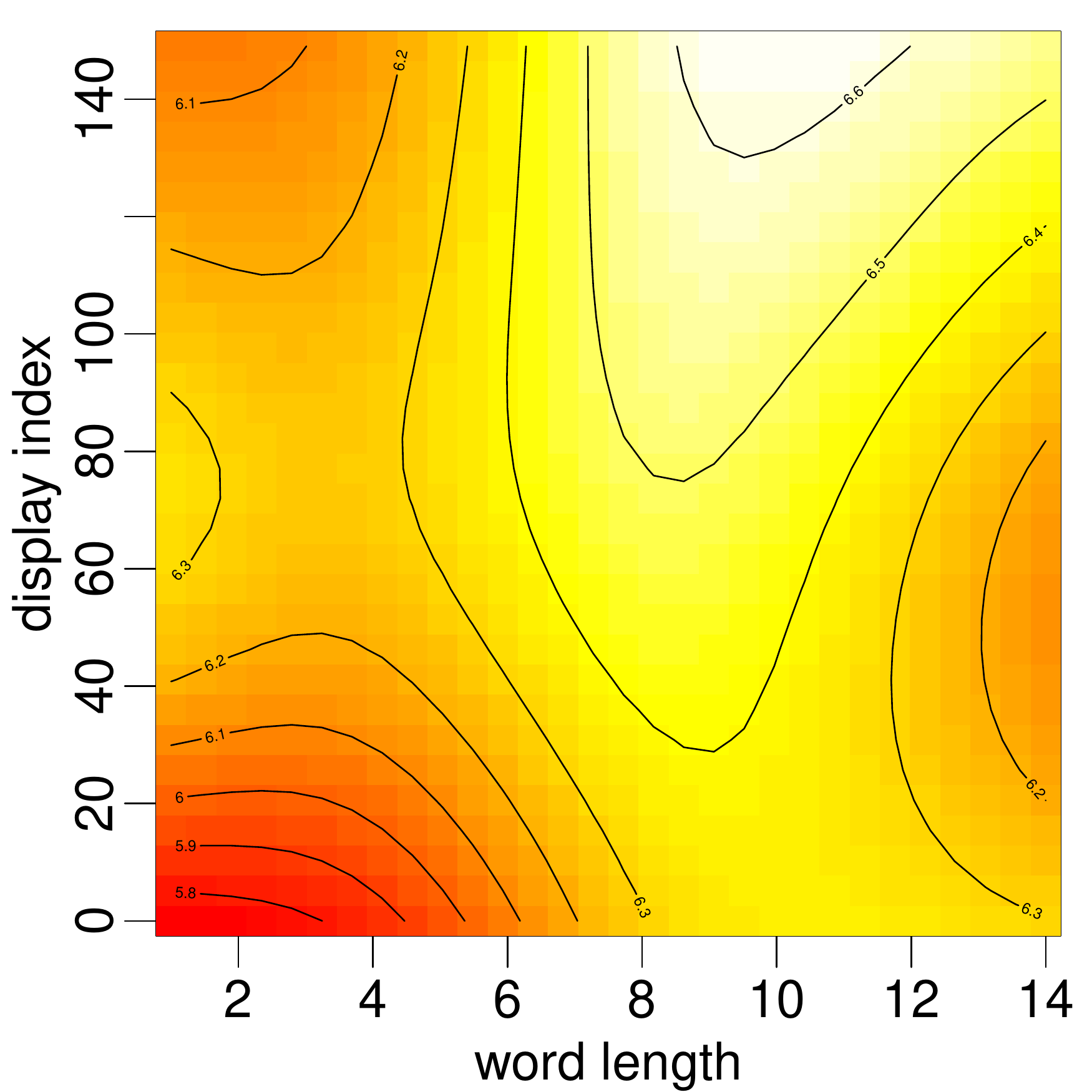}
    \caption{Display index * word length}\label{fig:interaction_word_length_display_index}
    \end{subfigure}%
    \begin{subfigure}[t]{0.33\textwidth}
        \centering
    \includegraphics[width=0.75\textwidth]{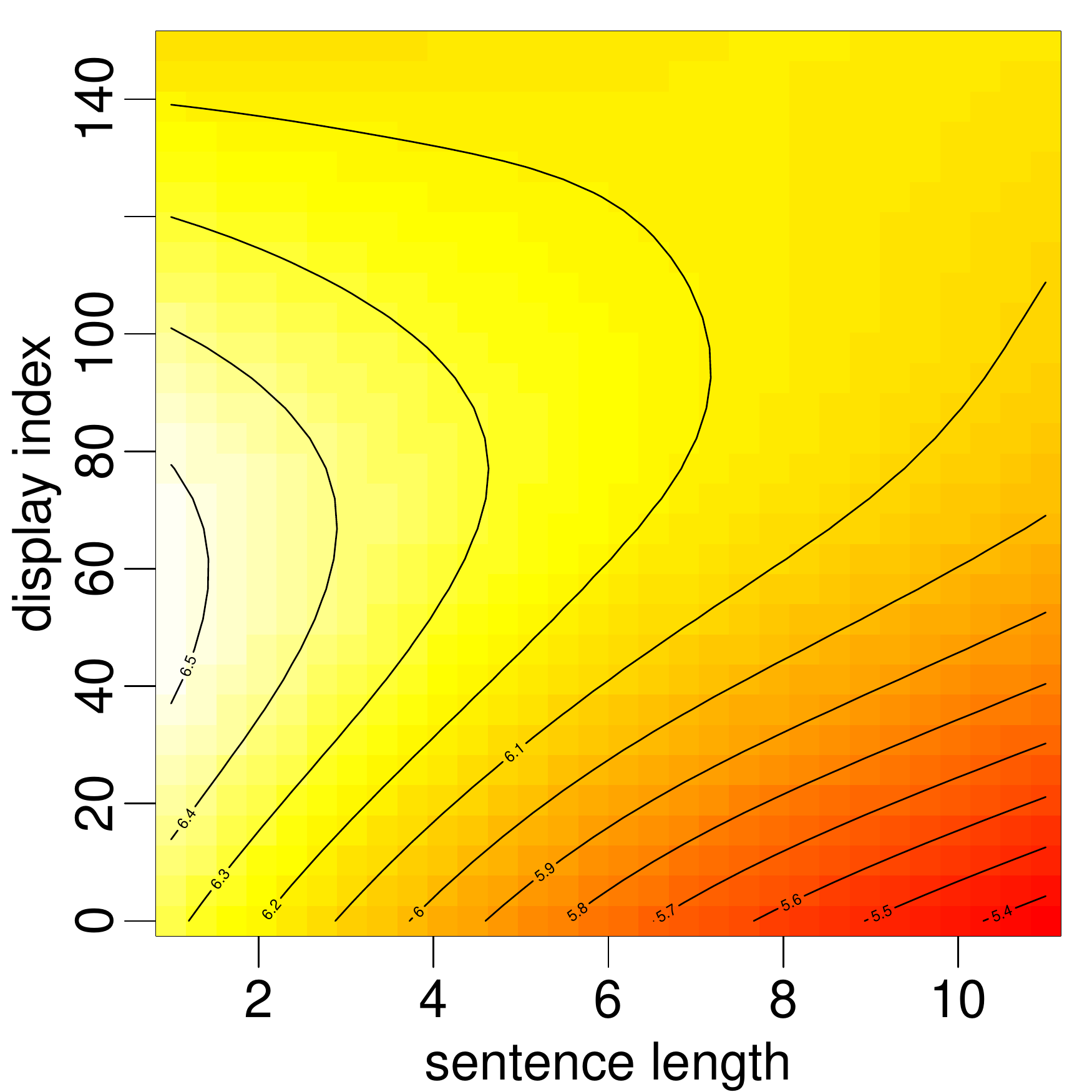}
    \caption{Display index * sentence length}
    \end{subfigure}
    \begin{subfigure}[t]{0.33\textwidth}
        \centering
    \includegraphics[width=0.7\textwidth]{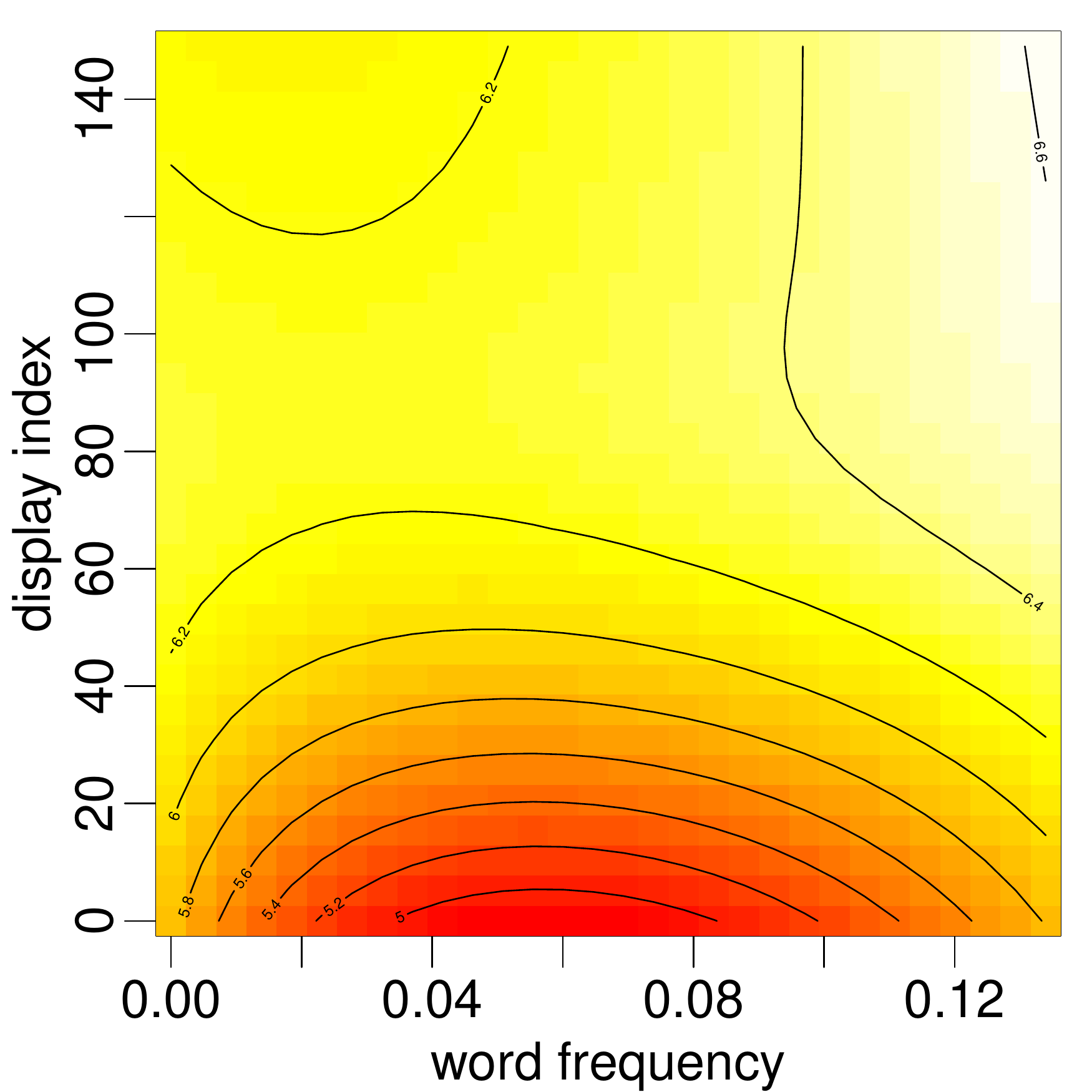}
    \caption{Display index * word frequency}
    \end{subfigure}%
    \begin{subfigure}[t]{0.33\textwidth}
        \centering
    \includegraphics[width=0.75\textwidth]{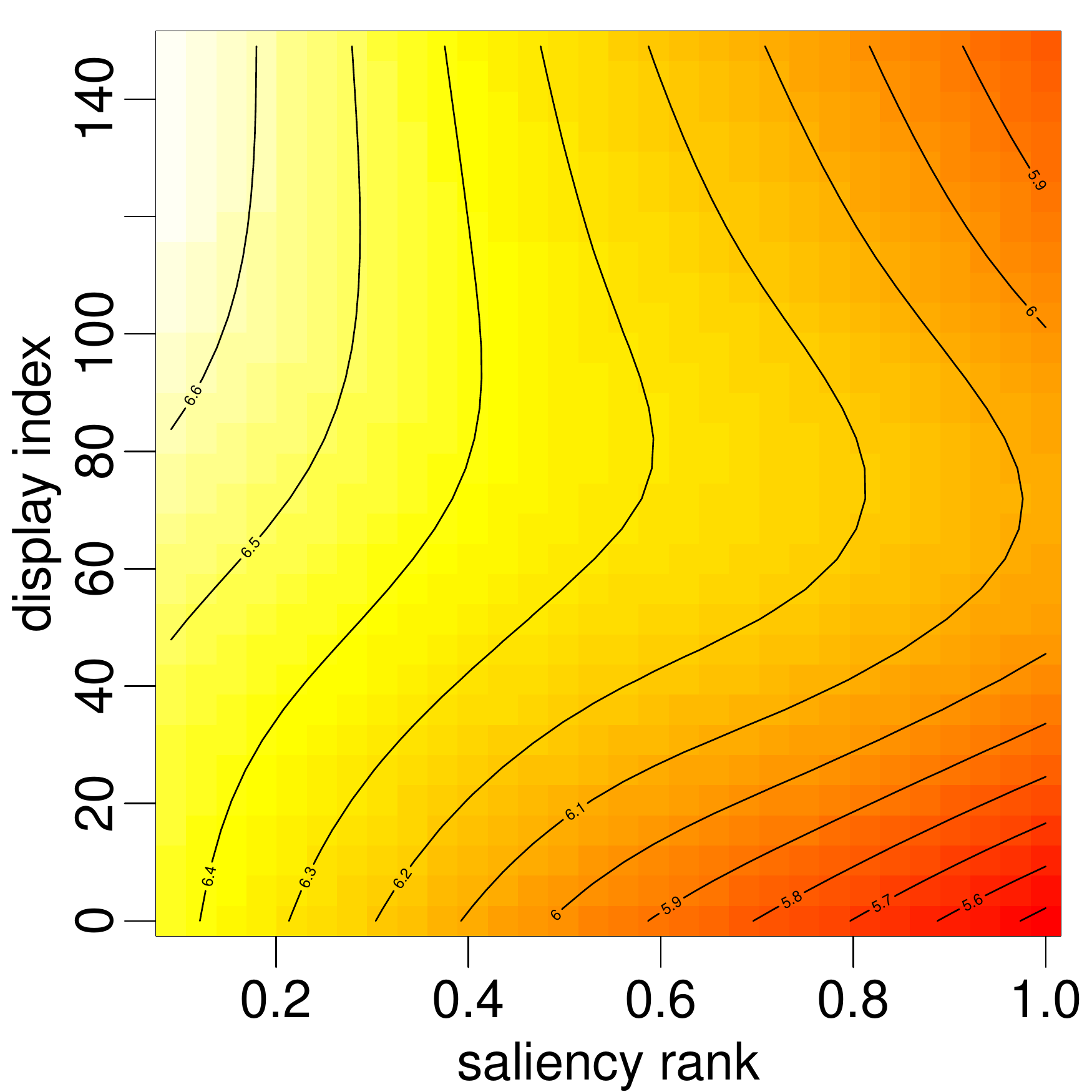}
    \caption{Display index * saliency rank}
    \end{subfigure}%
    \begin{subfigure}[t]{0.33\textwidth}
        \centering
    \includegraphics[width=0.75\textwidth]{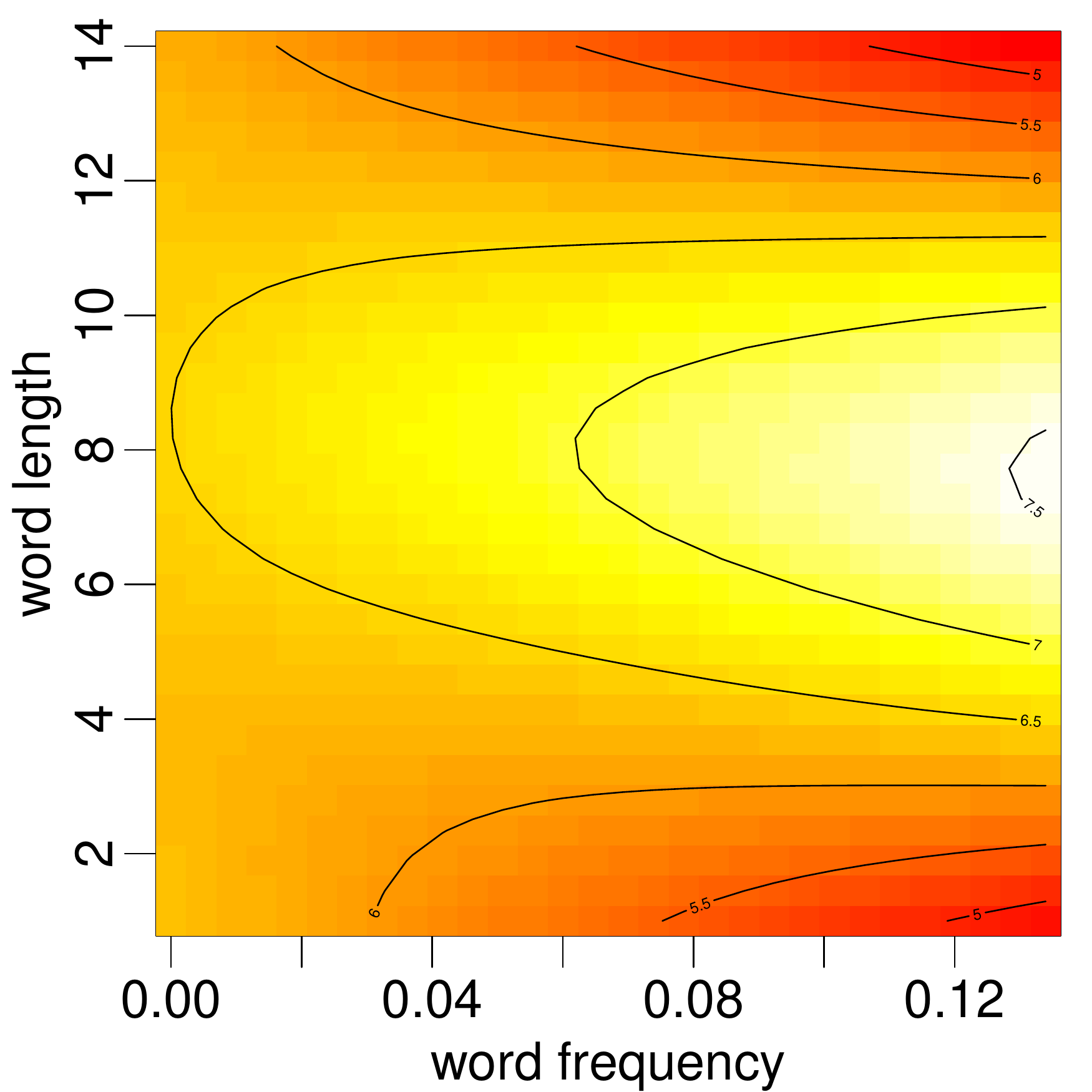}
    \caption{Word length * word frequency}\label{fig:interaction_word_length_relative_word_frequency}
    \end{subfigure}
    \begin{subfigure}[t]{0.33\textwidth}
        \centering
    \includegraphics[width=0.7\textwidth]{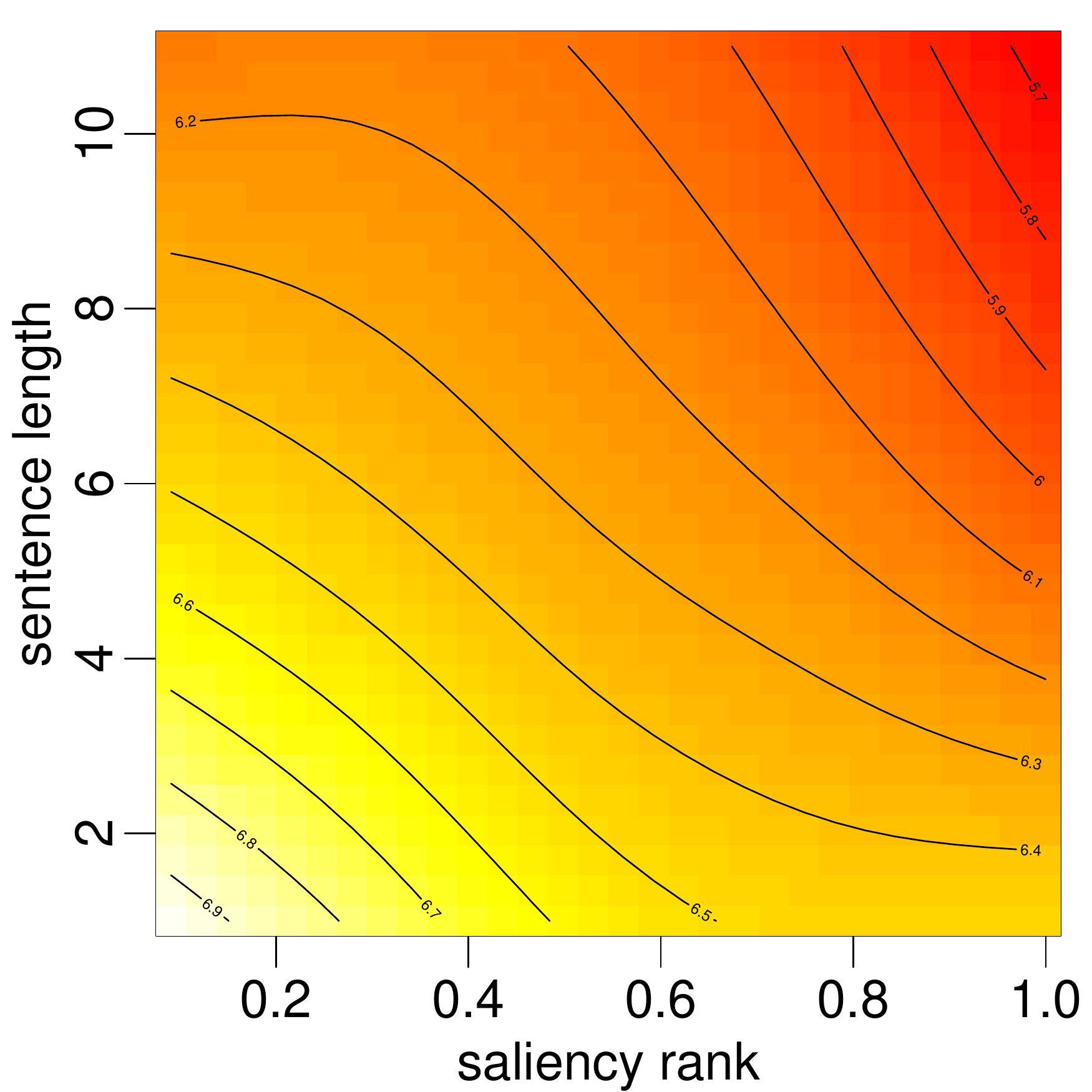}
    \caption{Sentence length * saliency rank}
    \end{subfigure}%
    \begin{subfigure}[t]{0.33\textwidth}
        \centering
    \includegraphics[width=0.75\textwidth]{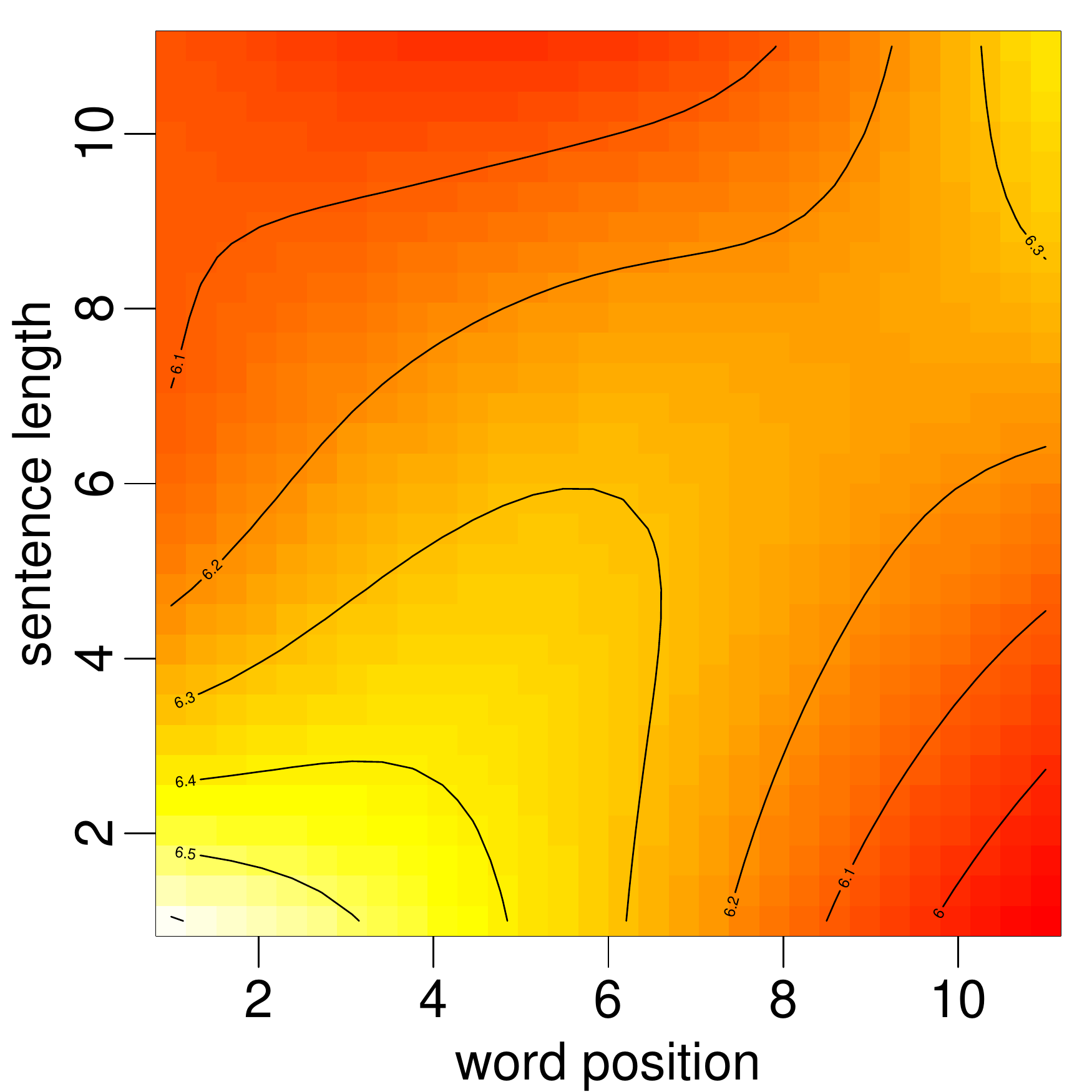}
    \caption{Sentence length * word position}
    \end{subfigure}%
    \caption{Summed effect plots of all significant pairwise interactions.
    }\label{fig:summed_effects}
\end{figure*}

Table~\ref{tab:tests_smooth_interactions} displays test statistics for all smooth pairwise interactions.
We make use of tensor interaction smooths following a functional ANOVA decomposition.
Figure~\ref{fig:summed_effects} shows summed effect plots for the respective significant interactions.
Ordered categorial cut points are located at -1, 1.31, 3.29, 5.15, 7.1 and 9.22.
\begin{table}[ht]
\centering
\ra{1.3}
\caption{Wald tests for the pairwise interactions (tensor interactions) (upper) and random effects (lower) of the English user study.}\label{tab:tests_smooth_interactions}
\resizebox{1\linewidth}{!}{
\begin{tabular}{lrrrr}
  \hline
 & \textbf{edf} & \textbf{ref. df} & \textbf{F} & \textbf{p} \\ 
  \hline
\rowcolor{lightgray}ti(saliency,display index) & 2.5102 & 16 & 2.1075 & 0.0001 \\ 
  ti(saliency,word length) & 6.0566 & 16 & 2.2698 & 0.0001 \\ 
\rowcolor{lightgray}ti(saliency,sentence length) & 3.1609 & 16 & 1.1203 & 0.0020 \\ 
  ti(saliency,word frequency) & 0.9176 & 12 & 1.8325 & 0.0004 \\ 
\rowcolor{lightgray}ti(saliency,sentiment polarity) & 2.9357 & 16 & 0.5553 & 0.0814 \\ 
  ti(saliency,saliency rank) & 0.0004 & 16 & 0.0000 & 0.5864 \\ 
\rowcolor{lightgray}ti(saliency,word position) & 0.6254 & 16 & 0.1144 & 0.1276 \\ 
  ti(display index,word length) & 1.5112 & 16 & 0.6637 & 0.0026 \\ 
\rowcolor{lightgray}ti(display index,sentence length) & 1.2776 & 16 & 1.0159 & 0.0010 \\ 
  ti(display index,word frequency) & 2.6938 & 16 & 1.7810 & 0.0001 \\ 
\rowcolor{lightgray}ti(display index,sentiment polarity) & 0.5386 & 16 & 0.0853 & 0.1678 \\ 
  ti(display index,saliency rank) & 1.3966 & 16 & 0.5272 & 0.0174 \\ 
\rowcolor{lightgray}ti(display index,word position) & 3.3649 & 16 & 0.6625 & 0.0520 \\ 
  ti(word length,sentence length) & 0.0004 & 16 & 0.0000 & 0.9236 \\ 
\rowcolor{lightgray}ti(word length,word frequency) & 2.1540 & 16 & 6.5510 & $<$ 0.0001 \\ 
  ti(word length,sentiment polarity) & 0.0014 & 16 & 0.0000 & 0.6790 \\ 
\rowcolor{lightgray}ti(word length,saliency rank) & 2.2175 & 16 & 0.3503 & 0.0573 \\ 
  ti(word length,word position) & 1.0296 & 16 & 0.1270 & 0.1222 \\ 
\rowcolor{lightgray}ti(sentence length,word frequency) & 0.0005 & 16 & 0.0000 & 0.8608 \\ 
  ti(sentence length,sentiment polarity) & 0.0013 & 16 & 0.0001 & 0.5113 \\ 
\rowcolor{lightgray}ti(sentence length,saliency rank) & 1.3045 & 16 & 0.2651 & 0.0453 \\ 
  ti(sentence length,word position) & 3.1995 & 16 & 0.8487 & 0.0067 \\ 
\rowcolor{lightgray}ti(word frequency,sentiment polarity) & 0.0015 & 16 & 0.0001 & 0.1969 \\ 
  ti(word frequency,saliency rank) & 0.0022 & 15 & 0.0001 & 0.3230 \\ 
\rowcolor{lightgray}ti(word frequency,word position) & 2.0375 & 16 & 0.3168 & 0.0924 \\ 
  ti(sentiment polarity,saliency rank) & 0.0006 & 16 & 0.0000 & 0.8407 \\ 
\rowcolor{lightgray}ti(sentiment polarity,word position) & 0.0005 & 16 & 0.0000 & 0.9558 \\ 
  ti(saliency rank,word position) & 0.0006 & 16 & 0.0000 & 0.6542 \\ 
  \hline
\rowcolor{lightgray}s(sentence\_id) & 0.0006 & 150 & 0.0000 & 0.9276 \\ 
  s(saliency,sentence\_id) & 9.1441 & 150 & 0.0676 & 0.2305 \\ 
\rowcolor{lightgray}s(worker\_id) & 48.1065 & 49 & 10640.8475 & $<$ 0.0001 \\ 
  s(saliency,worker\_id) & 48.0654 & 50 & 6593.7769 & $<$ 0.0001 \\ 
   \hline
\end{tabular}
}
\end{table}

\begin{table}[ht]
\centering
\ra{1.3}
\caption{Wald tests for the parametric terms of the German user study.}\label{tab:tests_parametric_de}
\resizebox{0.31\textwidth}{!}{
\begin{tabular}{rrrr}
  \hline
 &\textbf{df} & \textbf{F} & \textbf{p} \\ 
  \hline
capitalization & 2 & 7.62 & 0.0005 \\ 
  dependency relation & 33 & 2.57 & $<$ 0.0001 \\ 
   \hline
\end{tabular}
}
\end{table}

\begin{table}[ht]
\centering
\ra{1.3}

\caption{Wald tests for the smooth terms of the German user study.}\label{tab:tests_smooth_de}
\resizebox{1\linewidth}{!}{
\begin{tabular}{rrrrr}
  \hline
 & \textbf{edf} & \textbf{ref. df} & \textbf{F} & \textbf{p} \\ 
  \hline
\rowcolor{lightgray}s(saliency) & 8.2052 & 19 & 148.1115 & $<$ 0.0001 \\ 
  s(display index) & 1.5999 &9 & 2.4742 & $<$ 0.0001 \\ 
\rowcolor{lightgray}s(word length) & 2.0440 &9 & 3.9174 & $<$ 0.0001 \\ 
  s(sentence length) & 0.9073 &9 & 1.7657 & 0.0003 \\ 
\rowcolor{lightgray}s(word frequency) & 0.0017 &9 & 0.0002 & 0.2816 \\ 
  s(saliency rank) & 0.0004 &9 & 0.0000 & 0.7016 \\ 
\rowcolor{lightgray}s(word position) & 2.4429 &9 & 2.8142 & 0.0002 \\ 
  ti(saliency,display index) & 0.0007 & 16 & 0.0000 & 0.5846 \\ 
\rowcolor{lightgray}ti(saliency,word length) & 2.4114 & 16 & 1.1662 & 0.0013 \\ 
  ti(saliency,sentence length) & 1.8496 & 16 & 0.7410 & 0.0125 \\ 
\rowcolor{lightgray}ti(saliency,word frequency) & 0.6953 & 11 & 0.3084 & 0.0549 \\ 
  ti(saliency,saliency rank) & 1.4340 & 16 & 0.4958 & 0.0142 \\ 
\rowcolor{lightgray}ti(saliency,word position) & 0.0765 & 16 & 0.0053 & 0.2970 \\ 
  ti(display index,word length) & 0.3529 & 16 & 0.0477 & 0.1968 \\ 
\rowcolor{lightgray}ti(display index,sentence length) & 0.1902 & 16 & 0.0171 & 0.2622 \\ 
  ti(display index,word frequency) & 0.0005 & 15 & 0.0000 & 0.7096 \\ 
\rowcolor{lightgray}ti(display index,saliency rank) & 0.2967 & 16 & 0.0332 & 0.2325 \\ 
  ti(display index,word position) & 1.1440 & 16 & 0.4244 & 0.0168 \\ 
\rowcolor{lightgray}ti(word length,sentence length) & 0.9858 & 16 & 0.3138 & 0.0290 \\ 
  ti(word length,word frequency) & 0.9622 & 11 & 1.0293 & 0.0050 \\ 
\rowcolor{lightgray}ti(word length,saliency rank) & 0.0005 & 16 & 0.0000 & 0.8581 \\ 
  ti(word length,word position) & 0.8285 & 16 & 0.5132 & 0.0091 \\ 
\rowcolor{lightgray}ti(sentence length,word frequency) & 0.0009 & 15 & 0.0001 & 0.3536 \\ 
  ti(sentence length,saliency rank) & 0.0005 & 16 & 0.0000 & 0.9945 \\ 
\rowcolor{lightgray}ti(sentence length,word position) & 0.0005 & 16 & 0.0000 & 0.6862 \\ 
  ti(word frequency,saliency rank) & 0.0003 & 16 & 0.0000 & 0.9438 \\ 
\rowcolor{lightgray}ti(word frequency,word position) & 0.0005 & 15 & 0.0000 & 0.6085 \\ 
  ti(saliency rank,word position) & 0.0004 & 16 & 0.0000 & 0.8379 \\ 
\rowcolor{lightgray}s(sentence ID) & 0.0004 & 149 & 0.0000 & 0.9007 \\ 
  s(saliency,sentence ID) & 36.6567 & 150 & 0.3534 & 0.0087 \\ 
\rowcolor{lightgray}s(worker ID) & 23.5324 & 24 & 8128.6327 & $<$ 0.0001 \\ 
  s(saliency,worker ID) & 23.6122 & 25 & 5645.0812 & $<$ 0.0001 \\ 
   \hline
\end{tabular}}
\end{table}

\begin{table}[ht]
\centering
\ra{1.3}
\caption{Capitalization and dependency relation coefficients for the German user study.} \label{tab:coefficients_details_de}
\resizebox{1\linewidth}{!}{
\begin{tabular}{lrrrr}
   \hline
\textbf{Coefficients} & \textbf{$\beta$} & \textbf{SE} & \textbf{t} & \textbf{p} \\ 
\hline
\rowcolor{lightgray}capitalization: all capital & 1.9051 & 0.9638 & 1.9767 & 0.0481 \\ 
  capitalization: first capital & 0.4074 & 0.1151 & 3.5390 & 0.0004 \\ 
\hline
\rowcolor{lightgray}dependency relation: acl & -1.2155 & 0.6428 & -1.8910 & 0.0587 \\ 
  dependency relation: acl:relcl & 1.3605 & 0.5947 & 2.2878 & 0.0222 \\ 
\rowcolor{lightgray}dependency relation: advcl & 0.8647 & 0.7154 & 1.2087 & 0.2269 \\ 
  dependency relation: advmod & 0.3741 & 0.2369 & 1.5790 & 0.1144 \\ 
\rowcolor{lightgray}dependency relation: amod & 0.4794 & 0.2653 & 1.8072 & 0.0708 \\ 
  dependency relation: appos & 0.2823 & 0.4119 & 0.6852 & 0.4932 \\ 
\rowcolor{lightgray}dependency relation: aux & 0.6395 & 0.3138 & 2.0379 & 0.0416 \\ 
  dependency relation: aux:pass & -0.0679 & 0.3798 & -0.1789 & 0.8581 \\ 
\rowcolor{lightgray}dependency relation: case & 0.1169 & 0.2082 & 0.5613 & 0.5746 \\ 
  dependency relation: cc & 0.1126 & 0.2571 & 0.4379 & 0.6615 \\ 
\rowcolor{lightgray}dependency relation: cc:preconj & 0.8039 & 1.1491 & 0.6996 & 0.4842 \\ 
  dependency relation: ccomp & 1.1850 & 0.5206 & 2.2763 & 0.0229 \\ 
\rowcolor{lightgray}dependency relation: compound & 0.8738 & 0.4488 & 1.9470 & 0.0516 \\ 
  dependency relation: compound:prt & 0.4114 & 0.4577 & 0.8989 & 0.3688 \\ 
\rowcolor{lightgray}dependency relation: conj & 0.1673 & 0.2900 & 0.5769 & 0.5640 \\ 
  dependency relation: cop & 0.4169 & 0.2598 & 1.6043 & 0.1087 \\ 
\rowcolor{lightgray}dependency relation: csubj & 1.0154 & 0.7533 & 1.3480 & 0.1777 \\ 
  dependency relation: det & -0.1604 & 0.2088 & -0.7682 & 0.4424 \\ 
\rowcolor{lightgray}dependency relation: expl & -1.0130 & 0.4605 & -2.1998 & 0.0279 \\ 
  dependency relation: flat:name & 0.3786 & 0.5401 & 0.7010 & 0.4833 \\ 
\rowcolor{lightgray}dependency relation: iobj & -0.4807 & 0.5162 & -0.9312 & 0.3518 \\ 
  dependency relation: mark & 0.1537 & 0.3646 & 0.4216 & 0.6734 \\ 
\rowcolor{lightgray}dependency relation: nmod & 0.4656 & 0.2787 & 1.6707 & 0.0949 \\ 
  dependency relation: nmod:poss & -0.0658 & 0.3251 & -0.2025 & 0.8395 \\ 
\rowcolor{lightgray}dependency relation: nsubj & 0.4443 & 0.2369 & 1.8755 & 0.0608 \\ 
  dependency relation: nsubj:pass & 0.4296 & 0.4305 & 0.9979 & 0.3184 \\ 
\rowcolor{lightgray}dependency relation: nummod & 1.3866 & 0.3609 & 3.8419 & 0.0001 \\ 
  dependency relation: obj & 0.2406 & 0.2649 & 0.9082 & 0.3638 \\ 
\rowcolor{lightgray}dependency relation: obl & 0.3126 & 0.2679 & 1.1668 & 0.2434 \\ 
  dependency relation: obl:tmod & 1.7042 & 0.5544 & 3.0739 & 0.0021 \\ 
\rowcolor{lightgray}dependency relation: parataxis & -0.2780 & 0.8595 & -0.3234 & 0.7464 \\ 
  dependency relation: root & 0.5463 & 0.2432 & 2.2460 & 0.0248 \\ 
\rowcolor{lightgray}dependency relation: xcomp & 0.6718 & 0.4494 & 1.4948 & 0.1351 \\ 
\hline
\end{tabular}
}
\end{table}

\clearpage
\definecolor{c_255_210_210}{RGB}{255,210,210}
\definecolor{c_255_171_171}{RGB}{255,171,171}
\definecolor{c_255_192_192}{RGB}{255,192,192}
\definecolor{c_255_74_74}{RGB}{255,74,74}
\definecolor{c_255_173_173}{RGB}{255,173,173}
\definecolor{c_255_240_240}{RGB}{255,240,240}
\definecolor{c_255_57_57}{RGB}{255,57,57}
\definecolor{c_255_224_224}{RGB}{255,224,224}
\definecolor{c_255_20_20}{RGB}{255,20,20}
\definecolor{c_255_227_227}{RGB}{255,227,227}
\definecolor{c_255_111_111}{RGB}{255,111,111}
\definecolor{c_255_189_189}{RGB}{255,189,189}
\definecolor{c_255_138_138}{RGB}{255,138,138}
\definecolor{c_255_82_82}{RGB}{255,82,82}
\definecolor{c_255_176_176}{RGB}{255,176,176}
\definecolor{c_255_118_118}{RGB}{255,118,118}
\definecolor{c_255_56_56}{RGB}{255,56,56}
\definecolor{c_255_204_204}{RGB}{255,204,204}
\definecolor{c_255_83_83}{RGB}{255,83,83}
\definecolor{c_255_75_75}{RGB}{255,75,75}
\definecolor{c_255_125_125}{RGB}{255,125,125}
\definecolor{c_255_41_41}{RGB}{255,41,41}
\definecolor{c_255_78_78}{RGB}{255,78,78}
\definecolor{c_255_155_155}{RGB}{255,155,155}
\definecolor{c_255_91_91}{RGB}{255,91,91}
\definecolor{c_255_232_232}{RGB}{255,232,232}
\definecolor{c_255_30_30}{RGB}{255,30,30}
\definecolor{c_255_203_203}{RGB}{255,203,203}
\definecolor{c_255_241_241}{RGB}{255,241,241}
\definecolor{c_255_218_218}{RGB}{255,218,218}
\definecolor{c_255_173_173}{RGB}{255,173,173}
\definecolor{c_255_171_171}{RGB}{255,171,171}
\definecolor{c_255_73_73}{RGB}{255,73,73}
\definecolor{c_255_22_22}{RGB}{255,22,22}
\definecolor{c_255_30_30}{RGB}{255,30,30}
\definecolor{c_255_66_66}{RGB}{255,66,66}
\definecolor{c_255_74_74}{RGB}{255,74,74}
\definecolor{c_255_158_158}{RGB}{255,158,158}
\definecolor{c_255_3_3}{RGB}{255,3,3}
\definecolor{c_255_156_156}{RGB}{255,156,156}
\definecolor{c_255_2_2}{RGB}{255,2,2}
\definecolor{c_255_208_208}{RGB}{255,208,208}
\definecolor{c_255_101_101}{RGB}{255,101,101}
\definecolor{c_255_40_40}{RGB}{255,40,40}
\definecolor{c_255_191_191}{RGB}{255,191,191}
\definecolor{c_255_245_245}{RGB}{255,245,245}
\definecolor{c_255_164_164}{RGB}{255,164,164}
\definecolor{c_255_162_162}{RGB}{255,162,162}
\definecolor{c_255_84_84}{RGB}{255,84,84}
\definecolor{c_255_248_248}{RGB}{255,248,248}
\definecolor{c_255_44_44}{RGB}{255,44,44}
\definecolor{c_255_100_100}{RGB}{255,100,100}
\definecolor{c_255_51_51}{RGB}{255,51,51}
\definecolor{c_255_91_91}{RGB}{255,91,91}
\definecolor{c_255_239_239}{RGB}{255,239,239}
\definecolor{c_255_33_33}{RGB}{255,33,33}
\definecolor{c_255_5_5}{RGB}{255,5,5}
\definecolor{c_255_166_166}{RGB}{255,166,166}
\definecolor{c_255_118_118}{RGB}{255,118,118}
\definecolor{c_255_188_188}{RGB}{255,188,188}
\definecolor{c_255_212_212}{RGB}{255,212,212}
\definecolor{c_255_98_98}{RGB}{255,98,98}
\definecolor{c_255_211_211}{RGB}{255,211,211}
\definecolor{c_255_62_62}{RGB}{255,62,62}
\definecolor{c_255_193_193}{RGB}{255,193,193}
\definecolor{c_255_157_157}{RGB}{255,157,157}
\definecolor{c_255_232_232}{RGB}{255,232,232}
\definecolor{c_255_21_21}{RGB}{255,21,21}
\definecolor{c_255_194_194}{RGB}{255,194,194}
\definecolor{c_255_99_99}{RGB}{255,99,99}
\definecolor{c_255_144_144}{RGB}{255,144,144}
\definecolor{c_255_136_136}{RGB}{255,136,136}
\definecolor{c_255_71_71}{RGB}{255,71,71}
\definecolor{c_255_227_227}{RGB}{255,227,227}
\definecolor{c_255_165_165}{RGB}{255,165,165}
\definecolor{c_255_188_188}{RGB}{255,188,188}
\definecolor{c_255_36_36}{RGB}{255,36,36}
\definecolor{c_255_51_51}{RGB}{255,51,51}
\definecolor{c_255_187_187}{RGB}{255,187,187}
\definecolor{c_255_49_49}{RGB}{255,49,49}
\definecolor{c_255_81_81}{RGB}{255,81,81}
\definecolor{c_255_159_159}{RGB}{255,159,159}
\definecolor{c_255_238_238}{RGB}{255,238,238}
\definecolor{c_255_18_18}{RGB}{255,18,18}
\definecolor{c_255_132_132}{RGB}{255,132,132}
\definecolor{c_255_28_28}{RGB}{255,28,28}
\definecolor{c_255_36_36}{RGB}{255,36,36}
\definecolor{c_255_135_135}{RGB}{255,135,135}
\definecolor{c_255_200_200}{RGB}{255,200,200}
\definecolor{c_255_231_231}{RGB}{255,231,231}
\definecolor{c_255_70_70}{RGB}{255,70,70}
\definecolor{c_255_130_130}{RGB}{255,130,130}
\definecolor{c_255_214_214}{RGB}{255,214,214}
\definecolor{c_255_127_127}{RGB}{255,127,127}
\definecolor{c_255_234_234}{RGB}{255,234,234}
\definecolor{c_255_102_102}{RGB}{255,102,102}
\definecolor{c_255_230_230}{RGB}{255,230,230}
\definecolor{c_255_65_65}{RGB}{255,65,65}
\definecolor{c_255_151_151}{RGB}{255,151,151}
\definecolor{c_255_203_203}{RGB}{255,203,203}
\definecolor{c_255_211_211}{RGB}{255,211,211}
\definecolor{c_255_200_200}{RGB}{255,200,200}
\definecolor{c_255_66_66}{RGB}{255,66,66}
\definecolor{c_255_201_201}{RGB}{255,201,201}
\definecolor{c_255_16_16}{RGB}{255,16,16}
\definecolor{c_255_244_244}{RGB}{255,244,244}
\definecolor{c_255_220_220}{RGB}{255,220,220}
\definecolor{c_255_13_13}{RGB}{255,13,13}
\definecolor{c_255_8_8}{RGB}{255,8,8}
\definecolor{c_255_142_142}{RGB}{255,142,142}
\definecolor{c_255_205_205}{RGB}{255,205,205}
\definecolor{c_255_120_120}{RGB}{255,120,120}
\definecolor{c_255_126_126}{RGB}{255,126,126}
\definecolor{c_255_151_151}{RGB}{255,151,151}
\definecolor{c_255_174_174}{RGB}{255,174,174}
\definecolor{c_255_59_59}{RGB}{255,59,59}
\definecolor{c_255_174_174}{RGB}{255,174,174}
\definecolor{c_255_201_201}{RGB}{255,201,201}
\definecolor{c_255_233_233}{RGB}{255,233,233}
\definecolor{c_255_97_97}{RGB}{255,97,97}
\definecolor{c_255_209_209}{RGB}{255,209,209}

\begin{table*}
\centering

\ra{1.3} 
\caption{Comments of the participants of the German study. Participants were asked to rate the \underline{underlined} word or symbol.}\label{tab:participant_comments_de}
\includegraphics{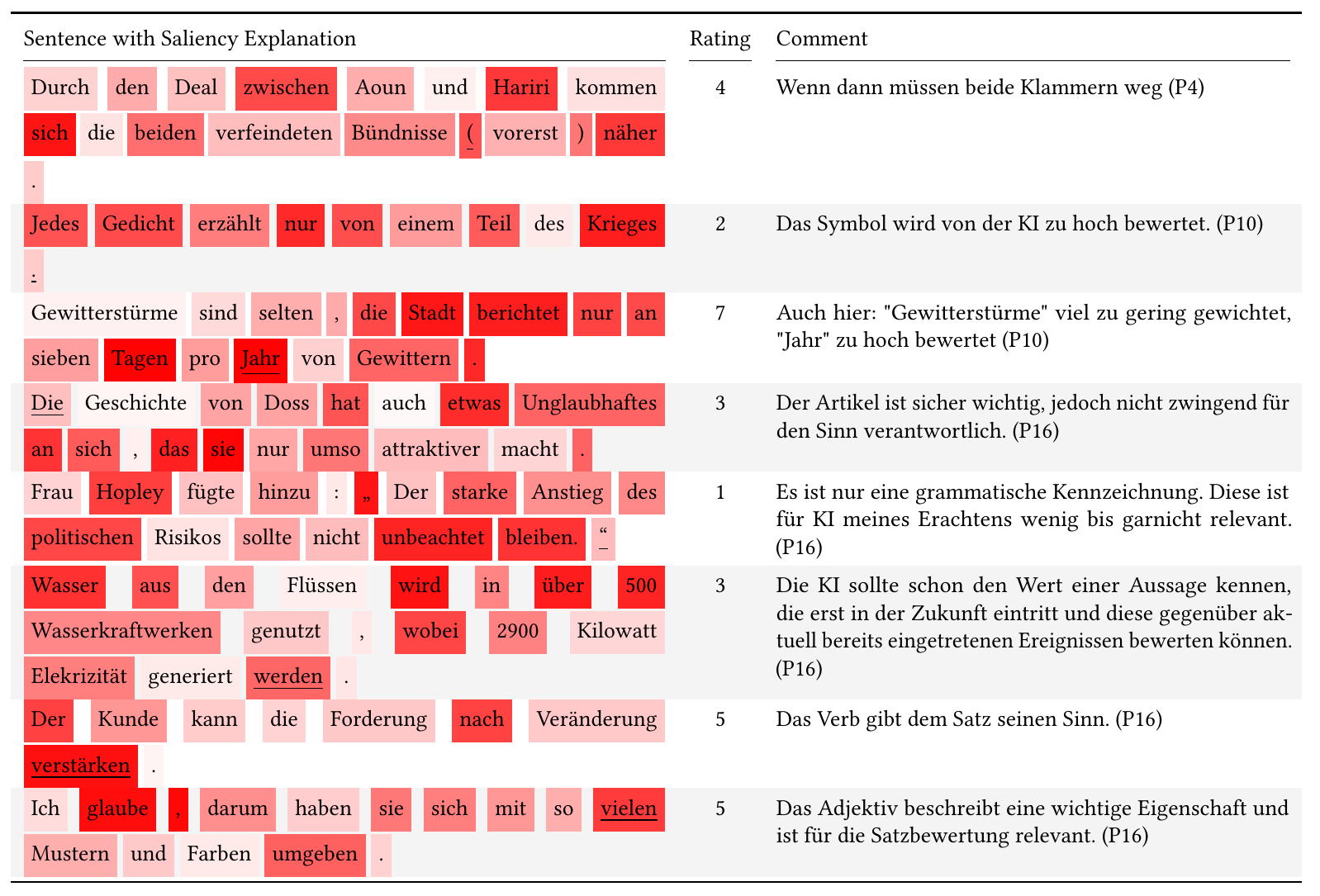}
\end{table*}
\section{German Study Details}\label{sec:appendix_de}
In this section, we provide details on the analysis of the German experiment.
Table~\ref{tab:tests_smooth_de} and Table~\ref{tab:tests_parametric_de} display test statistics for the smooth and parametric terms of the fitted GAMM model.
Table~\ref{tab:coefficients_details_de} shows statistics regarding parametric coefficient estimates.
Cut points are located at -1, 0.86, 2.42, 3.75, 5.53 and 7.67.
Table~\ref{tab:participant_comments_de} lists exemplary participant comments.
\clearpage

\section{Model-Based Bias Correction}\label{sec:saliency-correction-algorithm}
Our second approach to bias mitigation is to leverage the previously described GAMM model of human saliency perception and to \textit{correct} saliency perception by superimposing the initial saliency values with a correction signal.

Concretely, we want to increase the saliency scores for words which are predicted to be under-perceived (e.g., short words and words that appear in long sentences) and decrease the saliency scores for words which are predicted to be over-perceived (e.g., word's with a high polarity score or words that appear in very short sentences). 

When we want to \textit{correct} a user perception via the saliency scores, we cannot say whether a subjective user rating of importance is right or wrong.
However, the previously described GAMM model allows us to map a combination of a saliency score together with word/sentence properties to a perceived importance score (on a continuous latent scale).
In the following, we denote this mapping as:
\begin{equation}
    u(s, \vec{x}): [0,1] \times \mathbb{R}^d \rightarrow \mathbb{R}\label{eq:def_u},
\end{equation}
where $s$ is a saliency score and $\vec{x}$ is a $d$-dimensional feature vector representing the word/sentence properties.
This function allows us to take a fixed saliency score $s$ (e.g., 0.7) and predict its perceived importance given word and sentence features $\vec{\hat{x}}$ (corresponding to, e.g., a word length of 5 characters and a sentence length of 4).
We define this predicted importance score as
\begin{equation}
    p := u\left(s, \vec{\hat{x}} \right)\label{eq:u_orig}
\end{equation}
Additionally, it allows us to predict the perceived importance of that same saliency (0.7) in a hypothetical reference context $\vec{x_{\text{ref}}}$ (corresponding to, e.g., a word length of 3 and a sentence length of 6).
We define this second predicted importance score as
\begin{equation}
    p_{\text{ref}} := u\left(s, \vec{x_{\text{ref}}} \right)\label{eq:u_ref}
\end{equation}
We can now define a \textit{bias score} $b \in \mathbb{R}$ as the difference between the the importance score for the saliency in the observed context and the importance score for the same saliency in the reference context
\begin{equation}
    b := p - p_{\text{ref}}.\label{eq:b}
\end{equation}
The predicted bias score $b$ is positive if the saliency in the observed context is \textit{over-perceived} with respect to the reference level and negative if it is \textit{under-perceived} with respect to the reference level.
A bias score of zero corresponds to an \textit{unbiased} predicted perception.
Intuitively, this formalization allows us to answer the question ``\textit{In which direction do I have to change the saliency such that the predicted bias with respect to the reference context is decreased?}''.

To gain an executable process for bias mitigation we still lack (a) way to handle the random effects in the model, i.e., participant IDs and sentence IDs, (b) a definition of the reference context and (c) a procedure to minimize the absolute value of the bias score.
We detail these three aspects in the following.

\subsection{Including Random Effects}
So far, our definition of the model function $u$ ignores the random effects of the GAMM model, i.e., we did not specify which worker ID and which sentence ID should be used in predicting the importance score.
However, the choice of the respective levels directly influences the model predictions not only via a the random intercepts but also via the random slopes for each worker and sentence ID.
We see two options to address this.
While a first, intuitive remedy is to use an arbitrary worker ID and an arbitrary sentence ID for all predictions, this approach has the disadvantage of introducing an arbitrary bias. 
Therefore, we choose to make each model prediction not only for one participant ID and one sentence ID, but instead for all combinations of participant IDs and sentence IDs ($50 \times 150 = 7500$ combinations).
Thereby, we consider each combination of participant sentence as equally relevant for the prediction on unseen participants and sentences and smooth-out extreme influences of single participants or sentence IDs.
Formally, we thus update our definition of Equation~\ref{eq:def_u} to:
\begin{equation}
    u(s, \vec{x}, w, v): [0,1] \times \mathbb{R}^d \times W \times V  \rightarrow \mathbb{R},\label{eq:def_u_random_effects}
\end{equation}
where $W$ is the set of participant (or crowdworker) IDs ($|W| = 50$) and $V$ is the set of sentence IDs ($|V| = 150$).
Consequently, a single evaluation of $u\left(s, \vec{x} \right)$ is now replaced with
\begin{equation}
    \frac{1}{|W||V|} \sum_{w \in W} \sum_{v \in V} u\left(s, \vec{x}, w, v \right).\label{eq:u_summed}
\end{equation}

\subsection{Choosing the Reference Context}
So far, our definitions in Equations \ref{eq:u_orig} to \ref{eq:b} do not impose any constraints on the choice of reference context.
Why can we not just use an arbitrary reference context with, e.g., a word length of eight and a sentence length of one (and respective choices for all remaining covariates such as sentiment polarity etc.)?
The problem that arises for that concrete context is that the model assigns a very high importance prediction to words with eight characters within a sentence with length one.
Consequently, $p_{\text{ref}}$ will be larger than $p$ for most words and the bias score $b$ would get negative, indicating an under-perception for most words.
If we then increase all these words' saliency scores in order to minimize the absolute bias score, we, overall, have to make large changes to the saliencies.
In other words, this specific reference contexts corresponds to an, overall, raised level of saliency intensities.
While this is not bad per-se, we favor a reference context that is as neutral as possible regarding its impact on predicted importance ratings.

In order to find such a reference context, we sample 10001 random points from the space of possible contexts defined as the cross product of intervals of observed values (e.g., 1-37 characters word length) per variable if the variable is numeric (e.g., word length) and the set of possible values if the variable is categorial (e.g., dependency relation).
Each point is a candidate context.
We evaluate the term in Equation~\ref{eq:u_summed} for a saliency score of $0.5$ and each candidate context.
Among all predicted importance scores, we select the median score and choose the corresponding candidate context as our reference context $\vec{x_{\text{ref}}}$.\footnote{The concrete $\vec{x_{\text{ref}}}$ corresponds to a ``flat'' dependency relation, a ``first letter capitalized'' capitalization, a display index of 129.7, a word length of 24.6, a sentence length of 4.1, a relative word frequency of 0.04, a sentiment polarity of -0.78, a normalized saliency rank of 0.11 and a word position index of 1.08. While non-integer values for, e.g. word length cannot occur in any prediction, this does not limit the utility of $\vec{x_{\text{ref}}}$ as the reference context as it only serves as an arbitraty, but neutral reference point.}

\subsection{Iterative Bias Minimization}
In order to minimize the absolute predicted bias score, we have to modify each word's original saliency score $s_{\text{orig}}^{(i)} \in [0,1]$ into a corrected saliency score $s_{\text{corr}}^{(i)} \in [0,1]$.
While this seems to be a straight-forward minimization at first glance there is one covariate in the model that complicates optimization.
The value of the saliency rank variable depends on the saliencies of all words in the sentence.
Thus, changing one word's saliency can impact all other word's saliency rank.
We therefore propose an iterative minimization that (i) sequentially picks a token in the sentence (one after the other, round-robin) and (ii) updates this token's saliency score into the direction of a decreased absolute bias score while keeping all other tokens' saliencies fixed.
Algorithm~\ref{alg:correction} shows the complete correction procedure, Table~\ref{tab:correction_detailed} shows the procedure's impact on an example sentence over the course of 100 optimization steps.
Besides the examples shown in Table~\ref{tab:correction_examples_main}, we provide additional examples in Table~\ref{tab:correction_examples_appendix}.

\begin{algorithm}
\SetAlgoLined
\KwIn{$s_{\text{orig}}^{(i)}$: Original saliency scores for each word of the sentence with length $l$.}
\KwIn{$\vec{x_{\text{ref}}}$: Feature representation of the reference input.}
\KwOut{$s_{\text{corr}}^{(i)}$: Corrected saliency scores for each word of the sentence.}
$s_{\text{corr}}^{(i)} \gets s_{\text{orig}}^{(i)}$ for all $i$. \tcp{Initialization}
\tcp{Iterate for a fixed number of steps}
\For{$k\gets1$ \KwTo $n_{\text{steps}}$}{
    \tcp{Each iteration goes over all tokens in the sentence}
    \For{$i\gets1$ \KwTo $l$}{
        $\vec{\hat{x}} \gets$ feature representation of the $i$-th word (also depends on all other $s_{\text{corr}}^{(i)}$ via the saliency rank feature)\\
        $p \gets  \frac{1}{|W||V|} \sum_{w \in W} \sum_{v \in V} u\left(s_{\text{corr}}^{(i)}, \vec{\hat{x}}, w, v \right)$
        \tcp{Model-predicted perceived importance (on the latent continuous scale) averaged over participant IDs $W$ and sentence IDs $V$.}
        $p_{\text{ref}} \gets  \frac{1}{|W||V|} \sum_{w \in W} \sum_{v \in V} u\left(s_{\text{orig}}^{(i)}, \vec{x_{\text{ref}}}, w, v \right)$
        \tcp{Model-predicted perceived importance if the word would be the reference level word (in the reference level sentence).}
        $b \gets p - p_{\text{ref}}$
        \tcp{Define bias.}
        $s_{\text{corr}}^{(i)} \gets s_{\text{corr}}^{(i)} - \alpha \cdot \left(1-\frac{k-1}{n_{\text{steps}}}\right)^2 \cdot  \text{sgn}(b)$
        \tcp{Update saliency with quadratically-decaying step size (starting from $\alpha$) into the direction of reduced predicted bias.}
        $s_{\text{corr}}^{(i)} \gets \max\left(0, \min\left(s_{\text{corr}}^{(i)}, 1\right)\right)$
        \tcp{Make sure we stay within $[0,1]$.}
    }
}
\Return $s_{\text{corr}}^{(i)}$ for all $i$.
\caption{Saliency color correction procedure.}\label{alg:correction}
\end{algorithm}

\definecolor{c_255_241_241}{RGB}{255,241,241}
\definecolor{c_255_132_132}{RGB}{255,132,132}
\definecolor{c_255_247_247}{RGB}{255,247,247}
\definecolor{c_255_250_250}{RGB}{255,250,250}
\definecolor{c_134_134_255}{RGB}{134,134,255}
\definecolor{c_255_117_117}{RGB}{255,117,117}
\definecolor{c_182_182_255}{RGB}{182,182,255}
\definecolor{c_0_0_255}{RGB}{0,0,255}
\definecolor{c_255_229_229}{RGB}{255,229,229}
\definecolor{c_255_153_153}{RGB}{255,153,153}
\definecolor{c_255_240_240}{RGB}{255,240,240}
\definecolor{c_255_233_233}{RGB}{255,233,233}
\definecolor{c_255_253_253}{RGB}{255,253,253}
\definecolor{c_255_173_173}{RGB}{255,173,173}
\definecolor{c_255_239_239}{RGB}{255,239,239}
\definecolor{c_242_242_255}{RGB}{242,242,255}
\definecolor{c_255_231_231}{RGB}{255,231,231}
\definecolor{c_255_174_174}{RGB}{255,174,174}
\definecolor{c_255_241_241}{RGB}{255,241,241}
\definecolor{c_255_231_231}{RGB}{255,231,231}
\definecolor{c_204_204_255}{RGB}{204,204,255}
\definecolor{c_255_231_231}{RGB}{255,231,231}
\definecolor{c_248_248_255}{RGB}{248,248,255}
\definecolor{c_255_202_202}{RGB}{255,202,202}
\definecolor{c_255_230_230}{RGB}{255,230,230}
\definecolor{c_255_183_183}{RGB}{255,183,183}
\definecolor{c_255_241_241}{RGB}{255,241,241}
\definecolor{c_255_231_231}{RGB}{255,231,231}
\definecolor{c_247_247_255}{RGB}{247,247,255}
\definecolor{c_255_254_254}{RGB}{255,254,254}
\definecolor{c_252_252_255}{RGB}{252,252,255}
\definecolor{c_255_254_254}{RGB}{255,254,254}
\definecolor{c_255_229_229}{RGB}{255,229,229}
\definecolor{c_255_183_183}{RGB}{255,183,183}
\definecolor{c_255_241_241}{RGB}{255,241,241}
\definecolor{c_255_231_231}{RGB}{255,231,231}
\definecolor{c_255_251_251}{RGB}{255,251,251}
\definecolor{c_255_254_254}{RGB}{255,254,254}
\definecolor{c_254_254_255}{RGB}{254,254,255}
\definecolor{c_255_254_254}{RGB}{255,254,254}
\definecolor{c_255_230_230}{RGB}{255,230,230}
\definecolor{c_255_183_183}{RGB}{255,183,183}
\definecolor{c_255_241_241}{RGB}{255,241,241}
\definecolor{c_255_231_231}{RGB}{255,231,231}
\definecolor{c_254_254_255}{RGB}{254,254,255}
\definecolor{c_255_254_254}{RGB}{255,254,254}
\definecolor{c_254_254_255}{RGB}{254,254,255}
\definecolor{c_255_254_254}{RGB}{255,254,254}
\definecolor{c_255_230_230}{RGB}{255,230,230}
\definecolor{c_255_183_183}{RGB}{255,183,183}
\definecolor{c_255_241_241}{RGB}{255,241,241}
\definecolor{c_255_231_231}{RGB}{255,231,231}
\definecolor{c_254_254_255}{RGB}{254,254,255}
\definecolor{c_255_254_254}{RGB}{255,254,254}
\definecolor{c_255_254_254}{RGB}{255,254,254}
\definecolor{c_254_254_255}{RGB}{254,254,255}

\begin{table*}
\centering
 \ra{1.3}
 \caption{Evolution of saliency scores and corresponding bias estimates across 100 optimization steps of our bias correction procedure. The first row corresponds to the initial saliency scores. The first row of the right column shows that our method predicts that the word ``thanks'' is perceived as overly important, while the other parts of the sentence (especially ``...'') are under-perceived.
After 100 optimization steps, the saliencies of ``many'', ``2scompany'' and ``...'' have been increased while the saliency of ``thanks'' is decreased resulting in a removal of nearly all predicted bias.}\label{tab:correction_detailed}
\includegraphics{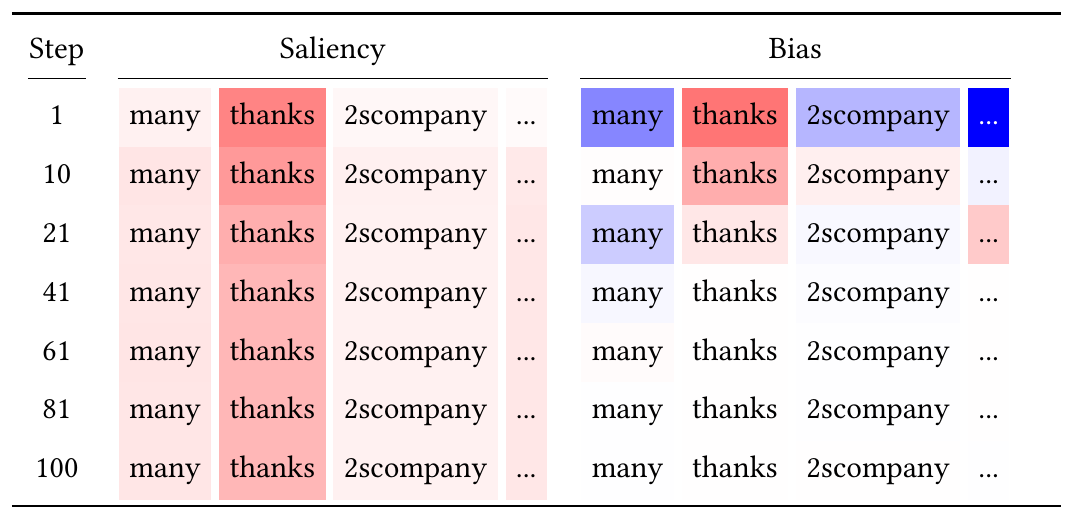}
\end{table*}

\definecolor{c_255_45_45}{RGB}{255,45,45}
\definecolor{c_255_209_209}{RGB}{255,209,209}
\definecolor{c_255_112_112}{RGB}{255,112,112}
\definecolor{c_255_209_209}{RGB}{255,209,209}
\definecolor{c_255_0_0}{RGB}{255,0,0}
\definecolor{c_255_254_254}{RGB}{255,254,254}
\definecolor{c_255_254_254}{RGB}{255,254,254}
\definecolor{c_254_254_255}{RGB}{254,254,255}
\definecolor{c_255_236_236}{RGB}{255,236,236}
\definecolor{c_255_226_226}{RGB}{255,226,226}
\definecolor{c_255_254_254}{RGB}{255,254,254}
\definecolor{c_255_221_221}{RGB}{255,221,221}
\definecolor{c_255_93_93}{RGB}{255,93,93}
\definecolor{c_255_132_132}{RGB}{255,132,132}
\definecolor{c_255_171_171}{RGB}{255,171,171}
\definecolor{c_255_146_146}{RGB}{255,146,146}
\definecolor{c_255_197_197}{RGB}{255,197,197}
\definecolor{c_255_146_146}{RGB}{255,146,146}
\definecolor{c_255_140_140}{RGB}{255,140,140}
\definecolor{c_255_126_126}{RGB}{255,126,126}
\definecolor{c_76_76_255}{RGB}{76,76,255}
\definecolor{c_60_60_255}{RGB}{60,60,255}
\definecolor{c_0_0_255}{RGB}{0,0,255}
\definecolor{c_114_114_255}{RGB}{114,114,255}
\definecolor{c_255_144_144}{RGB}{255,144,144}
\definecolor{c_186_186_255}{RGB}{186,186,255}
\definecolor{c_255_254_254}{RGB}{255,254,254}
\definecolor{c_255_253_253}{RGB}{255,253,253}
\definecolor{c_254_254_255}{RGB}{254,254,255}
\definecolor{c_212_212_255}{RGB}{212,212,255}
\definecolor{c_255_254_254}{RGB}{255,254,254}
\definecolor{c_255_254_254}{RGB}{255,254,254}
\definecolor{c_255_179_179}{RGB}{255,179,179}
\definecolor{c_255_228_228}{RGB}{255,228,228}
\definecolor{c_255_226_226}{RGB}{255,226,226}
\definecolor{c_255_192_192}{RGB}{255,192,192}
\definecolor{c_255_186_186}{RGB}{255,186,186}
\definecolor{c_255_189_189}{RGB}{255,189,189}
\definecolor{c_255_198_198}{RGB}{255,198,198}
\definecolor{c_255_186_186}{RGB}{255,186,186}
\definecolor{c_255_191_191}{RGB}{255,191,191}
\definecolor{c_0_0_255}{RGB}{0,0,255}
\definecolor{c_144_144_255}{RGB}{144,144,255}
\definecolor{c_153_153_255}{RGB}{153,153,255}
\definecolor{c_255_213_213}{RGB}{255,213,213}
\definecolor{c_255_254_254}{RGB}{255,254,254}
\definecolor{c_255_254_254}{RGB}{255,254,254}
\definecolor{c_255_252_252}{RGB}{255,252,252}
\definecolor{c_255_246_246}{RGB}{255,246,246}
\definecolor{c_255_34_34}{RGB}{255,34,34}
\definecolor{c_255_240_240}{RGB}{255,240,240}
\definecolor{c_255_243_243}{RGB}{255,243,243}
\definecolor{c_255_210_210}{RGB}{255,210,210}
\definecolor{c_255_86_86}{RGB}{255,86,86}
\definecolor{c_255_210_210}{RGB}{255,210,210}
\definecolor{c_255_209_209}{RGB}{255,209,209}
\definecolor{c_6_6_255}{RGB}{6,6,255}
\definecolor{c_255_45_45}{RGB}{255,45,45}
\definecolor{c_139_139_255}{RGB}{139,139,255}
\definecolor{c_0_0_255}{RGB}{0,0,255}
\definecolor{c_242_242_255}{RGB}{242,242,255}
\definecolor{c_255_254_254}{RGB}{255,254,254}
\definecolor{c_255_254_254}{RGB}{255,254,254}
\definecolor{c_255_254_254}{RGB}{255,254,254}
\definecolor{c_255_250_250}{RGB}{255,250,250}
\definecolor{c_255_244_244}{RGB}{255,244,244}
\definecolor{c_255_230_230}{RGB}{255,230,230}
\definecolor{c_255_204_204}{RGB}{255,204,204}
\definecolor{c_255_102_102}{RGB}{255,102,102}
\definecolor{c_255_242_242}{RGB}{255,242,242}
\definecolor{c_255_225_225}{RGB}{255,225,225}
\definecolor{c_255_179_179}{RGB}{255,179,179}
\definecolor{c_255_179_179}{RGB}{255,179,179}
\definecolor{c_255_157_157}{RGB}{255,157,157}
\definecolor{c_255_151_151}{RGB}{255,151,151}
\definecolor{c_255_158_158}{RGB}{255,158,158}
\definecolor{c_100_100_255}{RGB}{100,100,255}
\definecolor{c_42_42_255}{RGB}{42,42,255}
\definecolor{c_162_162_255}{RGB}{162,162,255}
\definecolor{c_174_174_255}{RGB}{174,174,255}
\definecolor{c_255_173_173}{RGB}{255,173,173}
\definecolor{c_0_0_255}{RGB}{0,0,255}
\definecolor{c_255_254_254}{RGB}{255,254,254}
\definecolor{c_255_221_221}{RGB}{255,221,221}
\definecolor{c_224_224_255}{RGB}{224,224,255}
\definecolor{c_255_254_254}{RGB}{255,254,254}
\definecolor{c_255_254_254}{RGB}{255,254,254}
\definecolor{c_255_254_254}{RGB}{255,254,254}
\definecolor{c_255_119_119}{RGB}{255,119,119}
\definecolor{c_255_181_181}{RGB}{255,181,181}
\definecolor{c_255_232_232}{RGB}{255,232,232}
\definecolor{c_255_230_230}{RGB}{255,230,230}
\definecolor{c_255_160_160}{RGB}{255,160,160}
\definecolor{c_255_164_164}{RGB}{255,164,164}
\definecolor{c_255_223_223}{RGB}{255,223,223}
\definecolor{c_255_222_222}{RGB}{255,222,222}
\definecolor{c_255_0_0}{RGB}{255,0,0}
\definecolor{c_157_157_255}{RGB}{157,157,255}
\definecolor{c_111_111_255}{RGB}{111,111,255}
\definecolor{c_132_132_255}{RGB}{132,132,255}
\definecolor{c_254_254_255}{RGB}{254,254,255}
\definecolor{c_254_254_255}{RGB}{254,254,255}
\definecolor{c_255_254_254}{RGB}{255,254,254}
\definecolor{c_255_254_254}{RGB}{255,254,254}
\definecolor{c_255_0_0}{RGB}{255,0,0}
\definecolor{c_255_86_86}{RGB}{255,86,86}
\definecolor{c_255_0_0}{RGB}{255,0,0}
\definecolor{c_255_194_194}{RGB}{255,194,194}
\definecolor{c_255_189_189}{RGB}{255,189,189}
\definecolor{c_255_173_173}{RGB}{255,173,173}
\definecolor{c_255_201_201}{RGB}{255,201,201}
\definecolor{c_255_201_201}{RGB}{255,201,201}
\definecolor{c_255_131_131}{RGB}{255,131,131}
\definecolor{c_255_0_0}{RGB}{255,0,0}
\definecolor{c_255_254_254}{RGB}{255,254,254}
\definecolor{c_255_254_254}{RGB}{255,254,254}
\definecolor{c_255_108_108}{RGB}{255,108,108}
\definecolor{c_255_236_236}{RGB}{255,236,236}
\definecolor{c_255_233_233}{RGB}{255,233,233}
\definecolor{c_255_194_194}{RGB}{255,194,194}
\definecolor{c_255_247_247}{RGB}{255,247,247}
\definecolor{c_255_166_166}{RGB}{255,166,166}
\definecolor{c_255_187_187}{RGB}{255,187,187}
\definecolor{c_255_171_171}{RGB}{255,171,171}
\definecolor{c_255_171_171}{RGB}{255,171,171}
\definecolor{c_255_171_171}{RGB}{255,171,171}
\definecolor{c_255_181_181}{RGB}{255,181,181}
\definecolor{c_172_172_255}{RGB}{172,172,255}
\definecolor{c_130_130_255}{RGB}{130,130,255}
\definecolor{c_231_231_255}{RGB}{231,231,255}
\definecolor{c_0_0_255}{RGB}{0,0,255}
\definecolor{c_255_254_254}{RGB}{255,254,254}
\definecolor{c_254_254_255}{RGB}{254,254,255}
\definecolor{c_232_232_255}{RGB}{232,232,255}
\definecolor{c_255_250_250}{RGB}{255,250,250}
\definecolor{c_207_207_255}{RGB}{207,207,255}
\definecolor{c_255_251_251}{RGB}{255,251,251}
\definecolor{c_255_238_238}{RGB}{255,238,238}
\definecolor{c_255_244_244}{RGB}{255,244,244}
\definecolor{c_255_41_41}{RGB}{255,41,41}
\definecolor{c_255_254_254}{RGB}{255,254,254}
\definecolor{c_255_245_245}{RGB}{255,245,245}
\definecolor{c_255_254_254}{RGB}{255,254,254}
\definecolor{c_255_189_189}{RGB}{255,189,189}
\definecolor{c_255_189_189}{RGB}{255,189,189}
\definecolor{c_255_189_189}{RGB}{255,189,189}
\definecolor{c_255_86_86}{RGB}{255,86,86}
\definecolor{c_255_189_189}{RGB}{255,189,189}
\definecolor{c_255_189_189}{RGB}{255,189,189}
\definecolor{c_255_189_189}{RGB}{255,189,189}
\definecolor{c_101_101_255}{RGB}{101,101,255}
\definecolor{c_188_188_255}{RGB}{188,188,255}
\definecolor{c_195_195_255}{RGB}{195,195,255}
\definecolor{c_255_185_185}{RGB}{255,185,185}
\definecolor{c_0_0_255}{RGB}{0,0,255}
\definecolor{c_151_151_255}{RGB}{151,151,255}
\definecolor{c_10_10_255}{RGB}{10,10,255}
\definecolor{c_229_229_255}{RGB}{229,229,255}
\definecolor{c_255_242_242}{RGB}{255,242,242}
\definecolor{c_235_235_255}{RGB}{235,235,255}
\definecolor{c_254_254_255}{RGB}{254,254,255}
\definecolor{c_132_132_255}{RGB}{132,132,255}
\definecolor{c_255_254_254}{RGB}{255,254,254}
\definecolor{c_251_251_255}{RGB}{251,251,255}
\definecolor{c_255_41_41}{RGB}{255,41,41}
\definecolor{c_255_226_226}{RGB}{255,226,226}
\definecolor{c_255_246_246}{RGB}{255,246,246}
\definecolor{c_255_250_250}{RGB}{255,250,250}
\definecolor{c_255_83_83}{RGB}{255,83,83}
\definecolor{c_255_210_210}{RGB}{255,210,210}
\definecolor{c_255_233_233}{RGB}{255,233,233}
\definecolor{c_255_210_210}{RGB}{255,210,210}
\definecolor{c_255_135_135}{RGB}{255,135,135}
\definecolor{c_216_216_255}{RGB}{216,216,255}
\definecolor{c_191_191_255}{RGB}{191,191,255}
\definecolor{c_0_0_255}{RGB}{0,0,255}
\definecolor{c_255_254_254}{RGB}{255,254,254}
\definecolor{c_233_233_255}{RGB}{233,233,255}
\definecolor{c_255_254_254}{RGB}{255,254,254}
\definecolor{c_255_225_225}{RGB}{255,225,225}
\definecolor{c_255_77_77}{RGB}{255,77,77}
\definecolor{c_255_180_180}{RGB}{255,180,180}
\definecolor{c_255_251_251}{RGB}{255,251,251}
\definecolor{c_255_122_122}{RGB}{255,122,122}
\definecolor{c_255_195_195}{RGB}{255,195,195}
\definecolor{c_255_210_210}{RGB}{255,210,210}
\definecolor{c_255_108_108}{RGB}{255,108,108}
\definecolor{c_255_214_214}{RGB}{255,214,214}
\definecolor{c_0_0_255}{RGB}{0,0,255}
\definecolor{c_254_254_255}{RGB}{254,254,255}
\definecolor{c_255_254_254}{RGB}{255,254,254}
\definecolor{c_254_254_255}{RGB}{254,254,255}
\definecolor{c_255_248_248}{RGB}{255,248,248}
\definecolor{c_255_254_254}{RGB}{255,254,254}
\definecolor{c_255_236_236}{RGB}{255,236,236}
\definecolor{c_255_231_231}{RGB}{255,231,231}
\definecolor{c_255_53_53}{RGB}{255,53,53}
\definecolor{c_255_251_251}{RGB}{255,251,251}
\definecolor{c_255_188_188}{RGB}{255,188,188}
\definecolor{c_255_188_188}{RGB}{255,188,188}
\definecolor{c_255_188_188}{RGB}{255,188,188}
\definecolor{c_255_188_188}{RGB}{255,188,188}
\definecolor{c_255_90_90}{RGB}{255,90,90}
\definecolor{c_255_188_188}{RGB}{255,188,188}
\definecolor{c_80_80_255}{RGB}{80,80,255}
\definecolor{c_36_36_255}{RGB}{36,36,255}
\definecolor{c_139_139_255}{RGB}{139,139,255}
\definecolor{c_220_220_255}{RGB}{220,220,255}
\definecolor{c_255_182_182}{RGB}{255,182,182}
\definecolor{c_0_0_255}{RGB}{0,0,255}
\definecolor{c_224_224_255}{RGB}{224,224,255}
\definecolor{c_210_210_255}{RGB}{210,210,255}
\definecolor{c_255_243_243}{RGB}{255,243,243}
\definecolor{c_243_243_255}{RGB}{243,243,255}
\definecolor{c_254_254_255}{RGB}{254,254,255}
\definecolor{c_238_238_255}{RGB}{238,238,255}
\definecolor{c_255_218_218}{RGB}{255,218,218}
\definecolor{c_255_241_241}{RGB}{255,241,241}
\definecolor{c_255_73_73}{RGB}{255,73,73}
\definecolor{c_255_231_231}{RGB}{255,231,231}
\definecolor{c_255_191_191}{RGB}{255,191,191}
\definecolor{c_255_191_191}{RGB}{255,191,191}
\definecolor{c_255_115_115}{RGB}{255,115,115}
\definecolor{c_255_154_154}{RGB}{255,154,154}
\definecolor{c_212_212_255}{RGB}{212,212,255}
\definecolor{c_7_7_255}{RGB}{7,7,255}
\definecolor{c_255_128_128}{RGB}{255,128,128}
\definecolor{c_0_0_255}{RGB}{0,0,255}
\definecolor{c_255_253_253}{RGB}{255,253,253}
\definecolor{c_208_208_255}{RGB}{208,208,255}
\definecolor{c_254_254_255}{RGB}{254,254,255}
\definecolor{c_254_254_255}{RGB}{254,254,255}
\definecolor{c_255_21_21}{RGB}{255,21,21}
\definecolor{c_255_233_233}{RGB}{255,233,233}
\definecolor{c_255_100_100}{RGB}{255,100,100}
\definecolor{c_255_195_195}{RGB}{255,195,195}
\definecolor{c_255_0_0}{RGB}{255,0,0}
\definecolor{c_153_153_255}{RGB}{153,153,255}
\definecolor{c_255_254_254}{RGB}{255,254,254}
\definecolor{c_254_254_255}{RGB}{254,254,255}

\begin{table*}
\centering
\caption{Examples of our proposed bias reduction method. The table shows sentences along with their initial saliency scores and the respective corrected saliency scores in the \textit{saliency} column. The \textit{bias} column shows the color-coded bias estimates as defined in our method. Predicted overestimations are colored in red whereas predicted underestimations are colored in blue. For each example, we scale the range of biases to use the full color spectrum in one direction. The column \textit{removed bias} lists how many percent of the initial bias were removed in the corrected saliencies.}\label{tab:correction_examples_appendix}
\includegraphics{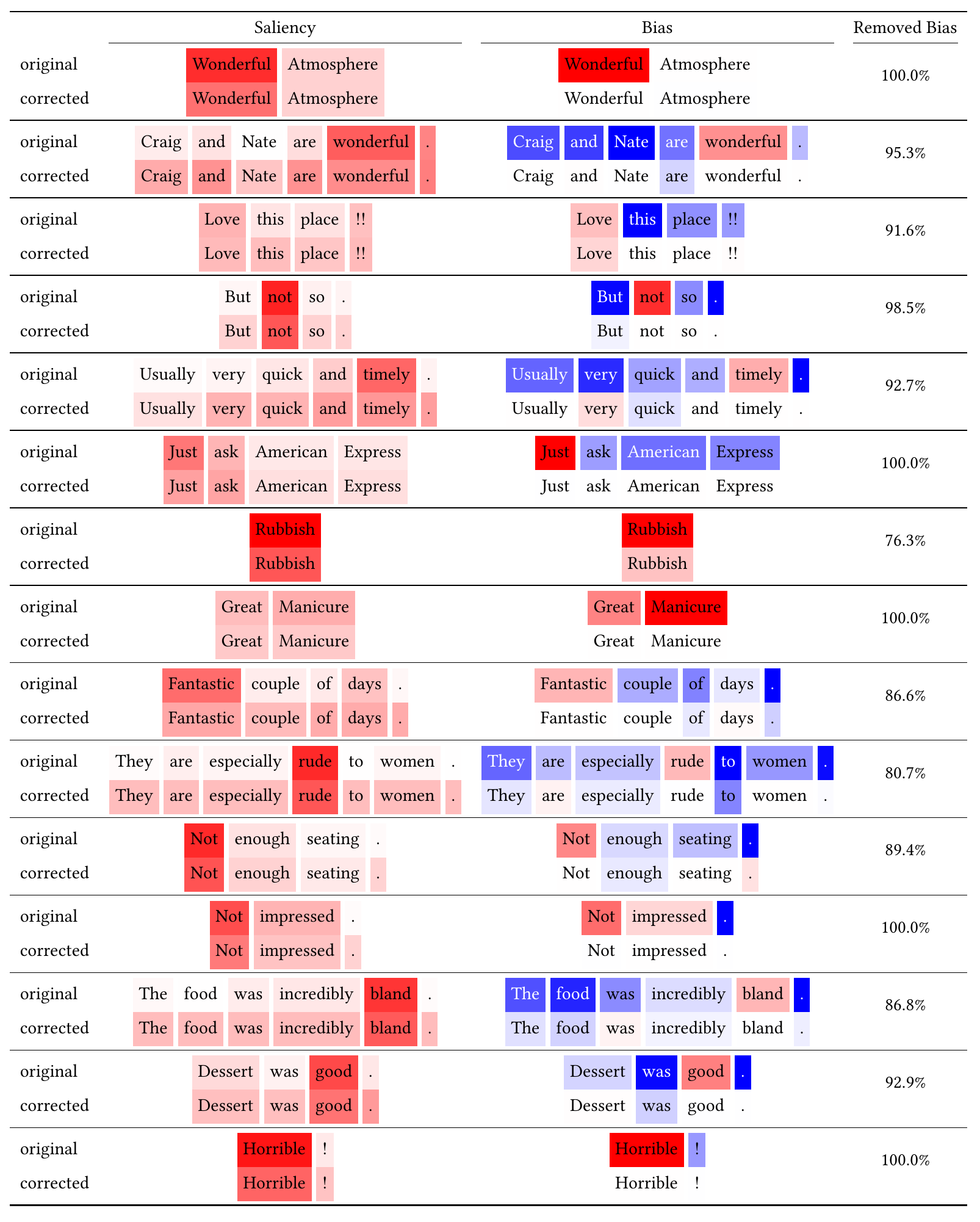}
\end{table*}

\section{Integrated Gradients and Correction Study}\label{sec:appendix_IG}
\begin{figure*}
    \centering
    \begin{subfigure}[t]{.3\textwidth}
        \centering
    \includegraphics[width=\textwidth]{figures/comparison_word_saliency_score.pdf}
    \caption{Saliency score}\label{fig:correction_comparison_saliency_score}
    \end{subfigure}%
    \hfill
    \begin{subfigure}[t]{.3\textwidth}
        \centering
    \includegraphics[width=\textwidth]{figures/comparison_num_characters.pdf}
    \caption{Word length}\label{fig:correction_comparison_word_length}
    \end{subfigure}%
    \hfill
    \begin{subfigure}[t]{.3\textwidth}
        \centering
    \includegraphics[width=\textwidth]{figures/comparison_display_index.pdf}
    \caption{Temporal display index}\label{fig:correction_comparison_display_index}
    \end{subfigure}
    \begin{subfigure}[t]{.3\textwidth}
        \centering
    \includegraphics[width=\textwidth]{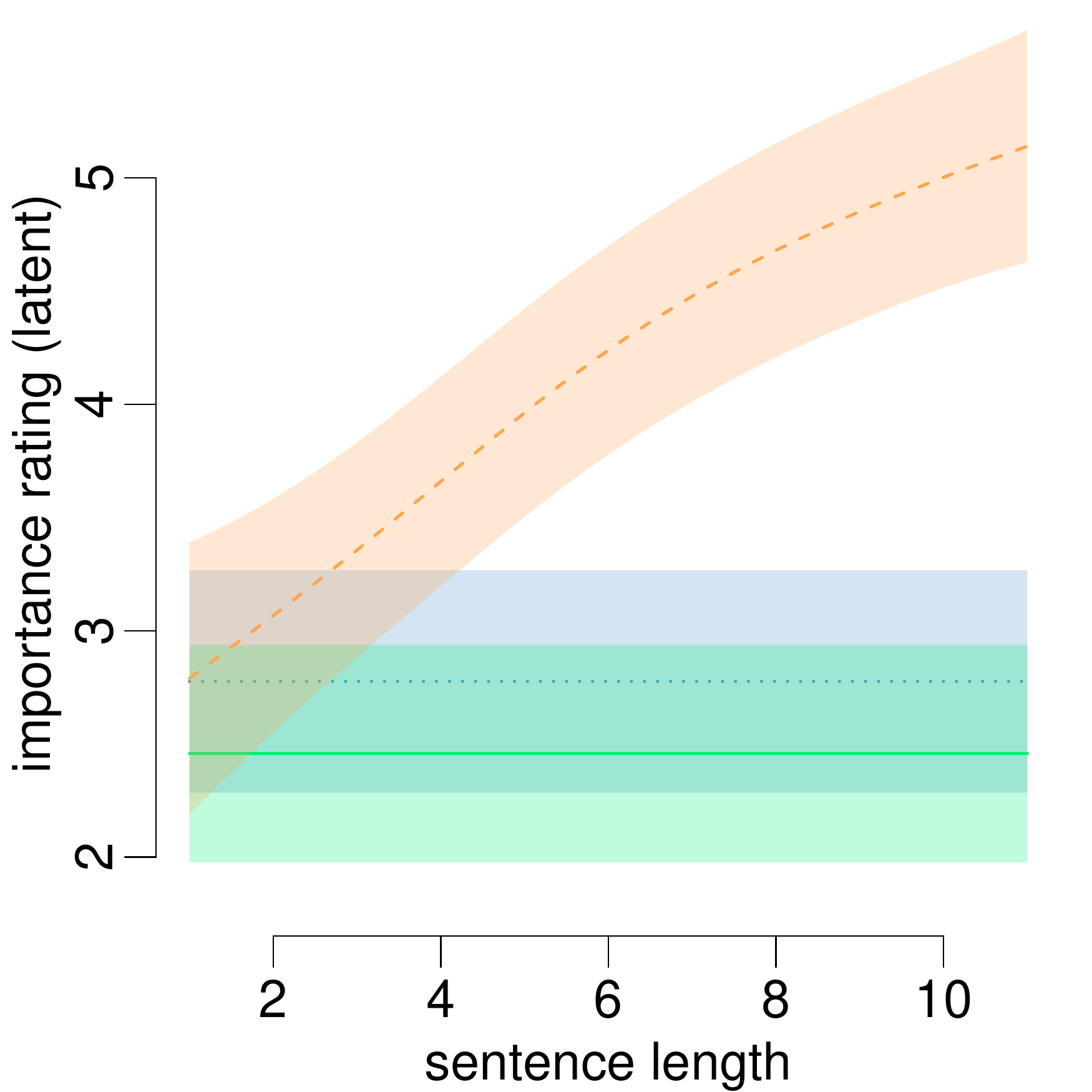}
    \caption{Sentence length}\label{fig:correction_comparison_sentence_length}
    \end{subfigure}%
    \hfill
    \begin{subfigure}[t]{.3\textwidth}
        \centering
    \includegraphics[width=\textwidth]{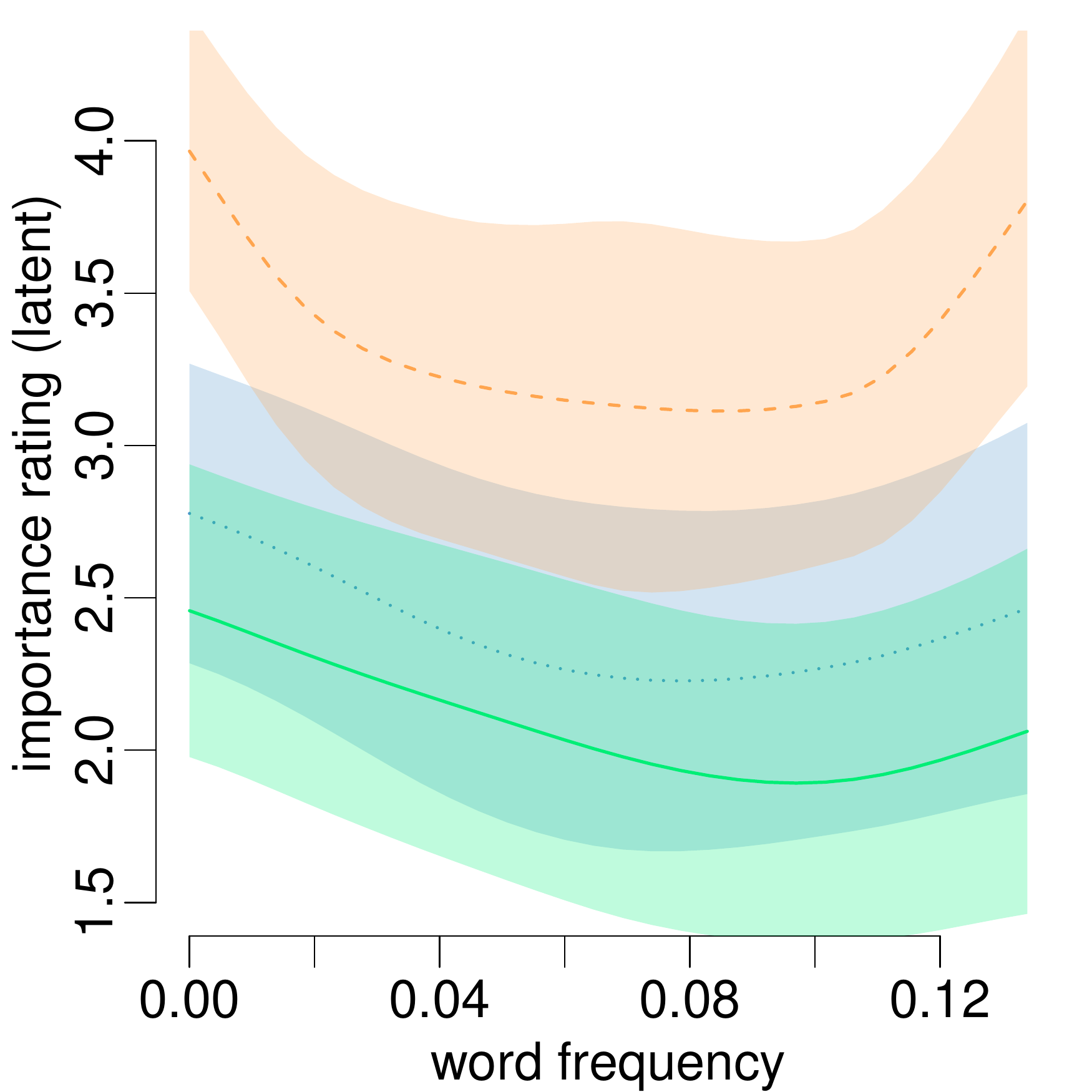}
    \caption{Word frequency}\label{fig:correction_comparison_relative_word_frequency}
    \end{subfigure}%
    \hfill
    \begin{subfigure}[t]{.3\textwidth}
        \centering
    \includegraphics[width=\textwidth]{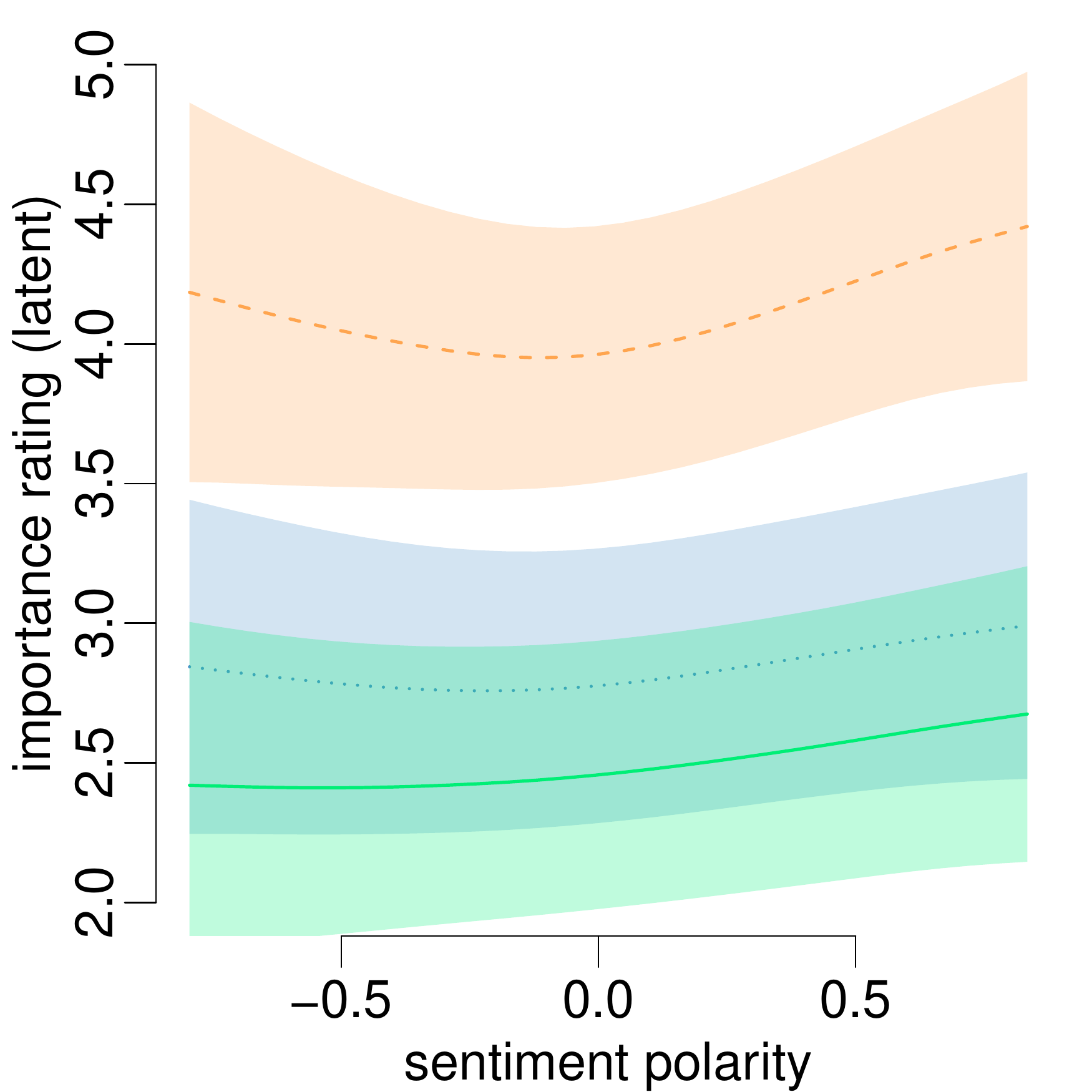}
    \caption{Sentiment polarity}\label{fig:correction_comparison_lemma_polarity}
    \end{subfigure}
    \begin{subfigure}[t]{.3\textwidth}
        \centering
    \includegraphics[width=\textwidth]{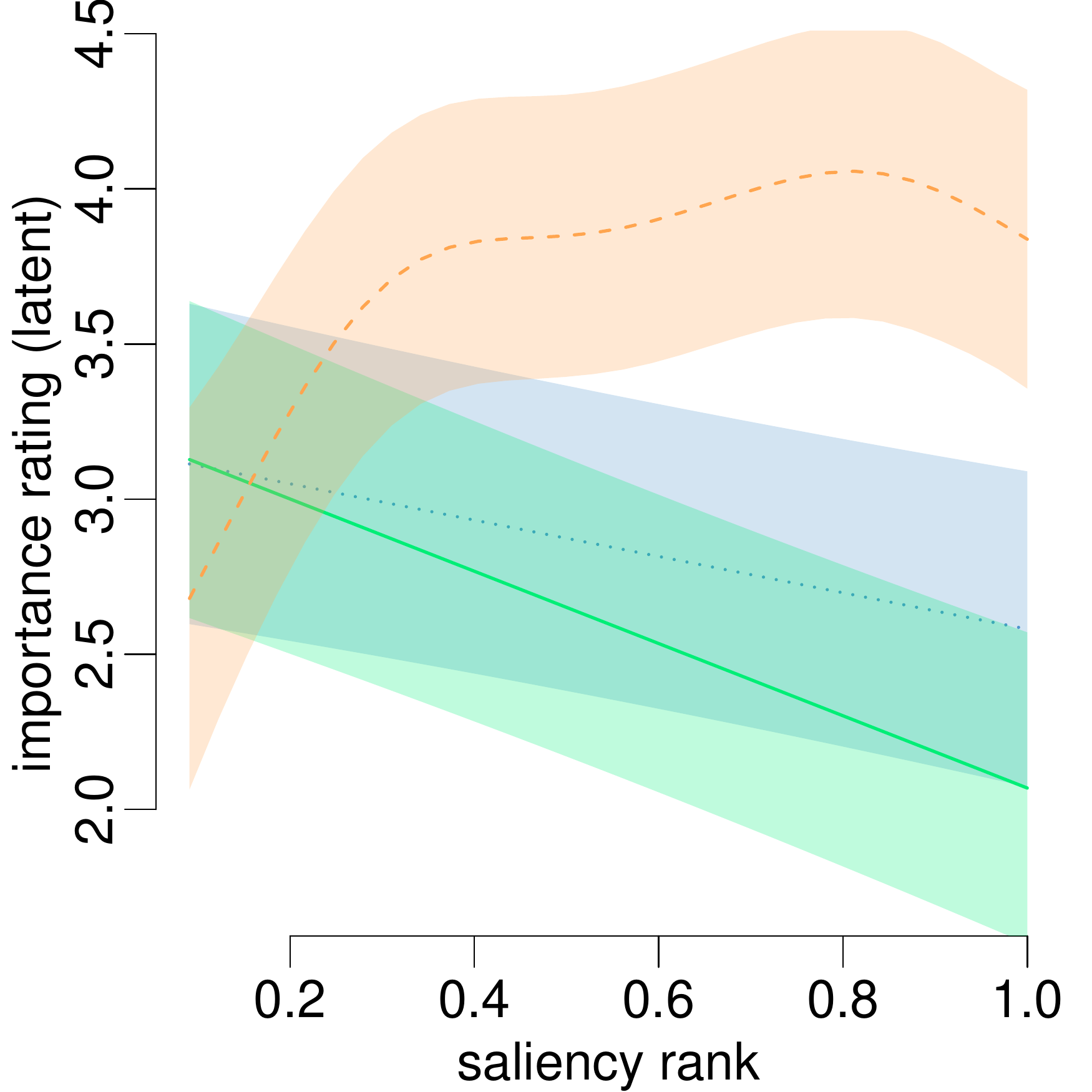}
    \caption{Saliency rank}\label{fig:correction_comparison_normalized_saliency_rank}
    \end{subfigure}%
    \hspace{1.5cm}
    \begin{subfigure}[t]{.3\textwidth}
        \centering
    \includegraphics[width=\textwidth]{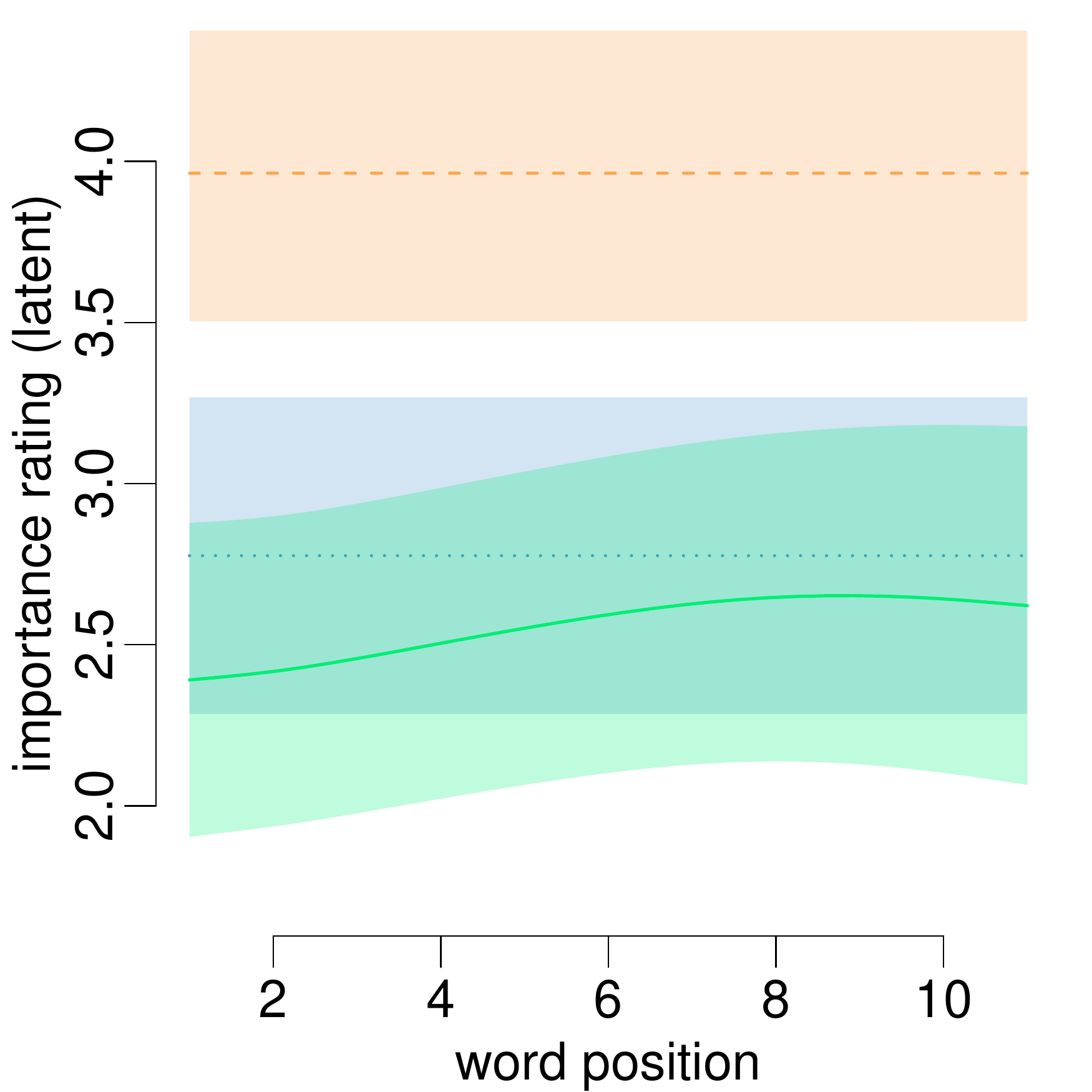}
    \caption{Word position}\label{fig:correction_comparison_word_index_in_sentence}
    \end{subfigure}
    \caption{Summed-effects comparison plots of the correction methods.}\label{fig:comparison_correction_all}
\end{figure*}

\begin{figure*}
    \centering
    \begin{subfigure}[t]{.33\textwidth}
        \centering
    \includegraphics[width=\textwidth]{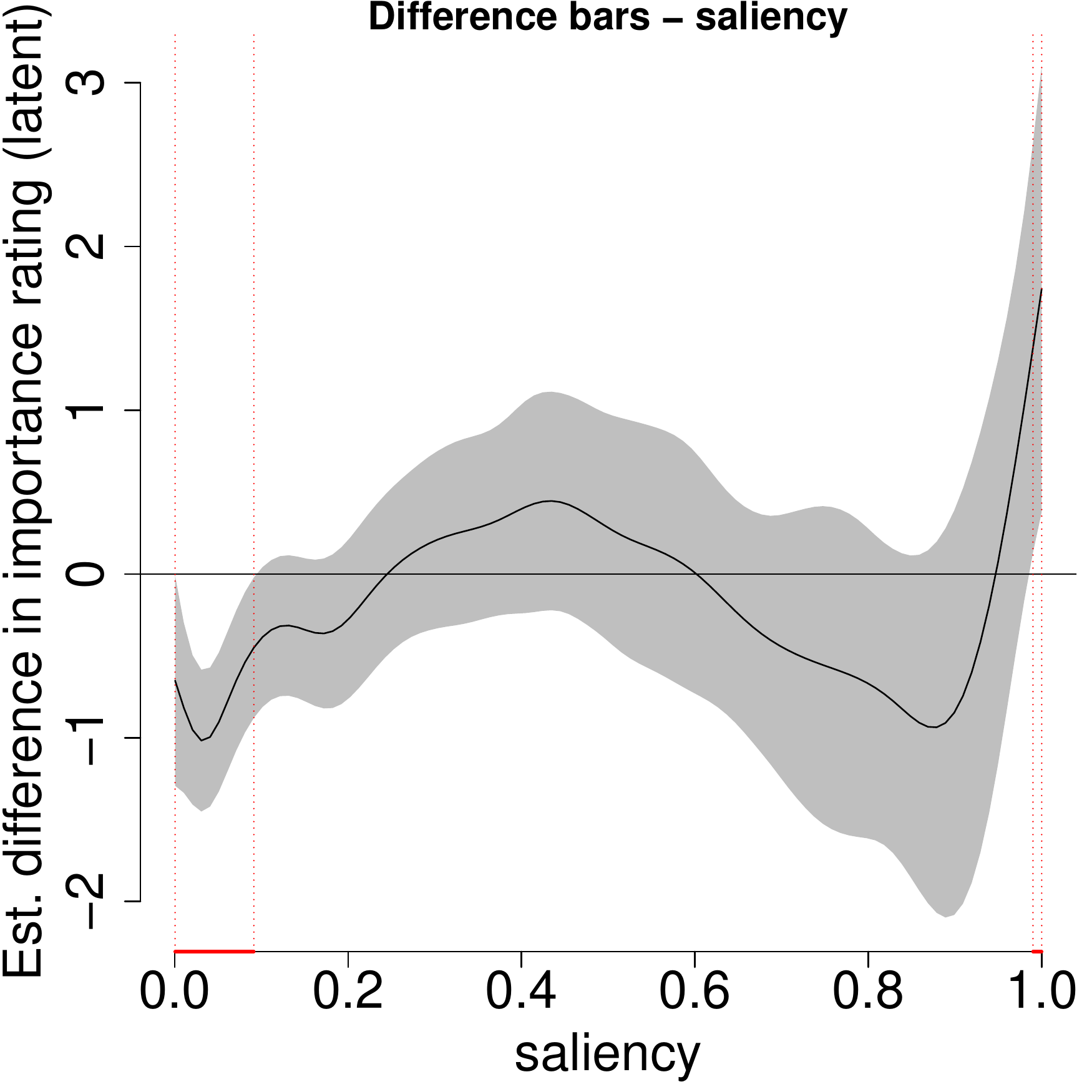}
    \caption{Saliency score}\label{fig:difference_bars_saliency_word_saliency_score}
    \end{subfigure}%
    \hfill
    \begin{subfigure}[t]{.33\textwidth}
        \centering
    \includegraphics[width=\textwidth]{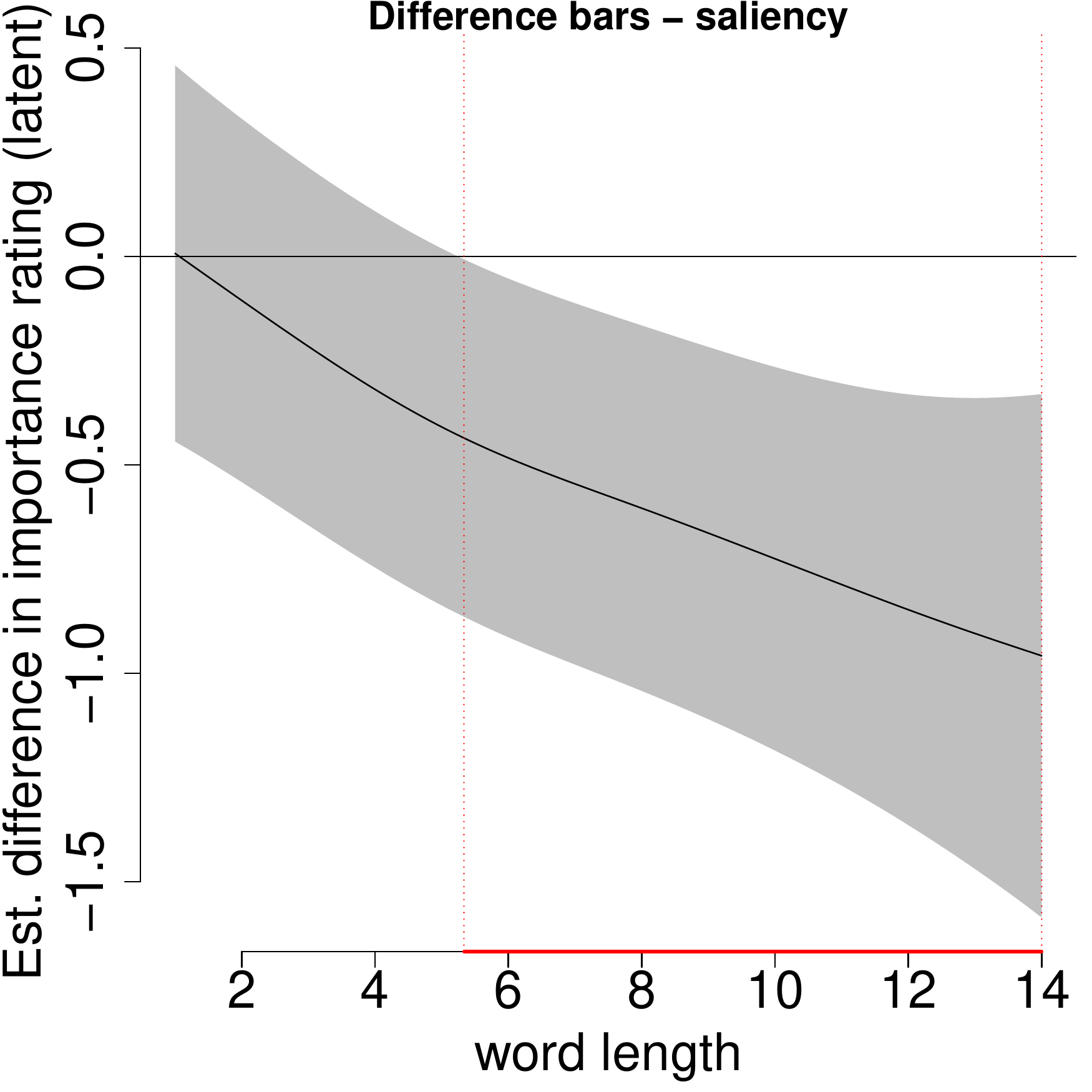}
    \caption{Word length}\label{fig:difference_bars_saliency_num_characters}
    \end{subfigure}%
    \begin{subfigure}[t]{.33\textwidth}
        \centering
    \includegraphics[width=\textwidth]{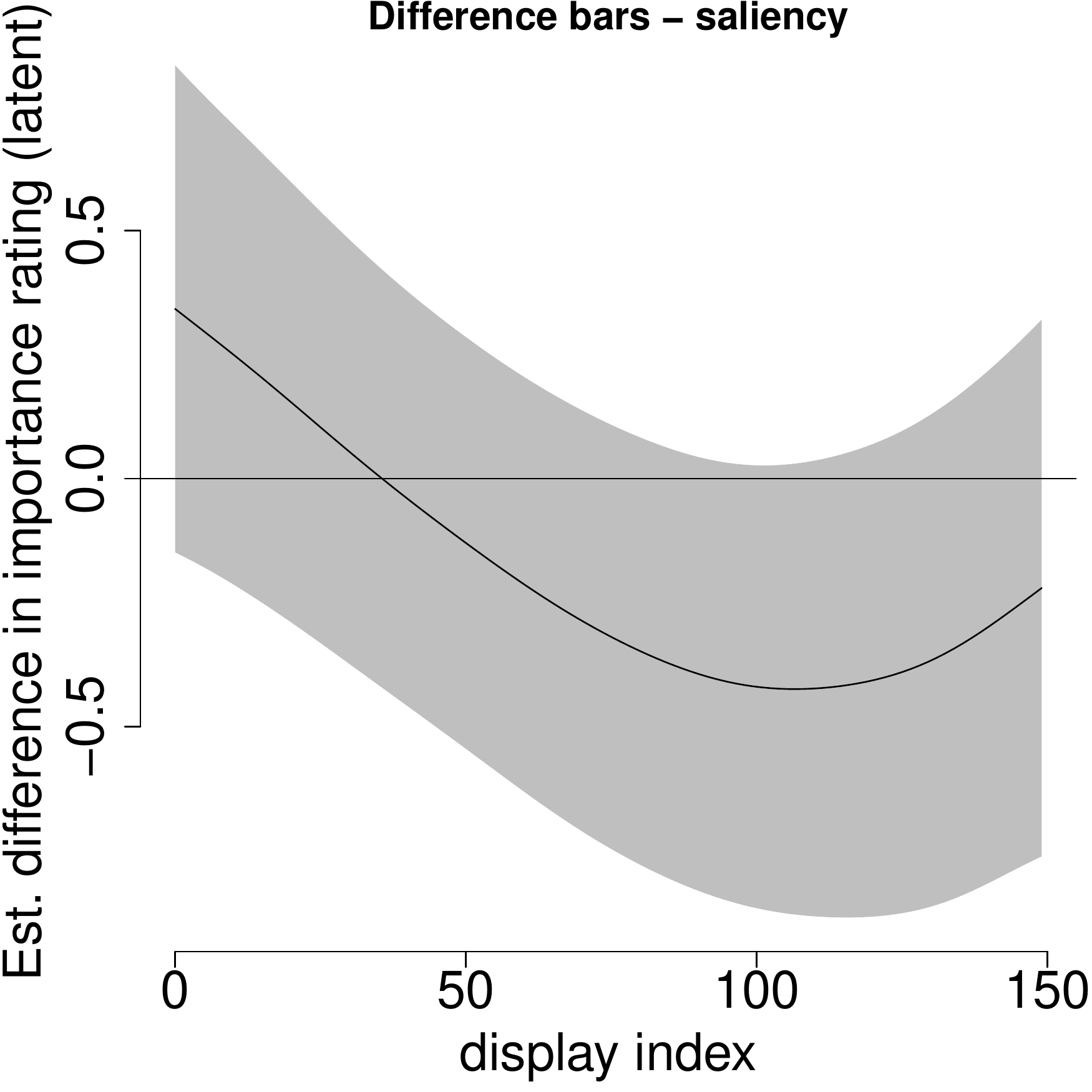}
    \caption{Temporal display index}\label{fig:difference_bars_saliency_display_index}
    \end{subfigure}
    \hfill
    \begin{subfigure}[t]{.33\textwidth}
        \centering
    \includegraphics[width=\textwidth]{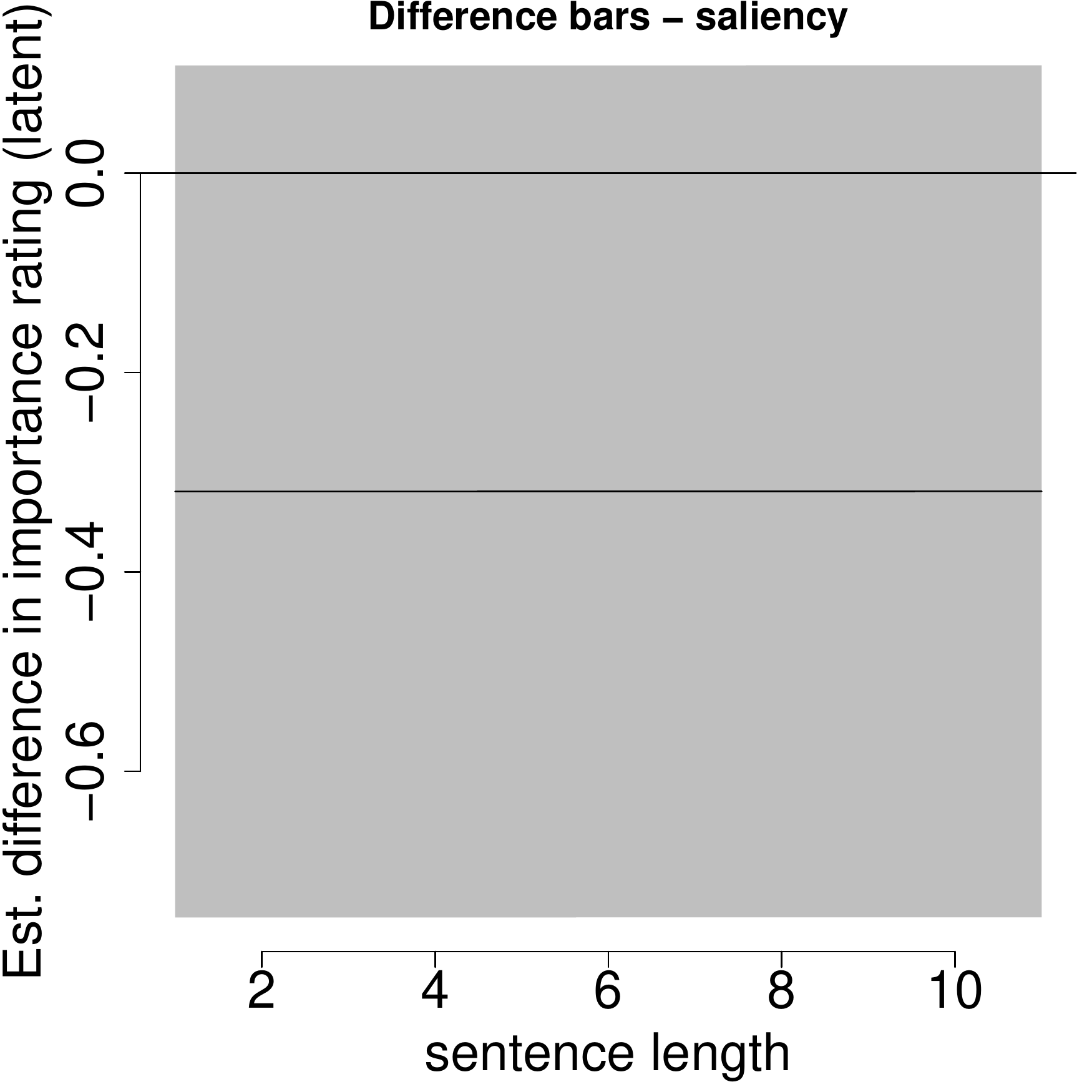}
    \caption{Sentence length}\label{fig:difference_bars_saliency_sentence_length}
    \end{subfigure}%
    \begin{subfigure}[t]{.33\textwidth}
        \centering
    \includegraphics[width=\textwidth]{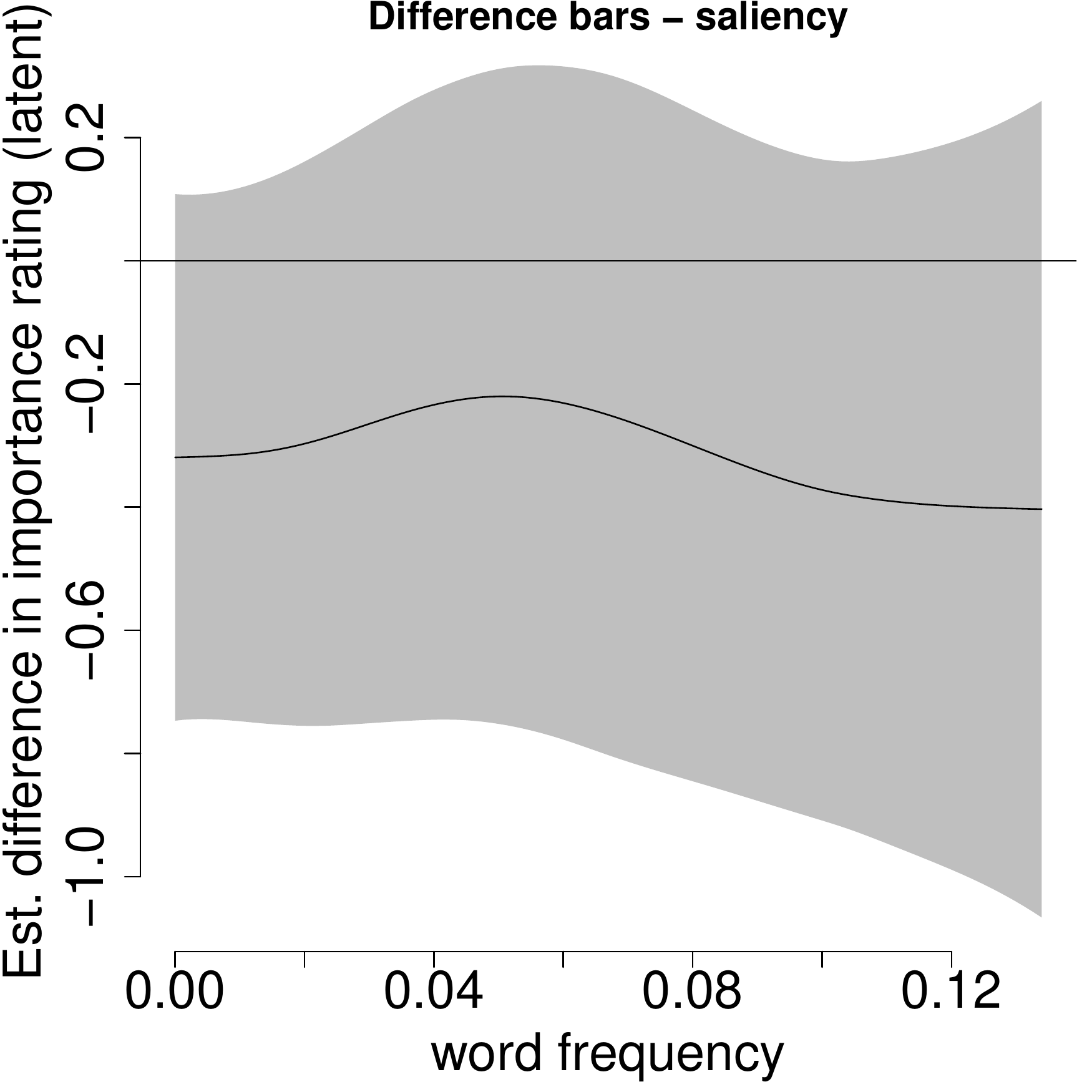}
    \caption{Word frequency}\label{fig:difference_bars_saliency_relative_word_frequency}
    \end{subfigure}%
    \begin{subfigure}[t]{.33\textwidth}
        \centering
    \includegraphics[width=\textwidth]{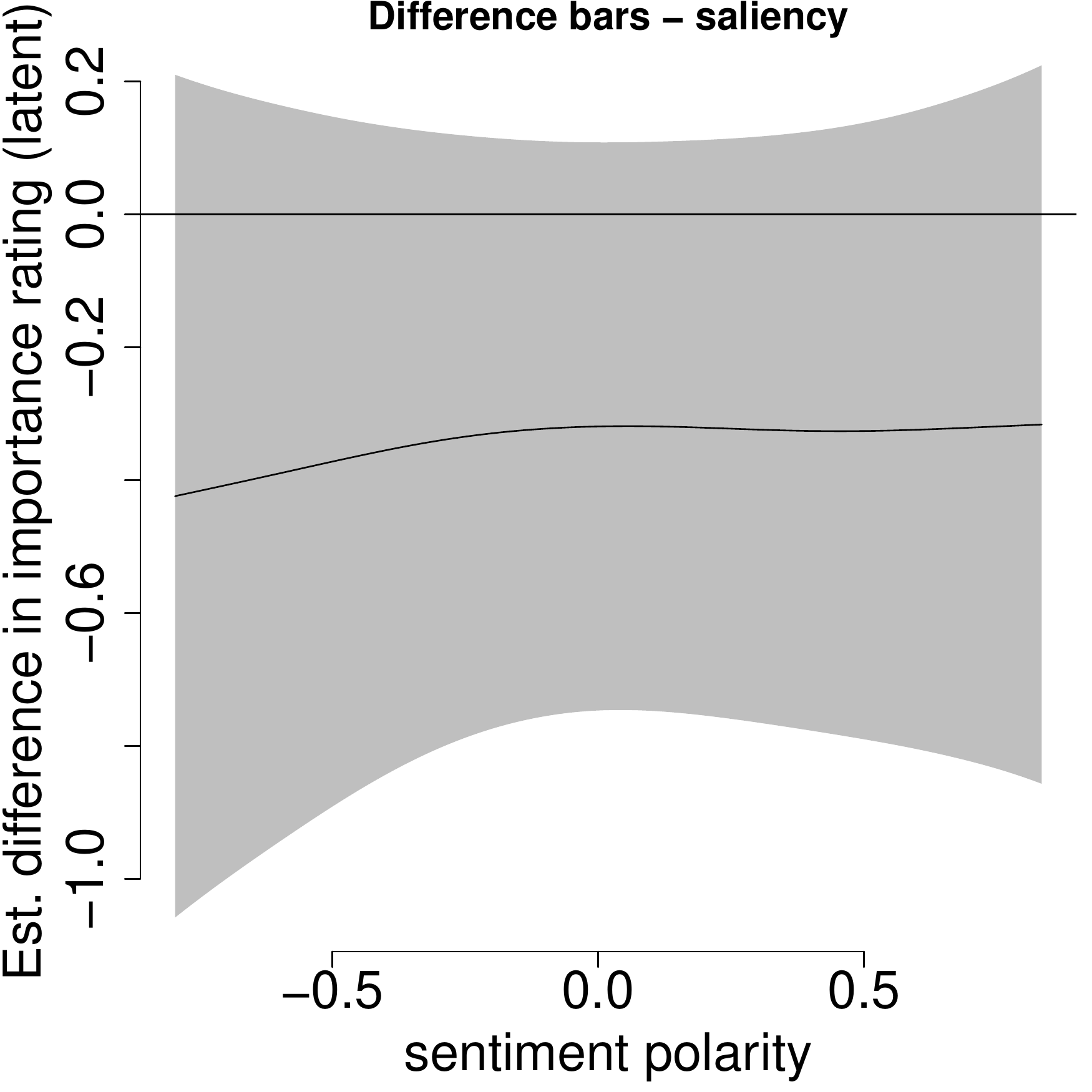}
    \caption{Sentiment polarity}\label{fig:difference_bars_saliency_lemma_polarity}
    \end{subfigure}
    \begin{subfigure}[t]{.33\textwidth}
        \centering
    \includegraphics[width=\textwidth]{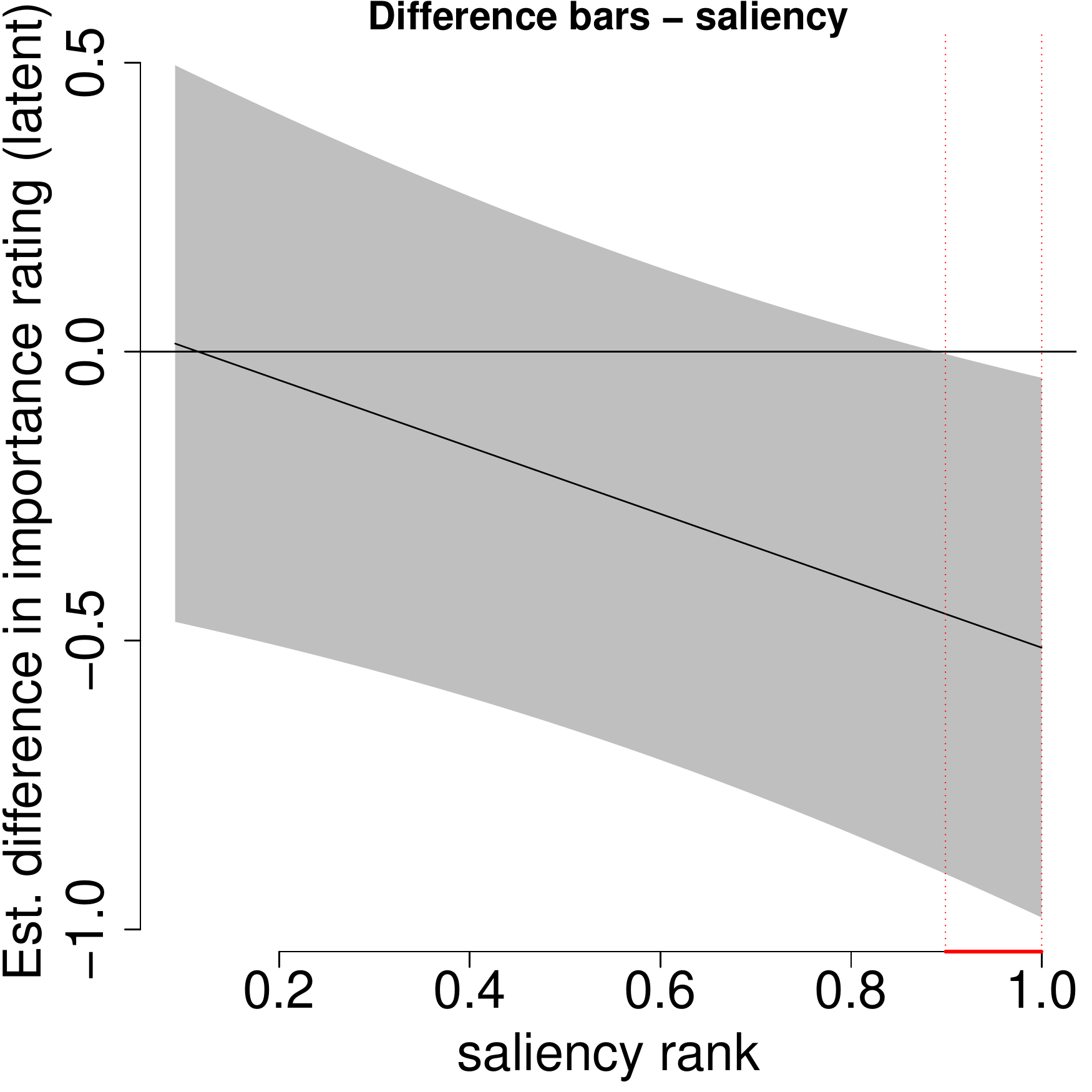}
    \caption{Saliency rank}\label{fig:difference_bars_saliency_normalized_saliency_rank}
    \end{subfigure}%
    \hspace{1cm}
    \begin{subfigure}[t]{.33\textwidth}
        \centering
    \includegraphics[width=\textwidth]{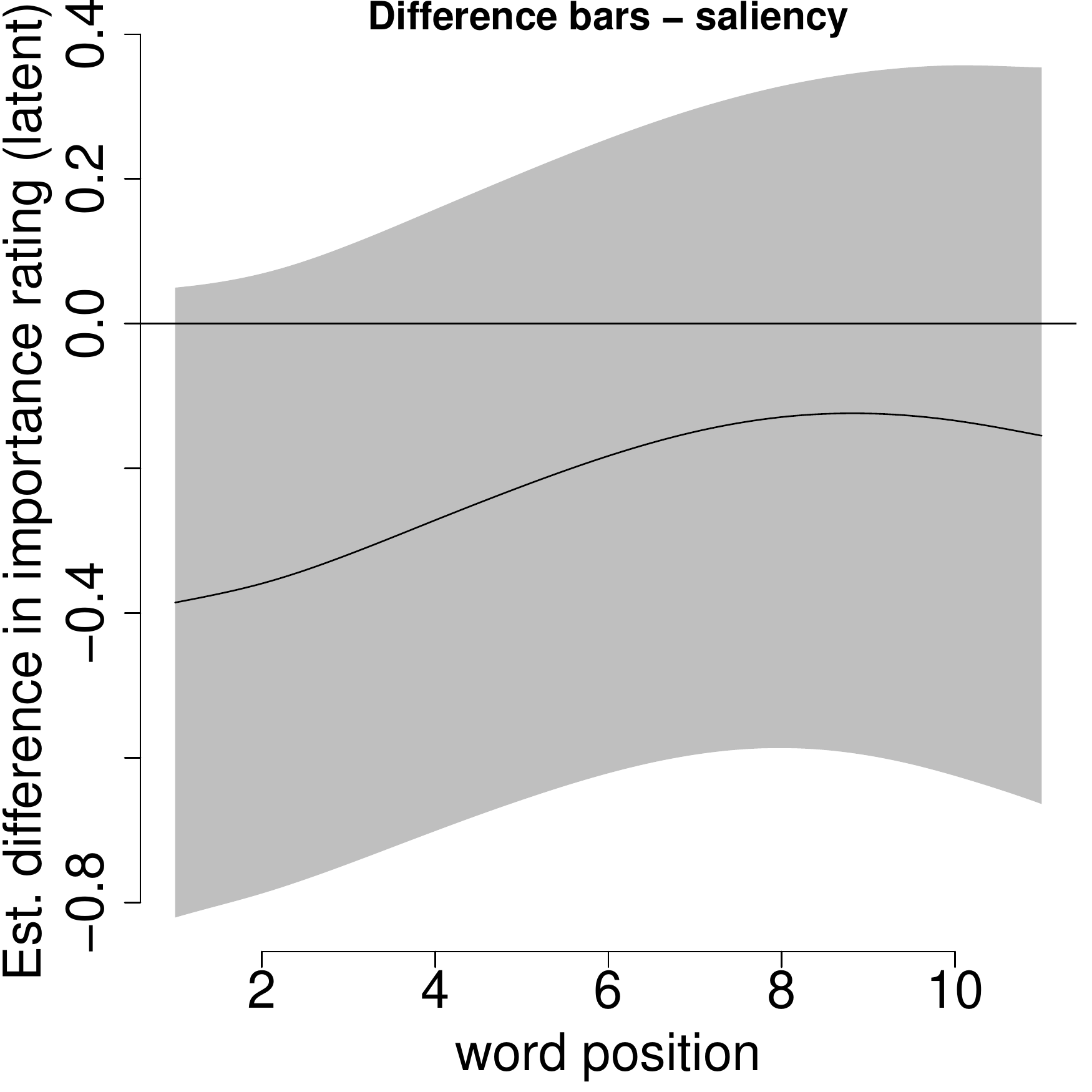}
    \caption{Word position}\label{fig:difference_bars_saliency_word_index_in_sentence}
    \end{subfigure}
    \caption{Difference plots between the bar visualization and the original visualization. Areas of significant differences are marked red.}\label{fig:differences_bars_saliency}
\end{figure*}

\begin{figure*}
    \centering
    \begin{subfigure}[t]{.33\textwidth}
        \centering
    \includegraphics[width=\textwidth]{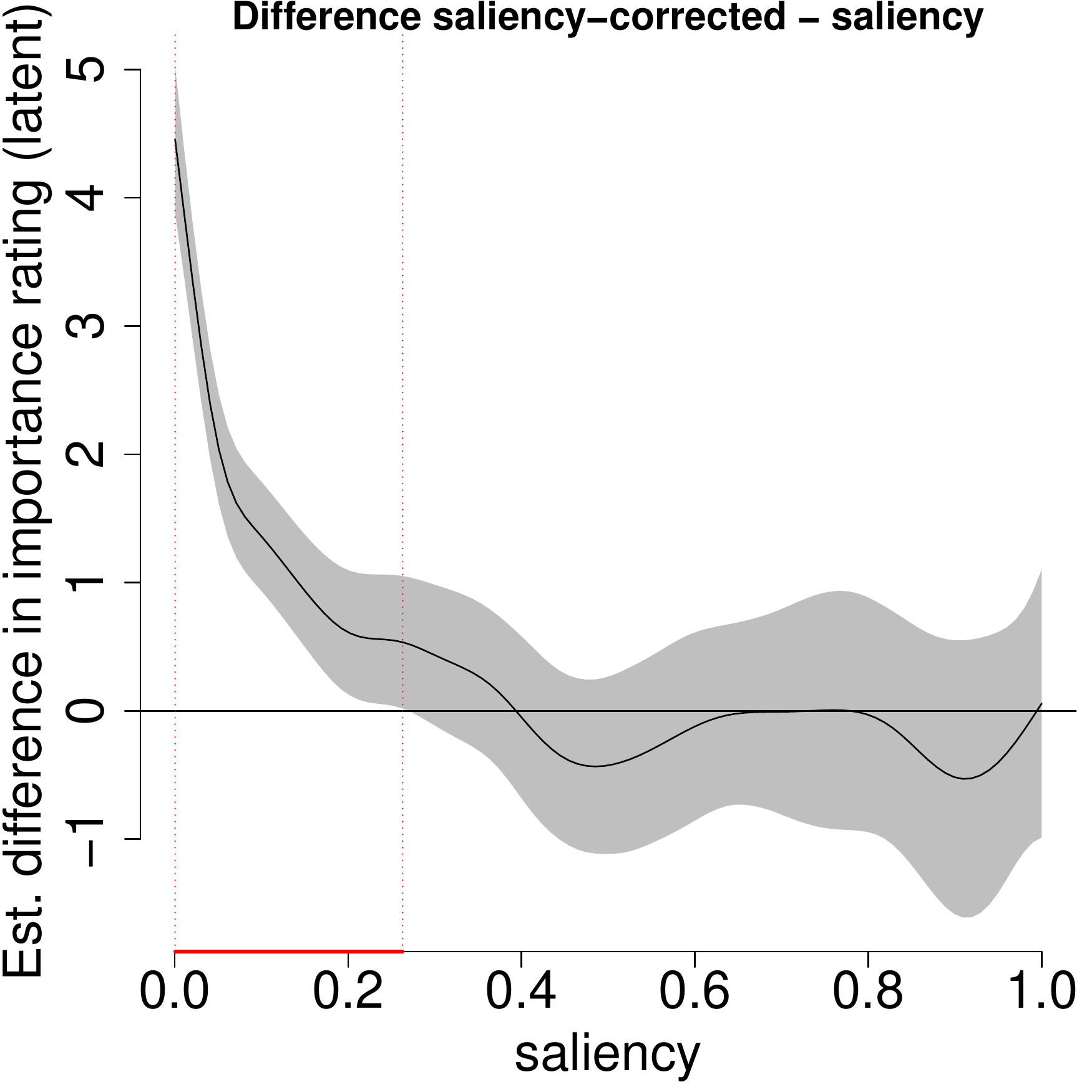}
    \caption{Saliency score}\label{fig:difference_corrected-saliency_saliency_word_saliency_score}
    \end{subfigure}%
    \hfill
    \begin{subfigure}[t]{.33\textwidth}
        \centering
    \includegraphics[width=\textwidth]{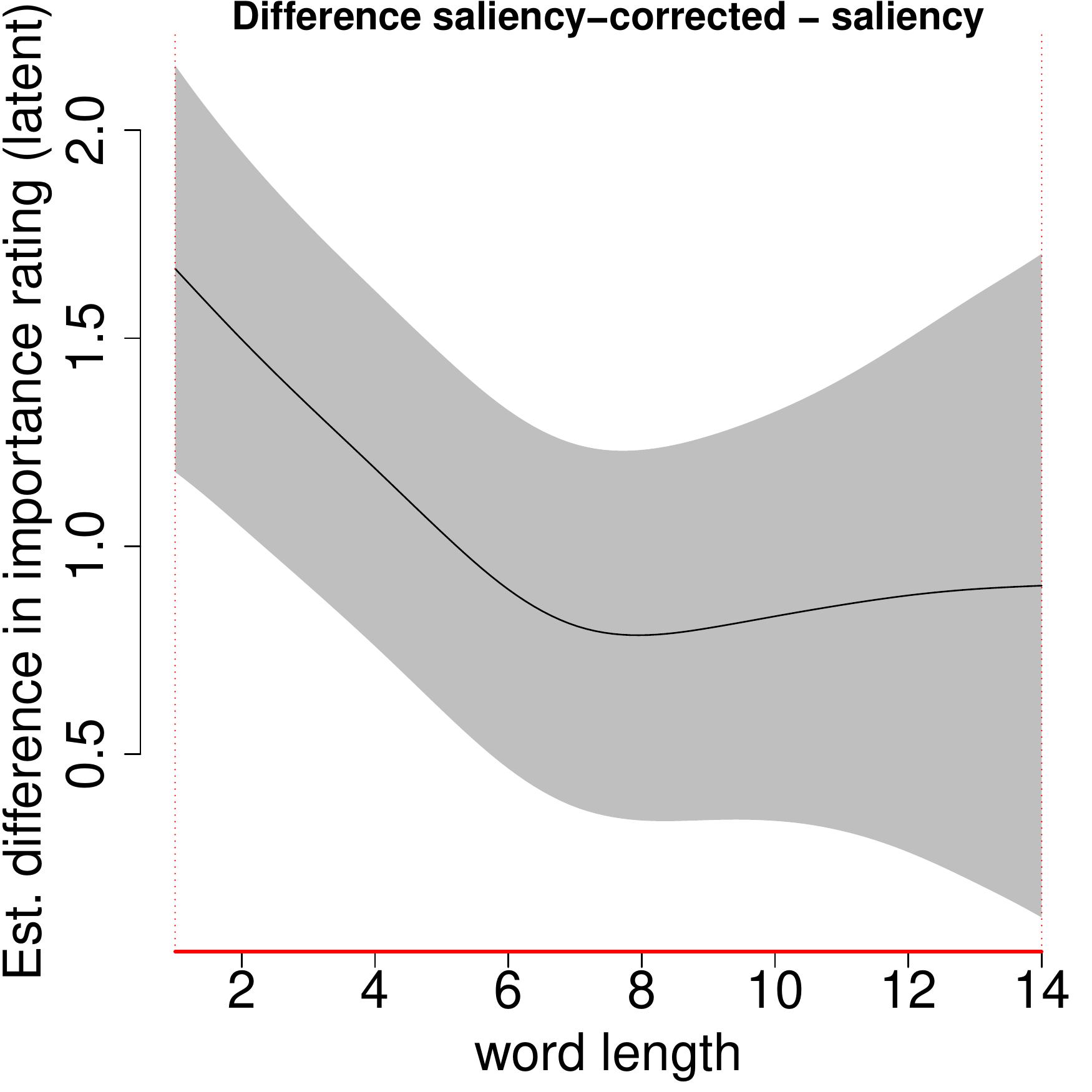}
    \caption{Word length}\label{fig:difference_corrected-saliency_saliency_num_characters}
    \end{subfigure}%
    \begin{subfigure}[t]{.33\textwidth}
        \centering
    \includegraphics[width=\textwidth]{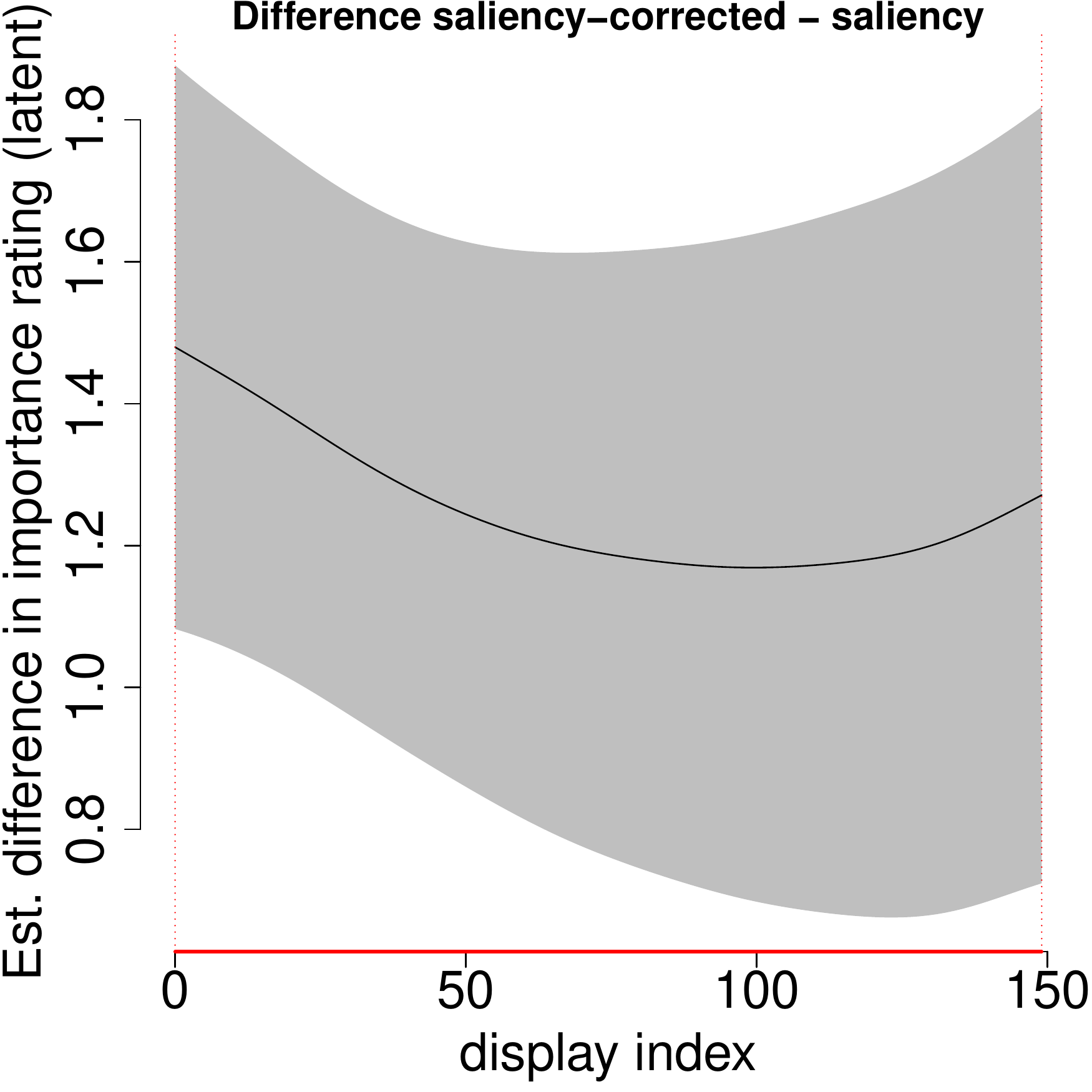}
    \caption{Temporal display index}\label{fig:difference_corrected-saliency_saliency_display_index}
    \end{subfigure}
    \hfill
    \begin{subfigure}[t]{.33\textwidth}
        \centering
    \includegraphics[width=\textwidth]{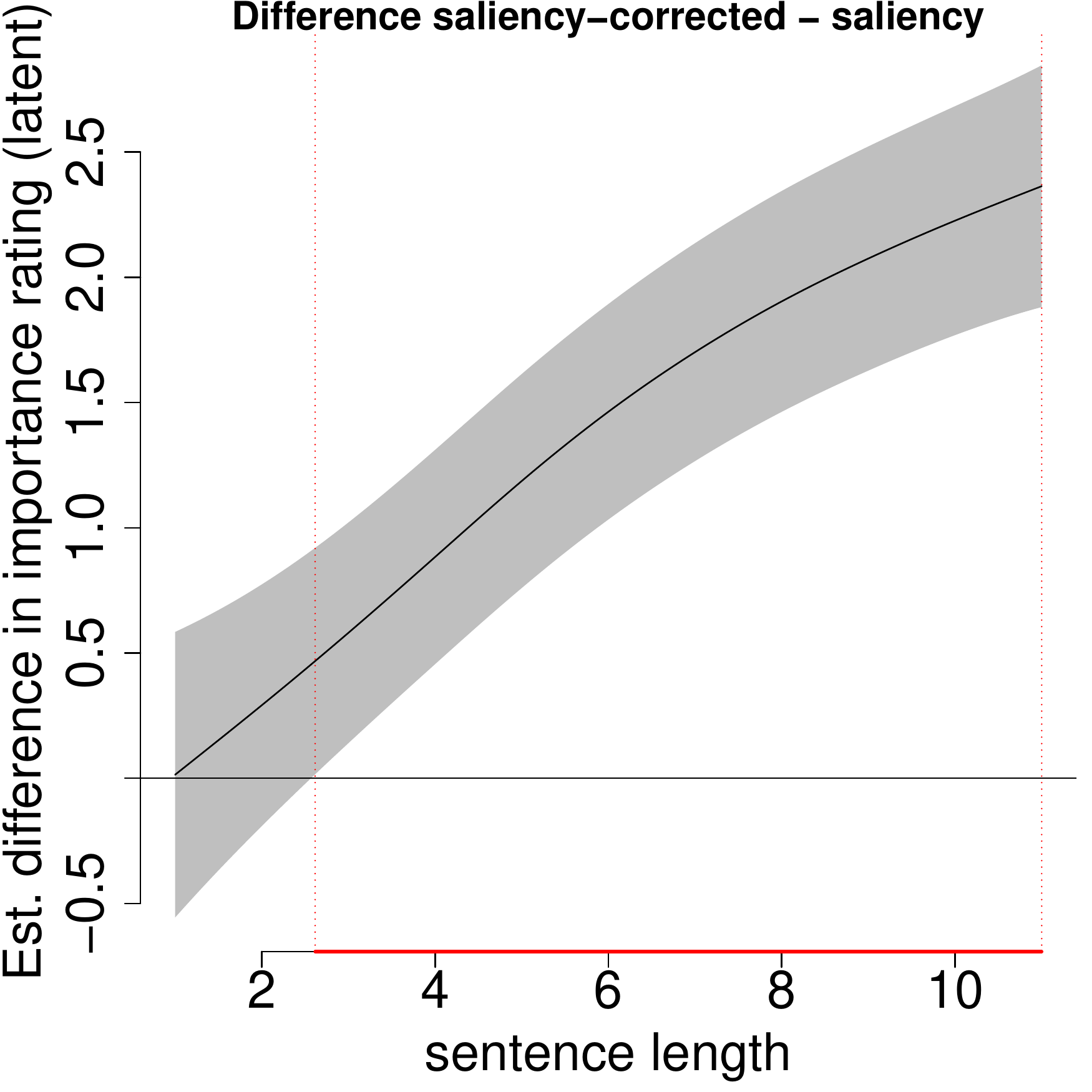}
    \caption{Sentence length}\label{fig:difference_corrected-saliency_saliency_sentence_length}
    \end{subfigure}%
    \begin{subfigure}[t]{.33\textwidth}
        \centering
    \includegraphics[width=\textwidth]{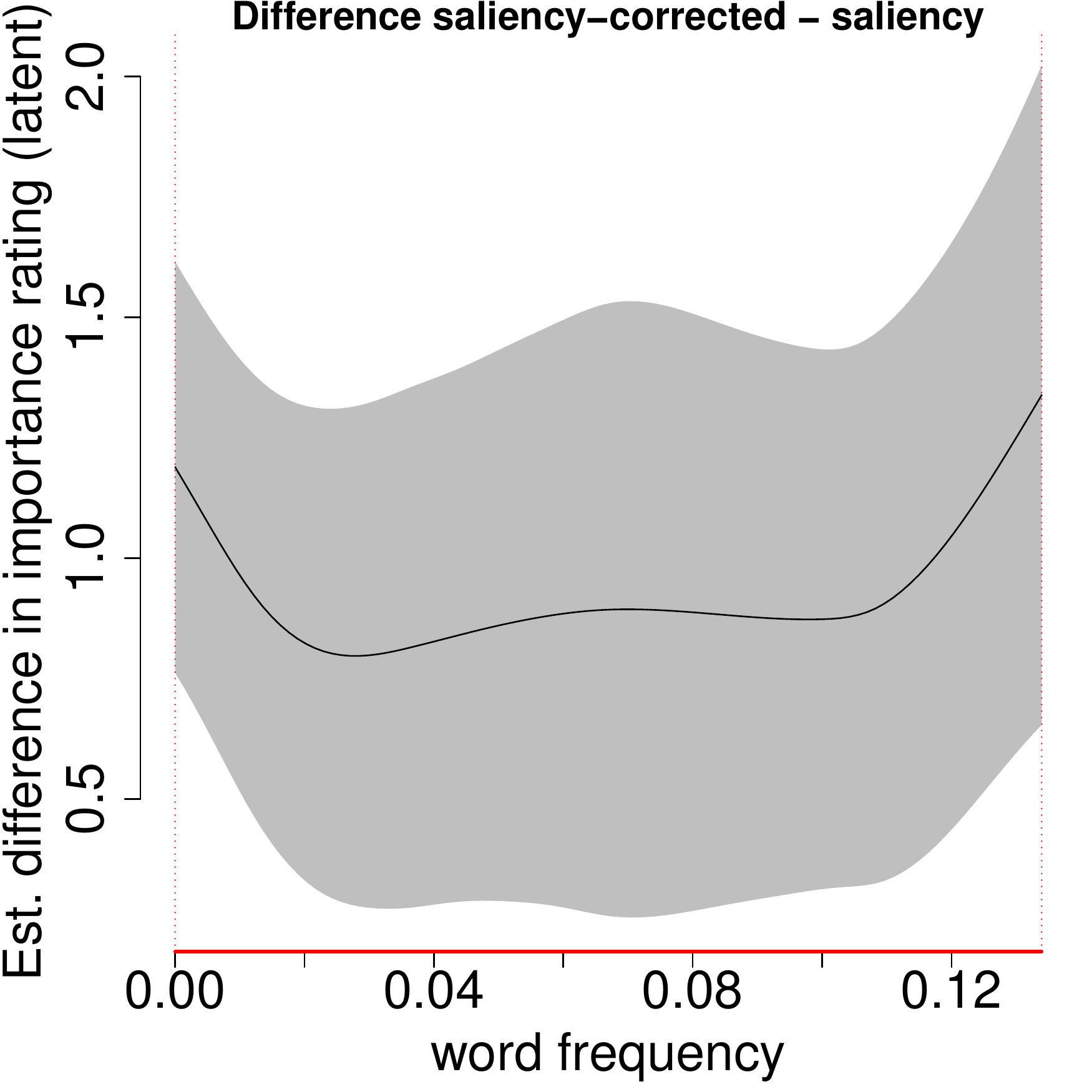}
    \caption{Word frequency}\label{fig:difference_corrected-saliency_saliency_relative_word_frequency}
    \end{subfigure}%
    \begin{subfigure}[t]{.33\textwidth}
        \centering
    \includegraphics[width=\textwidth]{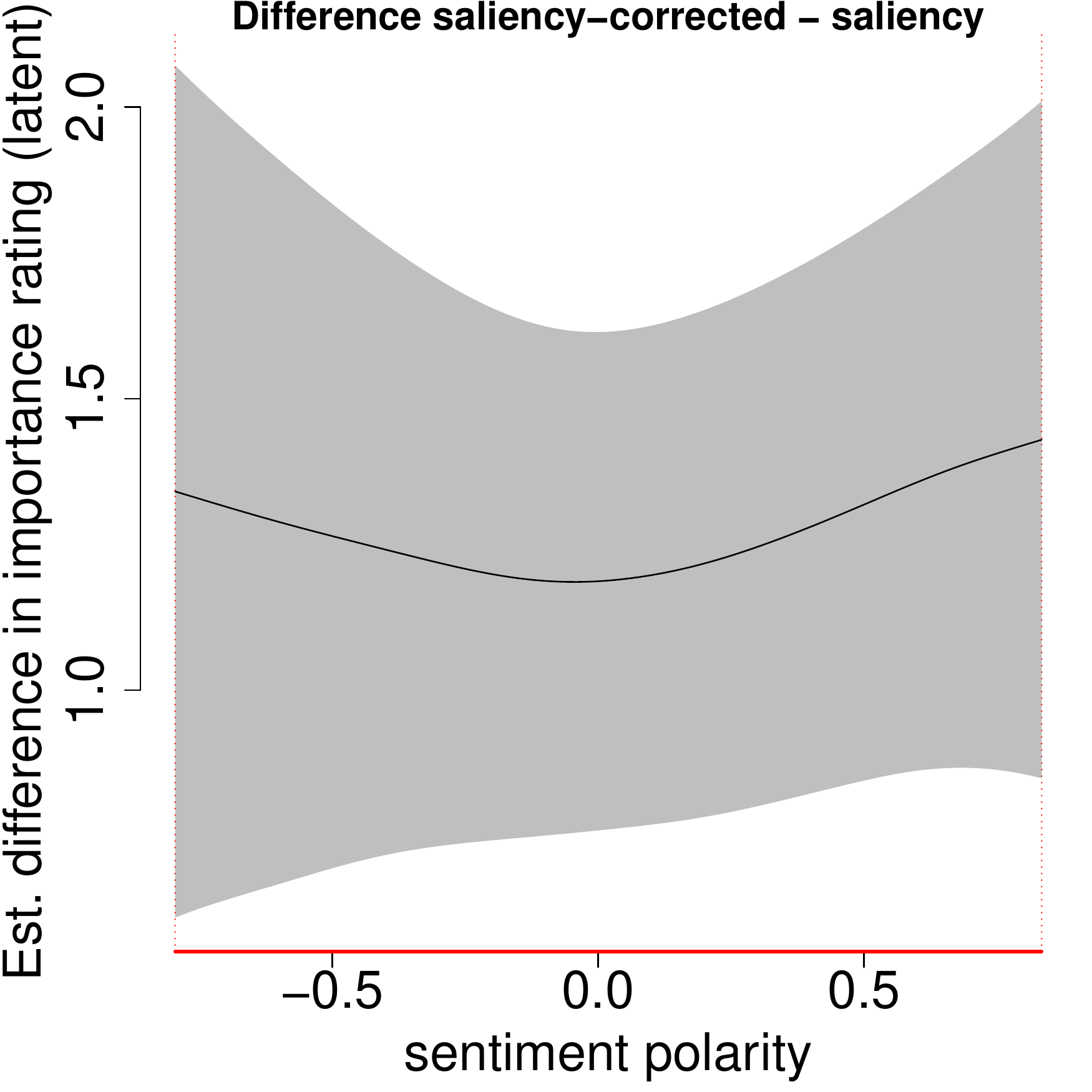}
    \caption{Sentiment polarity}\label{fig:difference_corrected-saliency_saliency_lemma_polarity}
    \end{subfigure}
    \begin{subfigure}[t]{.33\textwidth}
        \centering
    \includegraphics[width=\textwidth]{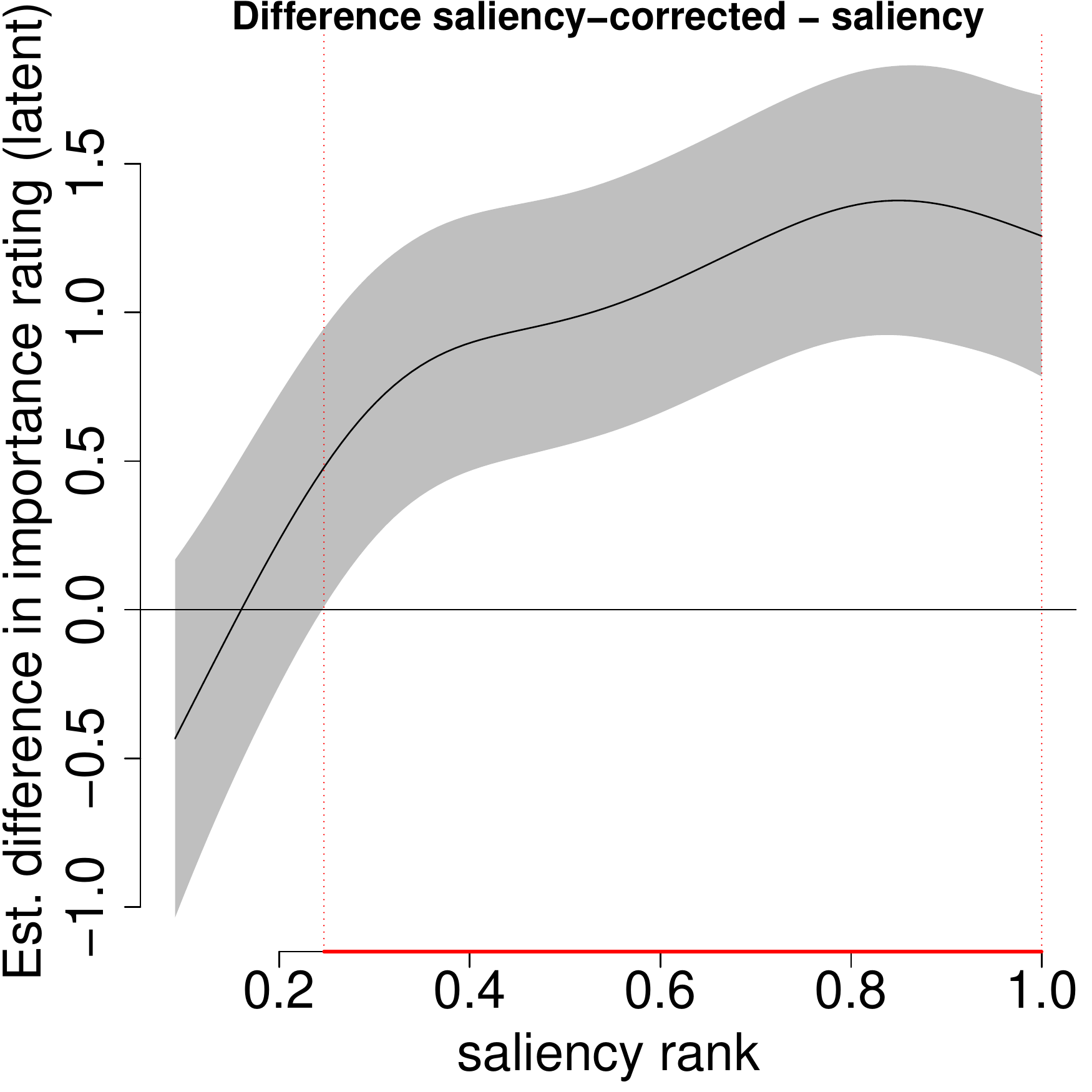}
    \caption{Saliency rank}\label{fig:difference_corrected-saliency_saliency_normalized_saliency_rank}
    \end{subfigure}%
    \hspace{1cm}
    \begin{subfigure}[t]{.33\textwidth}
        \centering
    \includegraphics[width=\textwidth]{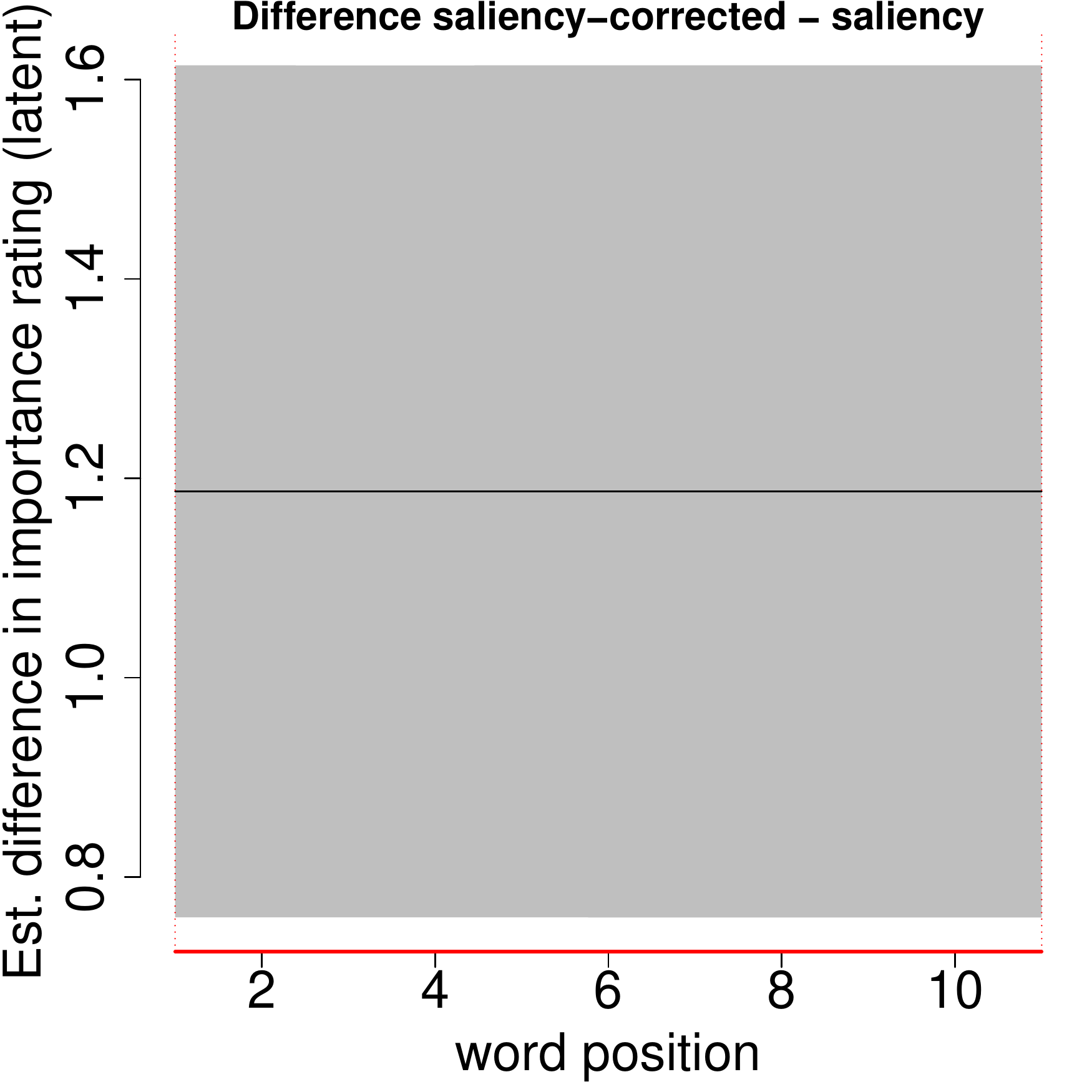}
    \caption{Word position}\label{fig:difference_corrected-saliency_saliency_word_index_in_sentence}
    \end{subfigure}
    \caption{Difference plots between the model-corrected saliencies and original saliencies. Areas of significant differences are marked red.}\label{fig:differences_corrected-saliency_saliency}
\end{figure*}

We report detailed estimates and test statistics regarding our third user study in Table~\ref{tab:model_params_corr_all}.
Figure~\ref{fig:comparison_correction_all} shows comparison plots for each smooth term and Figure~\ref{fig:differences_bars_saliency} as well as Figure~\ref{fig:differences_corrected-saliency_saliency} visualize the respective difference functions between visualizations along with highlighted regions of significant differences.
Cut points are located at -1, 0.95, 2.37, 3.67, 5.06 and 6.83.
\begin{table}[ht]
\centering
\ra{1.3}
\caption{Parametric and smooth coefficients of the GAMM corresponding to the third user study comparing the three visualizations.} \label{tab:model_params_corr_all}
\resizebox{1\linewidth}{!}{
\begin{oldtabular}{lrrrr}
   \hline
\textbf{Parametric Coefficients} & \textbf{$\beta$} & \textbf{SE} & \textbf{t} & \textbf{p} \\ 
  (Intercept) & 2.1119 & 0.1994 & 10.5909 & $<$ 0.0001 \\ 
  \hline
\rowcolor{lightgray}bars & -0.5991 & 0.1578 & -3.7974 & 0.0001 \\ 
\rowcolor{gray2}saliency-corrected & 1.1102 & 0.2515 & 4.4135 & $<$ 0.0001 \\ 
   \hline
\textbf{Smooth Terms} & \textbf{edf} & \textbf{ref. df} & \textbf{F} & \textbf{p} \\ 
  \hline
  s(saliency):saliency & 11.4304 & 19 & 283.3393 & $<$ 0.0001 \\ 
\rowcolor{lightgray}s(saliency):bars & 11.0767 & 19 & 321.0314 & $<$ 0.0001 \\ 
\rowcolor{gray2}s(saliency):saliency-corrected & 5.5202 & 19 & 113.9321 & $<$ 0.0001 \\ 
  s(display index):saliency & 1.4830 & 9 & 7.2492 & 0.2575 \\ 
\rowcolor{lightgray}s(display index):bars & 1.7044 & 9 & 15.3135 & 0.0254 \\ 
\rowcolor{gray2}s(display index):saliency-corrected & 0.0009 & 9 & 0.0001 & 0.6438 \\ 
  s(word length):saliency & 1.7724 & 9 & 4.1550 & $<$ 0.0001 \\ 
\rowcolor{lightgray}s(word length):bars & 0.0009 & 9 & 0.0001 & 0.3775 \\ 
\rowcolor{gray2}s(word length):saliency-corrected & 2.3645 & 9 & 1.3936 & 0.0213 \\ 
  s(sentence length):saliency & 0.0005 & 9 & 0.0001 & 0.2313 \\ 
\rowcolor{lightgray}s(sentence length):bars & 0.0004 & 9 & 0.0000 & 0.8967 \\ 
\rowcolor{gray2}s(sentence length):saliency-corrected & 2.4024 & 9 & 22.4406 & $<$ 0.0001 \\ 
  s(word frequency):saliency & 1.8086 & 9 & 2.3192 & $<$ 0.0001 \\ 
\rowcolor{lightgray}s(word frequency):bars & 1.7381 & 9 & 2.7043 & $<$ 0.0001 \\ 
\rowcolor{gray2}s(word frequency):saliency-corrected & 2.8913 & 9 & 7.2153 & $<$ 0.0001 \\ 
  s(sentiment polarity):saliency & 1.0751 & 9 & 0.4727 & 0.0633 \\ 
\rowcolor{lightgray}s(sentiment polarity):bars & 1.0022 & 9 & 0.5076 & 0.0507 \\ 
\rowcolor{gray2}s(sentiment polarity):saliency-corrected & 1.6991 & 9 & 2.2243 & 0.0020 \\ 
  s(saliency rank):saliency & 0.9279 & 9 & 2.0901 & 0.0002 \\ 
\rowcolor{lightgray}s(saliency rank):bars & 0.9764 & 9 & 6.5779 & $<$ 0.0001 \\ 
\rowcolor{gray2}s(saliency rank):saliency-corrected & 4.1893 & 9 & 6.8094 & $<$ 0.0001 \\ 
  s(word position):saliency & 0.0004 & 9 & 0.0000 & 0.9754 \\ 
\rowcolor{lightgray}s(word position):bars & 1.2970 & 9 & 0.7165 & 0.0167 \\ 
\rowcolor{gray2}s(word position):saliency-corrected & 0.0005 & 9 & 0.0000 & 0.9615 \\ 
  s(capitalization):saliency & 0.0009 & 2 & 0.0003 & 0.4268 \\ 
\rowcolor{lightgray}s(capitalization):bars & 0.0003 & 2 & 0.0001 & 0.4525 \\ 
\rowcolor{gray2}s(capitalization):saliency-corrected & 1.0644 & 2 & 3.2665 & 0.0245 \\ 
  s(dependency relation):saliency & 0.0057 & 29 & 0.0002 & 0.3443 \\ 
\rowcolor{lightgray}s(dependency relation):bars & 0.0010 & 28 & 0.0000 & 0.5819 \\ 
\rowcolor{gray2}s(dependency relation):saliency-corrected & 1.4715 & 28 & 0.0731 & 0.1955 \\ 
  s(condition order):saliency & 3.7653 & 6 & 30.7306 & 0.0044 \\ 
\rowcolor{lightgray}s(condition order):bars & 0.0007 & 6 & 0.0001 & 0.5619 \\ 
\rowcolor{gray2}s(condition order):saliency-corrected & 4.4665 & 6 & 150.1092 & $<$ 0.0001 \\ 
  s(sentence ID) & 12.7259 & 150 & 0.1028 & 0.2236 \\ 
  s(saliency,sentence ID) & 68.0861 & 150 & 1.7605 & $<$ 0.0001 \\ 
  s(worker ID) & 55.7637 & 59 & 313.9570 & $<$ 0.0001 \\ 
  s(saliency,worker ID) & 53.3619 & 60 & 230.3436 & $<$ 0.0001 \\ 
   \hline
\end{oldtabular}
}
\end{table}

\clearpage

\end{document}